\documentclass[botnum, fleqn]{dissert} 
\usepackage{cite}
\usepackage[utf8x]{inputenc}
\usepackage{amsmath,amssymb,amsfonts}
\usepackage{wrapfig}
\usepackage[pdftex]{graphicx}
\usepackage[table,xcdraw]{xcolor}
\usepackage{textcomp}
\usepackage{comment}

\usepackage{amsmath}
\usepackage{amssymb}
\usepackage{algorithm}
\usepackage{subcaption}
\usepackage{algpseudocode}
\usepackage{listings}
\usepackage{amsthm}
\usepackage{booktabs}
\usepackage{tabularx}
\usepackage{multirow}

\usepackage{xifthen}
\usepackage{mathtools}
\usepackage{bm}
\usepackage{xparse}
\usepackage{chapterbib}

\usepackage{lipsum}
\usepackage{subcaption}
\usepackage{hyperref} 
\hypersetup{
        colorlinks=true,                
        breaklinks=true,                
        urlcolor= red,                
        linkcolor= red,                
        bookmarksopen=false,
        filecolor=black,
        citecolor=blue,
        linkbordercolor=green
}

\AtBeginDocument{\hypersetup{pdfborder={1 0 0}}}

\usepackage{amsmath}
\usepackage{algorithm}
\usepackage{algpseudocode}
\usepackage{xcolor}
\usepackage{pdfpages}
\usepackage{xifthen}
\usepackage{comment}

\NewDocumentCommand{\vect}{ O{} O{} m }{\bm{#3}\ifthenelse{\isempty{#1}}{}{^{(#1)}}\ifthenelse{\isempty{#2}}{}{_{#2}}}

\NewDocumentCommand{\mat}{ O{} O{} m }{\bm{#3}\ifthenelse{\isempty{#1}}{}{^{(#1)}}\ifthenelse{\isempty{#2}}{}{_{#2}}}

\NewDocumentCommand{\ten}{ O{} O{} m }{\bm{\mathcal{#3}}\ifthenelse{\isempty{#1}}{}{^{(#1)}}\ifthenelse{\isempty{#2}}{}{_{#2}}}
\begin{document}

\frontmatter



\title{Integrating Deep Learning and Augmented Reality to Enhance Situational Awareness in Firefighting Environments}
\author{Manish Bhattarai}

\degreesubject{Ph.D., Engineering}

\degree{Doctor of Philosophy \\ Engineering}

\documenttype{Dissertation}

\previousdegrees{M.S., University of New Mexico, 2017}

\date{December, \thisyear}

\maketitle

\begin{dedication}
To my lord, ever-loving grandmother, my beloved parents and wonderful sisters. \\[3ex]
   ``We're here to put a dent in the universe. Otherwise why else even be here?''
         -- Steve Jobs
\setcounter{page}{3}
\end{dedication}

\begin{acknowledgments}
  \vspace{0.4in}
Firstly, I would like to express my gratitude to my advisor, Dr. Manel Martinez-Ramon, for his constant support and guidance. I feel honored to have had the opportunity to work under your guidance. Without your constant encouragement, it would not have been possible to complete my Ph.D. dissertation. 
\par
I would also like to thank my dissertation committee members Dr. Ramiro Jordan, Dr. Marios Pattichis, and Dr. Trilce Estrada, for their encouragement and valuable suggestions.
\par 
I want to express special thanks to Sophia Thompson for her constant motivation and support throughout my Ph.D. She has been a great companion for my growth and an undeniable source of inspiration. Besides, I am also thankful to my friends Anees Abrol, Pankaj Das and Abraham Vinod, who are always there to keep pushing and motivating me and show directions to me when I am in need.
\par
I would also like to thank my mentor at Los Alamos National Labs, Dr. Boian Alexandrov, for the valuable suggestions and for providing me an opportunity for growth in the field of research. Further, my co-mentors Benjamin Nebgen and Erik Skau have been very helpful in showing new research directions. I am also glad to have my friends Javier Santos, Renan Rosas, Hector Carillo, Miguel Herrera, whom I met during applied machine learning school at LANL and helped me grow a lot in the field of applied deep learning.  
\par
Finally, I am glad to have friends in Albuquerque who gave me more support and love. Finally, I thank my family for their love, support, and encouragement throughout this journey.   This work would not have been possible without my parents and family.  I dedicate this work to them.
\clearpage
\end{acknowledgments}

\maketitleabstract 

\begin{abstract}
    We present a new four-pronged approach to build firefighter's situational awareness for the first time in the literature. We construct a series of deep learning frameworks built on top of one another to enhance the safety, efficiency, and successful completion of rescue missions conducted by firefighters in emergency first response settings. First, we used a deep Convolutional Neural Network (CNN) system to classify and identify objects of interest from thermal imagery in real-time. Next, we extended this CNN framework for object detection, tracking, segmentation with a Mask RCNN framework, and scene description with a multimodal natural language processing(NLP) framework. Third, we built a deep Q-learning-based agent, immune to stress-induced disorientation and anxiety, capable of making clear navigation decisions based on the observed and stored facts in live-fire environments. Finally, we used a low computational unsupervised learning technique called tensor decomposition to perform meaningful feature extraction for anomaly detection in real-time. With these ad-hoc deep learning structures, we built the artificial intelligence system's backbone for firefighters' situational awareness. To bring the designed system into usage by firefighters, we designed a physical structure where the processed results are used as inputs in the creation of an augmented reality capable of advising firefighters of their location and key features around them, which are vital to the rescue operation at hand, as well as a path planning feature that acts as a virtual guide to assist disoriented first responders in getting back to safety. When combined, these four approaches present a novel approach to information understanding, transfer, and synthesis that could dramatically improve firefighter response and efficacy and reduce life loss.
    \setcounter{page}{5}
\end{abstract}

\tableofcontents
\listoffigures
\listoftables

\mainmatter

\chapter{Introduction}
\section{Overview}
\subsection{Firefighting Loss Statistics}
According to the last report from the US Fire Administration [USFA]\footnote{\href{https://www.usfa.fema.gov/downloads/pdf/publications/ff_fat17.pdf}{https://www.usfa.fema.gov/downloads/pdf/publications/ff\_fat17.pdf}}, around 13 million fires were declared in 2015, of which 380,900 were in residential buildings. The incidents in residential buildings produced 2565 deaths ($78\%$ of fire-related deaths in 2015)
and 11,475 injuries ($73\%$ of all fire-related injuries in 2015). 
Estimated losses due to residential fires totaled 7.1 billion ($49\%$ of total losses due to fires). While the number of fire incidents
has decreased over time (for example, there were 1,642 million fires in 2006 and
One thousand three hundred forty-five million in 2015), the number of fatalities is almost constant, with a mean of 3200 victims a year
since 2006. The National Fire Protection Association [NFPA] 2016 Annual Fire Loss report states that, while the number of fires is decreasing, the risk of death in the event of a fire has
remained relatively constant for the period from 1977 to 2016. This suggests that the improvements made to firefighting response have yet to address those responsible for the fatality.
\par
\subsection{Firefighter Current Strategy}
Fighting fires of any kind is a dangerous mission. Success hinges on making many decisions based on small amounts of current information by firefighters in high-stress environments. Firefighters must continuously assess the situation,
plan their next set of actions, and coordinate with other team members. Currently, these decisions and the resulting actions are often made with an incomplete picture of the current
situation. At the other end of the spectrum, firefighters' fireground communication approaches and the field commander’s decision-making had not changed significantly since the early 1970s when portable
radio devices became part of fire departments' standard equipment. This unfortunate fact is well reflected in
the annual statistics by the US Fire Administration, wherein one can observe that the fatality rate of the on-duty firefighters is, if not worsening, not getting better in recent decades. There
is a gap between the state-of-the-art communication and information technology being developed and existing firefighting protocols.
\par
Coordination of a multitude of simultaneously occurring operations during fire suppression and rescue operations is imperative to mission success and relies heavily on situation awareness. Maintaining situational awareness, defined as knowledge of current conditions and activities at the scene, is critical to accurate decision making. Firefighters often carry various sensors in their equipment, namely thermal cameras, gas sensors, and microphones. Current fire fighting systems do not involve any automated detection and characterization of the fire ground. The detection of objects of interest and the estimation of the firefighters’ health condition or other persons in the fire ground might be slowed down and obscured due to several factors that include smoke, fire, or other elements or anxiety levels exhaustion of the active firefighters. Under such hazardous conditions, firefighter and others’ lives may be compromised due to the rescue operation decisions based on incomplete or inaccurate environmental conditions. Firefighter mortality statistics are primarily due to inefficient decision-making protocol derived from data-informed by human senses impaired by on-the-ground conditions. \par

\subsection{Situational Awareness}
According to a report by the Department of Homeland Security \cite{royal2014project}, situational awareness is defined as “capability to obtain and distill specific knowledge concerning threats, hazards and conditions in a timely matter to support incident management decisions across all phases of a catastrophic incident response, including (1) The ability to know the location of responders and their proximity to risks and hazards in real-time,(2)The ability to detect, monitor and analyze passive and active threats and hazards at incident scenes in real-time,(3) The ability to rapidly identify hazardous agents and contaminants and (4) The ability to incorporate information from multiple and nontraditional sources (for example, crowdsourcing and social media) into incident command operations”. As per the definition, the terminology emphasizes the need for real-time data availability with an increased speed and volume of flow of information for the responders. Also, situational awareness, more than just providing extra information to firefighters, is based on the interplay of each individual’s attention, memory, and decision for action.  \par

\subsection{Firefighter Problems}
During a firefighting operation, the context is characterized by the criticality of various components such as time sensitivity, uncertainty in a dynamic environment, poorly defined problem structure with competing priorities (i.e., multiple goals but an unclear path to solving them quickly and efficiently due to lack of on the ground knowledge),  high stakes, numerous players and rapidly changing information flow due to inputs from different sources. Under such a critical situation, one needs to consider the diverse components of information ranging from auditory to visual information elements, including extent and size of the fire, floor plan of the building, accessibility, status of the victims, status of responders, etc.  Firefighters heavily rely on communication between one another to maintain adequate situational awareness. Incident commanders are responsible for making the final decisions based on established protocols and past experiences. When communication systems break, individual responders are responsible for making decisions based on their immediate surroundings and teammates’ interpretation. They may not be optimal in highly dynamic and extreme situations. We analyzed past fire response events to identify successful and unsuccessful management components to better understand those aspects that could benefit from technological assistance. Our analysis results showed communication issues as one of the components most often associated with operation failure. Improvements targeting this component are the focus of our research.
\par
The underlying drivers during a fire response can be summarized as 
i) Time sensitivity and pressure necessitating the simultaneous coordination and implementation of planning and execution phases,
ii)	A framework of communication between fire personnel for collaborative decision making and
iii) Uncertainty /unpredictability of such events. 
The above objectives are based on our analyses of past events, which show that by assessing threats, harm, risks, communication strategy, and coordination policy, a more effective fire response will result, and fatality figures can be improved. 
\section{Motivation}
Lack of effective data processing, interpretation, and communication is currently a significant hindrance to decision making on the fire ground. Enhanced data processing techniques can mine this data to effectively improve situational awareness at all times, thereby improving real-time decision making and minimizing errors stemming from environmentally induced poor judgment. We propose a supplemental situational awareness system that uses the firefighter gear’s information to extract the necessary knowledge about the firefighter’s environment at any given time and transmit it to the Command. This research employs state-of-the-art Machine Learning (ML) techniques to create an automated system capable of real-time object detection and recognition utilizing currently gathered data to achieve improved situational awareness of firefighters on the scene.

\section{Literature Review}

In this section, we review the works related to each of the chapters of this dissertation. Besides, we also discuss the limitations of those works and describe our contributions to address those limitations. 

\subsection{Prior Work on Machine Learning for Emergency Response}
\begin{figure}[!ht]
	\centering
	\includegraphics[width=.7\textwidth]{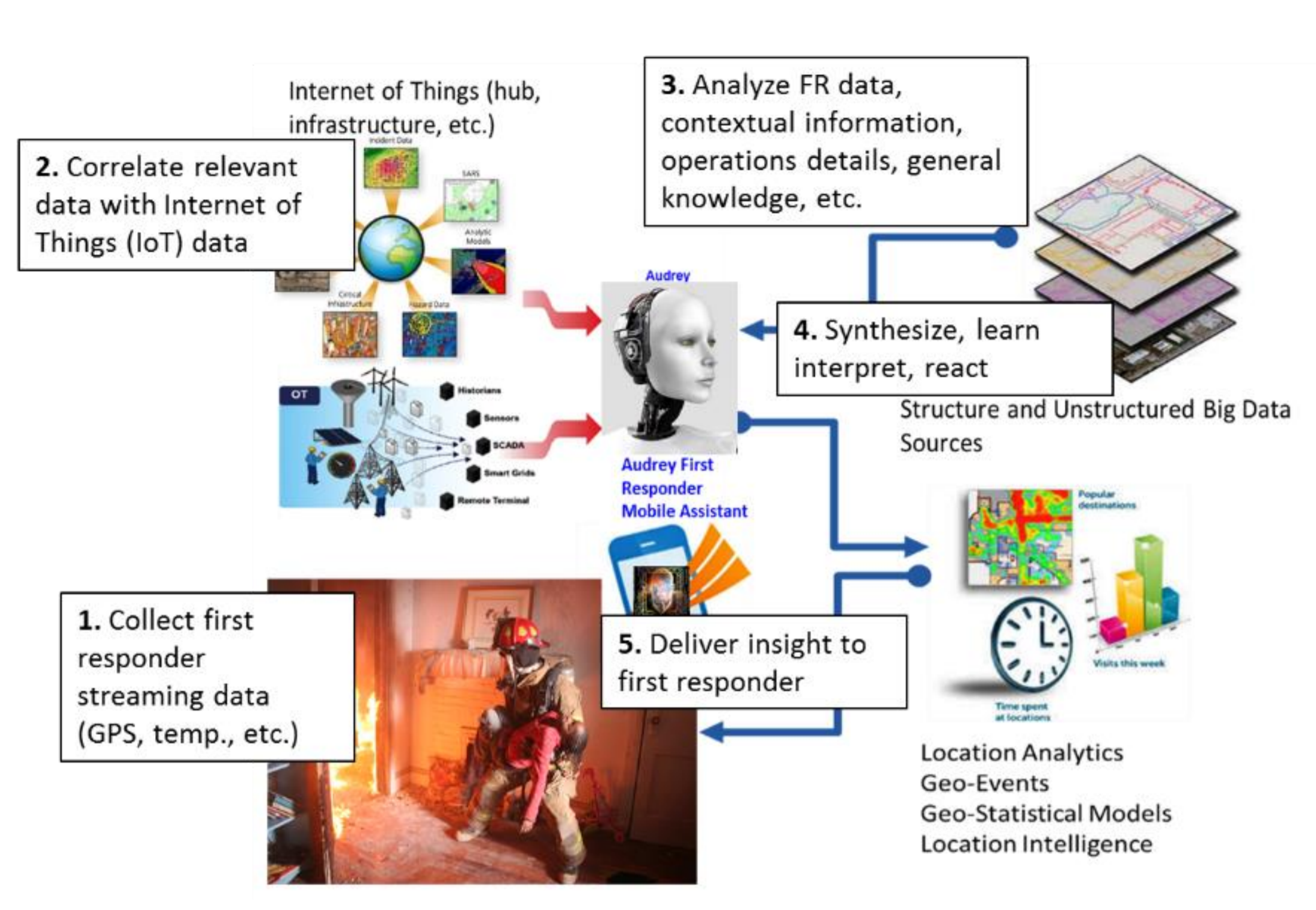}
	\caption{Subsystem interaction in the AUDREY}
	\label{fig:audrey1l}
\end{figure}

The idea of applying ML in emergency response systems is not new.  The first version of the platform Assistant for Understanding Data through
Reasoning, Extraction, \& sYnthesis(AUDREY) by NASA was released in 2016. This platform uses the idea of collecting data from firefighter’s sensors to create a situational awareness system that warns the Command about the potential risks in the fire ground and sends recommendations to the firefighters on duty. To date, the project Audrey, a collaboration between NASA, JPL, and the Department of Homeland Security, uses deep learning technology for cognitive reasoning to assist the first response teams \cite{castro2016promise}. An overview of the framework is shown in \ref{fig:audrey1l}.

Current various state-of-the-art artificial intelligence (AI) technologies show great promise in their application to emergency response environments. Our approach takes these technologies, powerful in their own right, and integrates them to form a new system with a broad capability. Only recently have other such integrations reached public outlets. As per \footnote{\href{https://c4i.gmu.edu/wp-content/uploads/AFCEA-GMU-2018-Chow.pdf}{https://c4i.gmu.edu/wp-content/uploads/AFCEA-GMU-2018-Chow.pdf}}, AUDREY is the first published application which displays a similar integration. AUDREY uses bio-inspired Neural-Symbolic Processing through a mixed neural and symbolic processing ability for higher-level cognitive reasoning. AUDREY’s underlying capabilities are:
Overlapped reasoning and learning timings.
Ability to analyze missing/biased data sets.
Ability to synthesize workflows for visual/textual/audio questioning and answering.
Learn from human community interaction. 
\par
\begin{figure}[!ht]
    \centering
    \includegraphics[width=.7\textwidth]{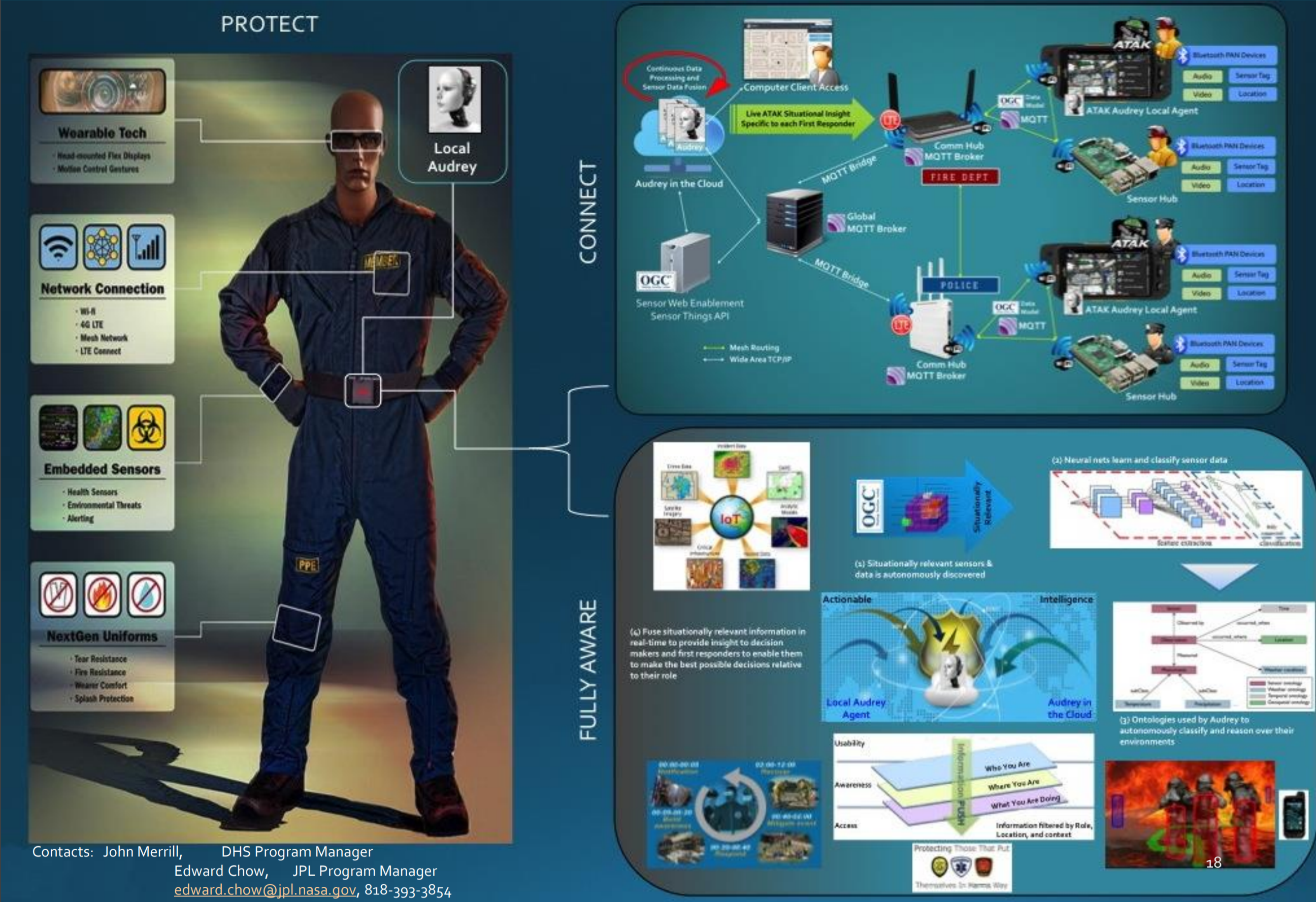}
    \caption{ AUDREY framework}
    \label{fig:audrey3}
\end{figure}

 Our approach, though developed independently prior to any publications covering Audrey, employs a similar situational awareness reasoning process as AUDREY, which incorporates the following steps:
Autonomously collect the data from situationally relevant sensors.
Use a Machine learning/deep learning framework to learn and classify the sensor data.
Use Incremental Learning to organize and reason over the performing environments autonomously.
Create a guide that synthesizes situationally relevant information in real-time into a system that can advise decision-makers and first responders of current local conditions and those areas of a scene that are undergoing rapid change.
This autonomous guide’s synthesis of this information can deliver pertinent details necessary to Command and firefighters to make more informed decisions. This process can be visually observed in \ref{fig:audrey3}.

\subsection{Prior Work on Object Detection and Classification in Firefighting Environments}
The application of CNN technology abounds in the Surveillance and Defense fields  \cite{4,5,6,7}. Still, very little research is documented in applying these principles to overcoming the navigational challenges faced by firefighters in live-fire events. The conventional fire alarm system uses different sensors such as temperature, UV, and fire detectors to determine the presence of fire, smoke, and other hazards \cite{8,9}. Their long-established usage, as evidenced by the prevalence of such sensors in all buildings and is a requirement in building codes. However, such detectors typically have a long response time in large spaces \cite{10}. Furthermore, they do not provide any more specific spatial information regarding detected hazards in the given scenario. More contemporary firefighting modalities for detection of fire, smoke and other targets in a fire environment rely on color \cite{10,11,12,13}, motion \cite{14,15} and texture \cite{13,16} features of the captured image. These vision-based approaches use histogram thresholding, optical flow- based motion vector computations, and texture analysis. These more modern techniques provide enhanced performance using RGB imagery and improve the more conventional techniques described above. These algorithms do not perform as robustly on Infrared (IR) or darker imagery and require longer computational time. IR imagery lacks sufficient complexity needed by such algorithms to perform well. Conversely, RGB images are hard to classify in active fire environments due to heavy smoke and poor lighting. IR image technology fills this gap. Furthermore, the presence of fire and smoke, by its very nature, creates a non-stationary environment and renders most existing stationary vision-based detection systems ineffectual in informing decision-making processes in real-time. To address this issue, a robust real-time detection system based on CNN, which can detect and localize the target of interest instantaneously, is proposed. The research presented in \cite{17} describes the usage of infrared images to extract motion and statistics features in real-time using a Bayesian classifier for multi-class identification. Significant research has also been done in human detection in other dark/ low visibility environments utilizing a single visible camera and fusion of the generated RGB image with an IR data set. Single IR camera-based detection mechanisms have been presented in \cite{18,19,20,21}. Most of these approaches use HOG-based feature extraction and a classifier using SVM or other ML-based techniques \cite{20,22}. Paper \cite{18} presents the use of a GMM system in human detection. Template matching techniques and thresholding techniques have also been reported in \cite{20,21,23}.

Other published works that have influenced our research perform classification tasks on thermal imaging. In \cite{Cho2018}, a CNN is used in material recognition with non-firefighter grade thermal cameras. A related study is reported in \cite{Vandecasteele2017}, where transfer learning is used to detect objects in a fireground. Nevertheless, the number of data samples used in the Vandecasteele paper is low, so deep learning cannot be applied. Saliency detection and Convolutional Neural Networks are applied to detect wildfires in \cite{Zhao2018}. 
In \cite{Kim2016}, a Bayesian procedure to detect fire and smoke and differentiate them from thermal reflections in infrared is used in \cite{Yun2018}. In the recent work \cite{Ajith19}, authors introduce a methodology based on Random Markov fields to segment fire, smoke, and background in a sequence of images.  

Further works that use thermal imagery and deep learning include \cite{Zhang2018}, where authors use CNN to detect known objects in infrared surveillance cameras. In \cite{Wang2016}, CNN is applied to the detection of vehicles in thermal imagery. Researchers in \cite{DeOliveira2018} introduce the use of CNNs to detect pedestrians to apply detection to unmanned aerial vehicles (UAV). Another application in UAV is presented in \cite{Rodin2018}, where authors train a CNN structure to detect objects of interest, such as bodies or body parts and other objects related to victims of avalanches.  A similar approach uses CNNs on long-wave infrared imagery to detect objects. Work \cite{Valldor2014} uses deep learning to detect people utilizing a semi-supervised approach that takes advantage of a large quantity of non-labeled images containing humans.

Panic detection work has utilized machine learning techniques to analyze video footage, typically acquired from stationary surveillance cameras. Two papers provide excellent summaries of the techniques authored to date, focused on crowd analysis and panic detection.  \cite{panic_zhang1} covers crowd state detection methods utilizing stationary RGB video surveillance footage capable of detection at micro/macro levels to analyze local/global individual movements in the frame. The analyses usually focus on small subsets of a crowd, which are then aggregated together to achieve global attributes. The analyses summarized discussed two possible approaches in their frameworks, physics-based and machine learning-based. Physics-based models use collected motion such as velocity, correlation function, fluid dynamics, energy and entropy, force model, and complex systems. Machine learning-based models utilize features extracted through signal processing/computer vision tools to detect crowd state. The panic detection algorithms are applied to three different types of datasets, which are generated under three situations: 1) controlled experiment 2) crowd model and simulation 3) crowd video surveillance. \cite{panic_kok} covers crowd state detection and compares a stationary crowd (the body movements of people who do not move) to a dynamic crowd (people moving from one place to another). The analysis of dynamic crowds focuses on movement patterns of the individuals in the scene to infer activity. The motion vectors detected by classical image processing techniques such as frame difference or optical flow are analyzed to deduce crowd activity. Information is then gathered by processing the frames for a dynamic crowd. The camera position is required to be fixed to obtain realistic motion features. \cite{panic_kok} found that very few studies have focused on panic detection within stationary crowds due to the challenges involved in detecting panic behaviors from small body movements such as expression or posture. However, in both stationary and dynamic crowd types, the camera’s fixed position is a requirement. These methodologies also require the analysis of a sequence of frames processed together to compute the motion vectors. In both papers  \cite{panic_zhang1} and \cite{panic_kok}, panic behavior is primarily based on the rate of displacement computed for extracted features from sequences of frames and requires the camera to be stationary.

Our work’s primary focus is object detection of key features like entrance/exit points (doors and windows) and persons needing rescue within an active fire environment. We also add a posture detection framework to assist in the prioritization of rescue. Due to the nature of the generation of our data (a handheld thermal imagery camera carried by one of the responding firefighters throughout the scene and thus non-stationary), the pose detection is limited to basic poses (sitting, standing, crawling) that are easily distinguishable in individual frame analyses and are therefore not dependant upon the surrounding images in a sequence. Furthermore, these largely different poses are not reliant on high detail levels and can be discerned in IR imagery. The application of the algorithms deployed in the panic as mentioned above detection research is difficult to deploy on our dataset because the moving camera necessary to complete our research objectives would produce a much noisier motion vector for the targets, and those movements needed to deduce panic state would be indistinguishable from those induced by the camera movements. To minimize such errors generated by camera movements in our dataset, we process one frame at a time and infer activity rather than relying on motion vectors determined from a sequence of frames analyzed together. 
  
\subsection{Prior Work on Object Detection and Tracking in Firefighting Environments}
Before the inception of CNN applied to object detection and tracking models, image processing techniques for preprocessing and Machine Learning (ML) techniques for recognition were widely used. The preprocessing techniques involved template matching, fixed thresholding, adaptive thresholding, background subtraction, feature extraction with Scale Invariant Feature Transform (SIFT), and Speeded Up Robust Features (SURG) methods; the ML techniques involved were Graph cuts, K-means clustering, and Gaussian mixture models.
One of the notable works in object detection and tracking is presented in \cite{ berg2016detection }, where the author uses a template-based tracking method for short-term and single-shot tracking on thermal infrared imagery. \cite{ jungling2009feature} uses SURF features whereas \cite{zhang2007pedestrian} use histogram of oriented gradients(HOG) for feature extraction for person detection. \cite{ benezeth2008real} use Gaussian back-ground model whereas \cite{jeon2015human} use a background subtraction technique for tracking people in infrared images. \par

Deep learning-based frameworks gained popularity in recent years for the detection and tracking task. The state-of-the-art CNN architectures for RGB datasets include: (1) single-pass approaches that perform detection within single step (single shot multi-box detector (SSD) \cite{liu2016ssd}, You Only Look Once (YOLO) \cite{redmon2016you}) and (2) region-based approaches that exploit a bounding box proposal mechanism prior to detection (faster Regional-CNN (R-CNN) \cite{ren2015faster}, Region-based Fully Convolutional Networks (R-FCN) \cite{dai2016r}, and lightweight deep neural networks for real-time object detection (PVANET) \cite{kim2016pvanet}). However, these frameworks do not work with infrared imagery without reconfiguration due to a lack of equivalent dynamic range and texture information in infrared images. Furthermore, due to lower resolution with poor pixel variations, a deep CNN backbone model may fail to extract significant features required for region-based proposal networks and feature pyramid networks, resulting in poor detection and tracking performances. \par

\cite{ liu2017deep } proposes a correlation filter based ensemble tracker with multi-layer convolutional features for thermal infrared tracking (MCFTS). The authors first use pre-trained convolutional neural networks to extract the features of the multiple convolution layers of the thermal infrared target, followed by applying a  correlation filter to construct multiple weak trackers with the corresponding convolution layer features. In \cite{ park2020cnn}, the authors develop a CNN framework for human-image detection with a Resnet backbone for feature extraction, which is processed by an atrous spatial pyramid pooling followed by a bilinear interpolation to generate the per-pixel based segmentation mask for the object. \cite{ zhang2018novel} first uses a super-resolution method to increase the target signature resolution followed by a faster RCNN- based CNN framework for detecting and tracking objects of interest for a surveillance operation. \cite{ herrmann2018cnn} uses an advanced approach called domain adaptation where they minimize the domain gap between the RGB and thermal images by preprocessing strategy followed by finetuning of an RGB pre-trained NN with thermal images. \cite{ivavsic2019human} use a YOLO- based CNN and \cite{valldor2018person} use a retinanet based CNN for human detection in infrared images. \par

For the segmentation task, Watershed \cite{mangan1999partitioning}, Otsu \cite{ otsu1979threshold } and Graph-Cut \cite{boykov2006graph} are among the well established classical methods. Other techniques are Thresholding and Edge-based Segmentation. However, their performances are limited when applied to infrared imagery due to the lack of features. Current state-of-the-art deep learning-based techniques for segmentation include  UNET \cite{ronneberger2015u}, RESUNET \cite{xiao2018weighted}, SEGNET \cite{badrinarayanan2017segnet}, MASK-RCNN \cite{he2017mask}, DeepLab \cite{chen2017deeplab}, FastFCN  \cite{wu2019fastfcn} and newly proposed space-time memory networks for video segmentation \cite{ oh2019video}.  Despite their proven segmentation performance in RGB benchmark datasets, most of these frameworks fail for the same task in infrared datasets. One of the works that apply deep learning-based techniques on infrared imagery for object segmentation includes \cite{masouleh2019development}, where the authors improve the deep learning combined model’s performance. Gaussian-Bernoulli Restricted Boltzmann Machine (GB-RBM) specifications for the segmentation of the ground vehicles from UAV-based thermal infrared imagery. \cite{nan2019infrared} proposed an improved mask-RCNN where they combine the advantages of generative adversarial network and mask-RCNN for the segmentation of the object of interest. \par

While several works exist covering object detection, tracking, and segmentation applied to infrared images, none of these techniques work in firefighting scenarios where detecting fire, humans, and objects of interest simultaneously is the objective. We address these problems with a novel Mask RCNN framework capable of performing image detection and segmentation tasks concurrently.

\subsection{Prior Work on Scene Description in Infrared Images}
Although several works have been published covering image captioning with RGB images,  no image captioning framework for infrared images exists. Most of the caption generation’s prior work covers caption generation of the whole image, where the scene is described categorically with broad-level scene details. \cite{karpathy2015deep} uses images and corresponding captions to learn about the inter-model correspondence between language and visual data. The framework combines a standard visual CNN with bidirectional Recurrent Neural Net(RNN) with an objective of manifold alignment on a multi-modal embedding.  \cite{liu2017mat} proposes a methodology to sequence RNN for image caption generation through a multi-modal translation task. The authors represent the input image as a sequence of detected objects and feed to the RNN model to generate a sequence of words at the RNN model’s output. In \cite{lu2017knowing}, the author proposes an adaptive attention model with a visual sentinel. In other words, the model intelligently determines where to attend to the image to extract meaningful information for sequential word generation. The show and tell framework presented in \cite{vinyals2016show} uses a generative model based on the Long Short Term Memory(LSTM) model that combines the CNN for visual feature extraction and machine translation to generate natural sequences describing the image. Our framework is based on this model, where the objective for training is to maximize the likelihood of a target description sentence given the input training image. The authors in \cite{xu2015show} extend the work in \cite{ vinyals2016show} by introducing an attention-based model to learn to describe the content of images automatically. \cite{ yang2016review } introduces an attention-based review network capable of performing several review steps to output a thought vector after each review step. \par
Most of the above frameworks cannot be applied directly to thermal images. Most of them tend to be data-hungry, and there is no captioning benchmark dataset corresponding to infrared images. Also, the lack of sufficient infrared features causes a huge domain gap that adds challenges in embedding the visual and word feature spaces. 
\subsection{Prior Work on Embedded Platform Implementation}
 The firefighter’s life and those needing rescue rely on the firefighter’s ability to make accurate decisions during scene navigation, and recent advancements in deep learning \cite{lecun2015deep}, data processing, and embedded systems technologies now offer a new way to assist and improve this decision-making process.
\par
Indeed, such technologies are already being implemented in firefighting applications. Just within the last decade, these applications have grown to include the use of wireless sensor networks (WSN) for early fire detection and response \cite{wilson2007wireless}, virtual reality platforms for firefighter training \cite{vichitvejpaisal2016firefighting, yuan2012building, yang2018train}, and autonomous or semi-autonomous firefighting robots \cite{lawson2016touch, ranaweera2018shortest,liu2016robot}. But fighting fires poses many challenges to the firefighters themselves that are not adequately addressed by these solutions. Continuously changing life-threatening situations can engender severe stress and anxiety and induce disorientation, leading to inaccurate decision making and decreased situational awareness. Recent research \cite{li2014situational} has demonstrated a discrepancy between the information available during fire emergencies and the information required for critical situational awareness and the importance of information accuracy to error-free decision making. \par

During firefighting operations, the most important things that the incident commander needs to know for enhanced operational capability and responder safety are building floor plans, team members’ location, and the status of critical supplies such as oxygen levels in real-time. The early usage of  head-mounted display reported in the literature are for crime investigation \cite{baber2005wearable}, firefighting operations \cite{wilson2005design} and paramedic use \cite{sasse1999coordinating}. It has also been reported that proper and well-designed wearable technologies can augment situational awareness in firefighters as they tend to change the conscious experience of humans, correcting their attention, intentions, reaction rate, and delivering meaningful insights about the situation. 
\par
Inaccurate judgment calls are consistently among the top causes of firefighter fireground injuries each year, and firefighter personal protective equipment (PPE), contrary to its purpose, is often a contributing factor \cite{petrucci2016inaccuracy}. Although firefighter PPE is designed to protect the wearer during extreme fire conditions, it is not without its drawbacks, including reduced visibility and range of motion \cite{petrucci2016inaccuracy,kozlovszky2014environment}. These drawbacks have led to integrating various sensors, such as infrared cameras, gas sensors, and microphones, into firefighter PPE to assist and monitor firefighters remotely \cite{kozlovszky2014environment}. The data acquired by these sensors can also be leveraged for additional purposes: to provide firefighters with object recognition and tracking and 3D scene reconstruction in real-time, creating a next-generation smart and connected firefighting system.

\subsection{Prior Work on Path Planning and Navigation}
This section’s work is based on two distinct fields: 1) Path planning and navigation, and 2) deep reinforcement learning.  
\subsubsection{Path Planning and Navigation}


A large amount of work focused on path planning and navigation to aid firefighting has been done, but few works address dynamic, continuously changing environments. \cite{su2012path} proposes a mobile robot with various sensors to detect fire sources and use the so-called A* search algorithm for rescue. An algorithm based on fire simulation to plan safe trajectories for an unmanned aerial system in a simulator environment is presented in \cite{beachly2018fire}. \cite{jarvis2005robot} shows the efficacy of the covert robotic algorithmic tool for robot navigation in high-risk fire environments. The usage of an ant colony optimization tool to automatically find the safest escape routes in an emergency in a simulator environment is shown in \cite{goodwin2015escape} whereas \cite{zhang2020path} formulates the navigation problem as a “Traveling Salesman” problem and proposes a greedy-algorithm-based route planner to find the safest route to aid firefighters in navigation. \cite{ranaweera2018shortest} proposes a particle swarm optimization for shortest path planning for firefighting robots, whereas \cite{zhang2018intelligent} proposes approximate dynamic programming
to learn the terrain environment and generate the motion policy for optimal path planning for UAV in forest fire scenarios. A methodology for path reconstruction based on the analysis of thermal image sequences is demonstrated in \cite{vadlamani2020novel}, which is based on the estimation of camera movement through assessment of the relative orientation with SIFT and Optical flow. \par

Despite the large quantity of work in the literature to aid the firefighters in path planning and navigation, most tend to solve the path planning considering a static environment where a one-time decision is made to guide the agent from source to destination. Such algorithms fail when the environment is dynamic. Furthermore, these algorithms do not allow the agent to make immediate decisions when encountered with a sudden fire in the chosen navigation path. We propose a deep reinforcement learning-based agent capable of making an instantaneous decision based on past learned experiences when subjected to sudden environment changes during navigation. 



\subsubsection{Deep Reinforcement Learning}
Reinforcement learning (RL) is a technique that tends to learn an optimal policy by choosing actions based on maximizing the sum of expected rewards. Even though several works exist for path planning in fire environments, no RL based path planning implementations were found for fire scenarios. Outside of the fire scenario, several RL based path planning implementations do exist. \cite{romero2016navigation} demonstrates an RL based navigation of a robot, which is provided with a topological map.\cite{li2006q} uses a Q-learning based path planning for an autonomous mobile robot for dynamic obstacle avoidance. 
An RL based complex motion planning for an industrial robot is presented in \cite{meyes2017motion}. 
RL integrated with deep learning has demonstrated phenomenal breakthroughs that are able to surpass human-level intelligence for computer games such as Atari 2600 games \cite{ mnih2013playing} \cite{ mnih2015human} , AlphaGo zero \cite{ silver2017mastering} \cite{tang2017recent} along with various other games. In these frameworks, the AI agent was trained by receiving only the game’s snapshots and the game score as inputs. Deep RL (DRL) has also been used for autonomous navigation based on inputs of visual information \cite{surmann2020deep} \cite{kiran2020deep}. \cite{bae2019multi} Proposes a multi-robot path planning algorithm based on deep Q-learning, whereas  \cite{lei2018dynamic} demonstrates the autonomous navigation of a robot in a complex environment via path planning based on deep Q-learning(DQL) with SLAM. 
Most of these deep reinforcement learning-based path planning and navigation tasks are based on visual input, i.e., raw images/depth data, which encodes the information about the environment. Based on this information, the navigation agents can establish the relationship between action and the environment. The agent in the DRL system embeds the action-policy map in the modal parameters of the neural nets. \par

Despite the efficacy of the DRL system in navigation, they are based on a learning experience of trial and error where the agent goes through numerous failures before actual success. Considering the fire-environments’ dangerous behavior, training, and evaluation of such DRL systems is very dangerous and practically infeasible. In addition to that, it is costly as well as time-consuming. To address these challenges, we developed the RL agent’s training environment in a virtual gaming environment Unreal Engine. The virtual environment depicts the actual firefighting scenario and enables the user to collect many visual observations for action and reaction in various fire environments. The agent can interact with the environment through the actions and be trained with multiple user-defined rewards and goals.  The framework also allows a  plug and play option for the firefighting environment. One can depict a variety of fire scenarios from structural fire to wildfires for training the DRL agent. 

\subsection{Prior Work on Tensor Decomposition in Infrared Images}
Earlier works on tensor decomposition for computer vision in thermal images are presented in this section. \cite{song2019ensemble} proposes an ensemble tensor factorization to extract defects signal of infrared thermography videos by jointly modeling the weighted low rank and sparse tensor patterns. Similarly, \cite{khanzadeh2018dual} proposes a tensor decomposition-based approach to detect the process changes in an additive manufacturing process using thermal images. \cite{song2020ensemble} uses ensemble tensor decomposition to extract the weak signal from infrared thermography videos for crack detection. In the paper \cite{lopez2020spectral}, the authors first use the higher-order orthogonal iteration(HOOI) for a reduced dimensional representation of the hyperspectral data followed by a fully connected network to classify the reduced dataset.
On the other hand,\cite{ wu2020hierarchical} develops a hierarchical low-rank and sparse tensor decomposition to mine anomalous patterns in thermal images for defect detection. \cite{ahmed2020sparse} proposes a low-rank tensor with a sparse mixture of the Gaussian decomposition algorithm for natural crack detection. The authors in \cite{zhang2015compression} use the hyperspectral image compression and reconstruction method based on the multi-dimensional or tensor data processing approach.\par

Various other applications of tensor decomposition for blind source separation are presented.
\cite{bousse2016tensor} first performs signal compression via low-rank tensor representation and then exploits the low-rank structure and uses a deterministic tensorization technique for blind source separation. Similarly, \cite{xie2018underdetermined} proposes a tensor decomposition to estimate the mixing matrix and NMF source model to estimate the source spectrogram factors for blind source separation. \cite{teschendorff2018tensorial} proposes a novel tensorial independent component analysis (tICA) algorithm for identifying biological sources of data variation. \cite{papalexakis2016tensors} describe the use of tensor decompositions for various data mining tasks such as Social and Collaboration Network Analysis, Web Mining, and Web Search, Knowledge Bases, Information Retrieval, and Topic Modeling, Brain Data Analysis, recommendation Systems, Urban Computing, healthcare, and Medical Applications and Speech and Image Processing and Computer Vision. Non-negative tensor and matrix factorizations are used in audio source separations. 
\cite{fevotte2018single,sprechmann2015supervised, ozerov2018introduction,leplat2020blind}   demonstrate the time-frequency/spectral decomposition with NMF for source separation for  unsupervised or supervised settings or both. The authors in \cite{innami2012nmf} propose a monaural sound source separation of environmental sounds via a two-phase clustering using non-negative matrix factorization (NMF).

\par
\section{Contributions of Dissertation}
The research presented in this dissertation demonstrates a novel fusion of a variety of artificial intelligence and deep learning techniques that greatly expand the usability of low quality thermal imagery to produce an innovative path planning and navigation system that can revolutionize the firefighting response system.

Firstly, we developed an algorithm capable of effectively performing object identification, segmentation, and tracking on thermal imagery with accuracies greater than 70\% and as high as 90\%. The development of a scene description system that describes the semantic information utilizing Natural Language Processing adds to this image recognition system’s utility by creating a system that can supplement firefighter vision in near-zero visibility conditions.
Second, we built a prototype hardware system capable of implementing our algorithm (processing, analyzing, and projecting the results) and relaying the processed information into a display viewable by firefighters in real-time via augmented reality.
Third, we created the foundation for a virtual guide capable of assessing actively changing environments and advising a firefighter on alternative paths utilizing Deep Q- Learning. 
Fourth, we implement tensor decomposition technology, which introduces the ability to detect rapidly changing environments or other anomalous activities within the scene. This addition to the system functions as a warning system, alerting firefighters to abnormal events such as rapidly increasing temperature or smoke levels in a room or rapidly decreasing oxygen levels. This knowledge may assist firefighters in accelerating evacuation of an area due to an impending catastrophic change in the local fire conditions.\par

This dissertation’s foundation is the development of the algorithms and models that effectively exploit the information gathered from the infrared camera. We use a trained deep Convolutional Neural Network (CNN) system to provide all necessary information to the firefighters to supplement their situational awareness gaps. This is achieved by identifying, classification, and tracking objects of interest (i.e., exits such as doors and windows). Crucial information (i.e., location and condition of victims) is identified and relayed back to firefighters to help their decision-making processes and safely navigate the environment. The deep Neural Net (DNN)-based algorithm we authored is sufficient to infer human recognition and posture detection to deduce a victim’s health level. This information can help prioritize rescue based on the immediacy of need and guide firefighters accordingly. We also employ deep-learning based systems path planning and navigation, path reconstruction, scene segmentation, estimation of firefighter condition, and Natural Language Processing to inform firefighters about the scene. This integrated image recognition system could dramatically reduce fatality through improved decision-making due to a better understanding of the ground circumstances. \par

The second piece of this research integrates FLIR One G2 and Intel RealSense D435i thermal and depth cameras with an NVIDIA  Jetson embedded Graphics Processing Unit (GPU) platform that deploys the developed machine and deep learning models to process captured datasets and then wirelessly stream the augmented images in real-time to a  secure local network router based on a mesh node communication topology. This communication topology allows for the transfer of the data to a viewer, in this prototype, a Microsoft HoloLens. This viewing system’s idea is that processed data could be projected onto a corner of the firefighter face shield, allowing them to view both the actual scene and what the CNNs have processed from the scene. In near-zero visibility, the augmented scene may show features that the firefighter could miss, such as a body or a doorway.\par

Disorientation within such circumstances as near-zero visibility is very common. In such a scenario, a virtual guide, immune to the pressures of fear, anxiety, and stress, could assist the disoriented firefighter in retracing their steps and making it out of the burning building. Such a virtual guide must understand the harsh realities of life or death consequences and learn from these experiences. We utilize Unreal Engine gaming software and Deep Q-Learning training techniques to train such a virtual guide to avoid life-threatening hazards that occur unpredictably within the chosen path. Through this training, the guide can respond instantaneously and re-path a route of escape. While this methodology is currently demonstrated in a simulation environment, it provides the foundation for transferring such learned information into a real-world fire scene navigation application. \par

The last piece of this research utilizes tensor decomposition for anomaly detection and blind source separation. We exploit the non-negative matrix and tensor decomposition techniques to explore the underlying processes for a given data set. Such methods enable us to explore the events and activities which are missed by the deep neural network algorithms to inform the firefighters about possible catastrophes. We take advantage of such frameworks’ compressed representation ability to instantly recognize the novel incidents such as rapidly changing temperature levels and heat distribution and then cautioning the firefighters from the further movement towards such regions. We also use the non-negative matrix factorization for spectral decomposition of the speech recordings to monitor the stress level and inform the commanders if the firefighter is no longer able to safely perform the task.

\section{Dissertation Summary}
\par
\begin{figure}[!ht]
    \centering
    \includegraphics[width=.8\textwidth]{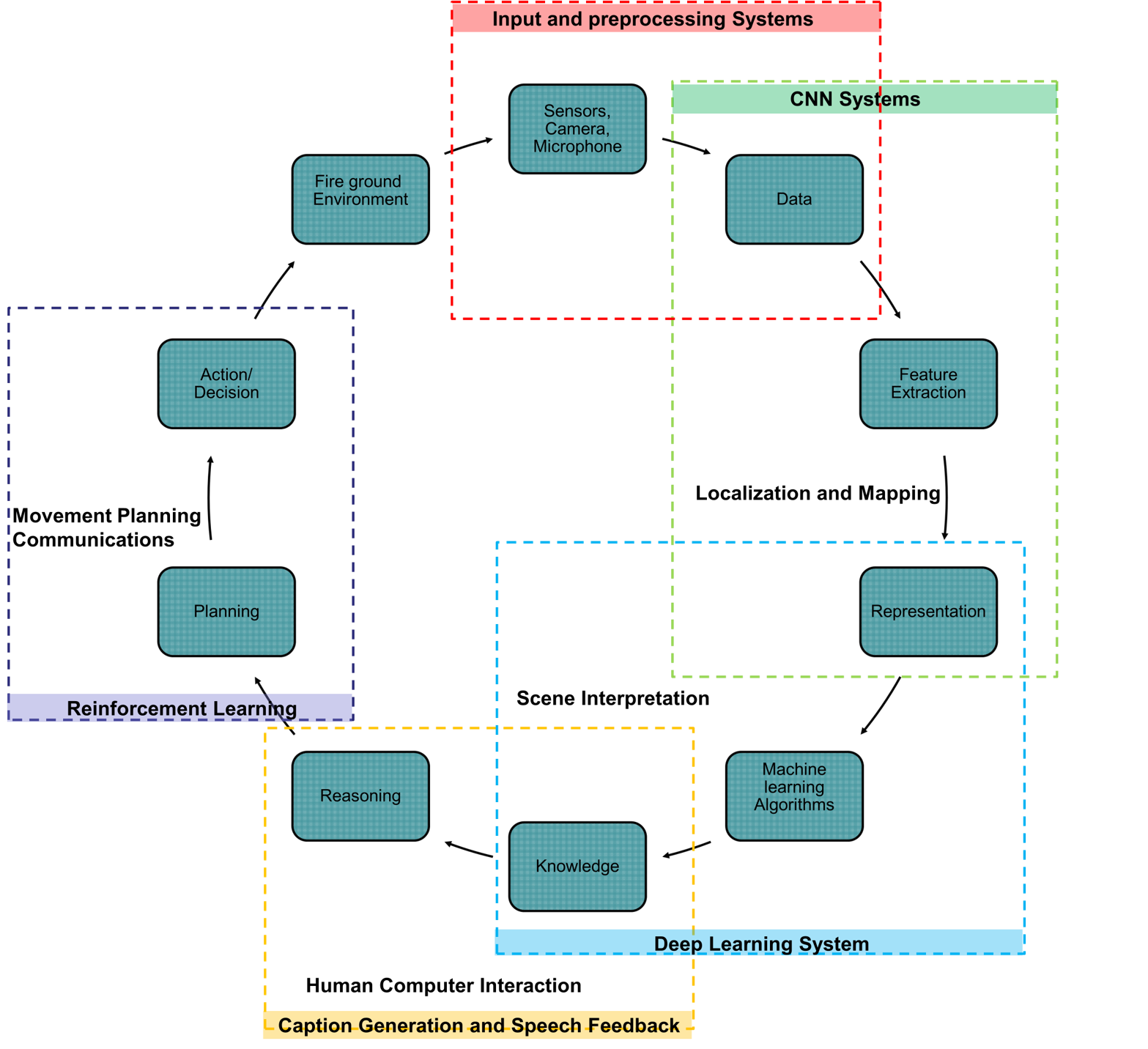}
    \caption{Subsystem interaction in the devised cyber-physical system}
    \label{fig:my_label}
\end{figure}
We propose a cyber-physical system to enable
firefighters with sensing capabilities for enhanced
awareness and decision making. State-of-the-art
computing technologies facilitate a remarkable opportunity for upgrading existing fire fighting protocols. The convergence of small, cheap wearable sensors paired with
wireless networks offer the tremendous potential to gather and share
critical fireground information in real-time. Meanwhile, advanced computing methodologies such as
machine learning (ML) algorithms make analyzing,
processing, modeling, and predicting utilizing multimodality data collected from different sensors possible, empowered by the latest mobile computing devices. Our system will enable the design of new first responder coordination
protocols by providing a predictive modeling capability that supports firefighters and the incident commanders to evaluate alternative actions and choose the best approach based on their experience and available
resources promptly. In summary, the intellectual merits of the proposed cyber-firefighter system
reside in technical innovations from the following five perspectives:

\begin{itemize}
    \item \textbf{Localization and Mapping.}  Smoke, lack of light, and high temperatures are aspects of the harsh environments endured by first responders entering buildings ablaze to rescue occupants. These conditions
 hamper visibility and can make navigation through the building in question nearly impossible without the
use of thermal imaging cameras which use heat bodies to their advantage, something RGB cameras are
incapable of. However, the lack of edges combined with poor contrast and limited dynamic range makes
the imagery obtained from thermal imaging cameras difficult to decipher or use to perform 3D reconstructions.
 This research aims to create an automated system that is capable of real-time, intelligent object detection and recognition and facilitates the improved situational awareness of firefighters during an emergency response. We have explored state-of-the-art machine/deep learning techniques to achieve this objective. This work aims to enhance the situational awareness of firefighters by effectively exploiting the infrared video that is actively recorded by firefighters on the scene. To accomplish this, we use a trained deep Convolutional Neural Network (CNN) system to classify and identify objects of interest from thermal imagery in real-time. During those critical circumstances created by a structure fire, this system can accurately inform firefighters’ decision-making process with up-to-date scene information by extracting, processing, and analyzing crucial information. Utilizing the new information produced by the framework, firefighters can make more informed inferences about their safe navigation circumstances through such hazardous and potentially catastrophic environments.

\item \textbf{ Scene Understanding.} Achieving a complete understanding of the surrounding scene is crucial to
safe and successful navigation in a fire environment. This portion of the project seeks to identify/detect
necessary background information followed by localization and tracking the object of interest. The
objects of interest to be identified include humans, doors, windows, fire, furniture, smoke, etc. The
system also seeks to prioritize the victims in the greatest need of assistance using human pose identification.
We have implemented a CNN based deep- learning system which can demonstrate \textgreater95\% detection
accuracy and excellent tracking accuracy on the stated objects of interest. We employ a Faster R-CNN backbone based Mask-RCNN for simultaneous object detection, tracking, and segmentation to inform firefighters about the position of the object of interest in reference to the position of the firefighter to avoid confusion leading to disorientation. 

\item \textbf{ Movement Planning.}  Movement planning is the most crucial outcome of the entire project. Fires
create actively changing environmental conditions. Within such an environment, the system needs to inform firefighters of the best possible movement options available to them while accounting
for these real-time changes. Live fire creates a dynamic, rapidly changing environment that claims many lives annually. Fighting these fires requires quick thinking, planning, and fearlessness. However, even with these aptitudes, the current lack of ability to pass information about rapidly changing conditions can prove fatal to even the most experienced firefighter. Thus, live-fire environments
present a worthy challenge for deep learning and artificial intelligence methodologies to help firefighters with scene comprehension maintain their situational awareness, tracking and relaying important features necessary for key decisions as they tackle these catastrophic events. To augment a firefighter’s understanding of their environment, 
We propose a deep Q-learning based agent who is immune to stress-induced disorientation and anxiety and can make clear navigation decisions based on the observed and stored facts in live-fire environments. As a proof of concept, we imitate structural fire in a gaming engine called Unreal Engine, enabling the agent’s interaction with the environment. The agent is trained with a deep Q-learning algorithm based on a set of rewards and penalties for its actions on the environment.  We exploit experience replay to accelerate the learning process and augment the agent’s learning with human-derived experiences. The agent trained under this deep Q- learning approach outperforms agents trained through alternative path planning systems and demonstrates this methodology as a good foundation for building a path planning navigation assistant capable of safely guiding firefighters through live-fire environments. Upon prolific experience of the virtual agent under such an environment, we aim to transfer such knowledge to a decision system that aids firefighters navigation in a fire environment by providing step by step directions that, by design, choose the safest path of navigation and avoid fire and obstructions. This
approach provides an efficient decision-making system that aids firefighters whose decision-making abilities may be impaired due to anxiety/stress levels and presents a novel approach to eliminating faulty
decisions made under duress.

\item \textbf{ Human-Computer Interaction.}  This research implements recent advancements in technology such as deep learning, point cloud, and thermal imaging, and augmented reality platforms to improve a firefighter’s situational awareness and scene navigation through an improved interpretation of that scene. We have designed and built a prototype embedded system that can leverage data streamed from cameras built into a firefighter’s personal protective equipment (PPE) to capture thermal, RGB color, and depth imagery and then deploy already developed deep learning models to analyze the input data in real-time. The embedded system analyzes and returns the processed images via wireless streaming, where they can be viewed remotely and relayed back to the firefighter using an augmented reality platform that visualizes the results of the analyzed inputs and draws the firefighter’s attention to objects of interest, such as doors and windows otherwise invisible through smoke and flames.  We also explored the feasibility of gauging firefighter’s stress levels generated from multi-modal data, which comprise analysis of breathing levels from microphone recording. This data will be fed to a Tensor decomposition system, which will learn from
the joint features from all of these. We will integrate our developed algorithm for robust non-invasive
respiration analysis in the presence of heavy noise, together with other non-invasive measures in these tasks.

\item \textbf{Communication.}  Our team has developed a prototype for robust communications that are intended to transmit data
in a changing and inconsistent environment, which is based on a mesh architecture constructed using a
Software-Defined Network. The network uses an algorithmic approach to determine the possible communication paths and resend those packets that did not reach the hub. It is endowed with a radio awareness
system that estimates the radio quality among nodes. The author’s contribution to this project is to provide the image dataset to benchmark the network’s performance and advise the students for the implementation.  We plan to extend the network’s capabilities
to make it adaptive by optimizing the computational burden/bandwidth of the system and introducing
sparse, distributed computation. The computational burden must be minimized to extend the batteries’ life, but it must be increased to lower the needed bandwidth of the data communication. The
present prototype is constructed using Raspberry Pi nodes that we will substitute with small GPU based
computers. The network’s endpoints will be endowed with a user interface to generate the situational
awareness available for the commander and the fire officers. The SOP protocols will be integrated into
them.
\end{itemize}

\section{Organization of Dissertation}
This dissertation is organized as follows. Chapter 2 presents a deep convolutional neural net for object recognition and classification from infrared images in a firefighting environment. This framework can identify essential objects of interest such as a person, fire, doors, windows, ladder, and estimating the human poses such as standing, crawling, and sitting to prioritize the rescue of individuals in a fire environment. \par 

To further improve the firefighters’ situational awareness in chapter 3, we demonstrate a Mask R-CNN-based framework capable of performing object detection, tracking, and segmentation simultaneously from infrared images. \par 

Chapter 4 shows a caption generation framework combining the CNN model developed in chapter 2 with an NLP framework for describing the scene. The framework combines visual features extracted from the CNN framework and textual features extracted with a Long Term Short Term Memory (LSTM) to predict a sequence of words for making a meaningful description of the scene.\par 

Chapter 5 demonstrates a prototype hardware implementation of the proposed frameworks from chapters 2,3 and 4 into an embedded platform to process the visual information locally and relay the processed data to the augmented reality worn by firefighters to aid them in navigation. 
\par
Chapter 6 proposes and demonstrates a virtual path planning agent based on deep Q-learning on a gaming engine called Unreal Engine. The simulation environment depicts smoke and fire and enables a virtual agent to learn the navigation through a dynamic setting filled with fire using rewards and penalties. This learned experience can then be transferred to an AI system to aid firefighters in navigation in a real-world fire scenario. 

\par
Chapter 7 introduces tensor decompositions for infrared images to extract hidden representations for blind source separation and anomaly detection. This technique is beneficial for data compression for storing redundant filtering information for post-processing to save power and memory. \par 

Finally, we conclude our works and provide future research directions in chapter 8. 
\section{ Publications Resulting from the Dissertation}

A list of our publications related to this dissertation is as follows:
\begin{itemize}
    \item Bhattarai, Manish, and Manel MartíNez-Ramón. “A Deep Learning Framework for Detection of Targets in Thermal Images to Improve Firefighting.” IEEE Access 8 (2020): 88308-88321.
    \item Bhattarai, Manish, Aura Rose Jensen-Curtis, and Manel Martínez-Ramón. “An embedded deep learning system for augmented reality in firefighting applications.” IEEE International Conference on Machine Learning and Applications(ICMLA), Florida, USA, 2020.
    \item Bhattarai, Manish, and Manel Martínez-Ramón. “A deep Q-learning based  path planning and navigation system for firefighting environments.” International Conference on Agents and Artificial Intelligence(ICAART), Virtual,2021.
    \item Bhattarai, Manish, and Manel Martínez-Ramón. “Tensor decomposition for firefighting environments.” (Under Submission)
\end{itemize}

\chapter{Object Recognition and Classification in Firefighting Environments}
\section{Motivation}

Current firefighting modalities do not involve any automated detection mechanism, and the target is identified solely by the firefighter. Detection processes can be adversely affected by environmental factors inherent in active fire scenes. High temperatures, near-zero visibility caused by debris, smoke, lack of lighting, and a rapidly changing environment can combine to disorient and further inhibit decision-making processes, affecting even experienced firefighters. Under these hazardous conditions, lives can be lost due to rescue operation decisions based on an incomplete or inaccurate understanding of the most current environmental conditions within the structure. Federal Emergency Management Agency studies \footnote{\url{https://www.usfa.fema.gov/downloads/pdf/publications/ff\_fat17.pdf}}
show a  majority of firefighter mortalities reported resulted from inefficient decision-making protocol. Heightened anxiety levels, leading to misinterpretation of the scene, and lack of a complete understanding of the environment are cited as factors.  We propose an Artificial Neural Network-based system capable of autonomously identifying objects and humans in the scene of the event in real-time to improve on-the-ground knowledge that dictates decision-making protocol. The artificial intelligence (AI) based results can be used to assist in reducing these mortality statistics by minimizing anxiety-induced errors. The AI-based system is also capable of accurately differentiating between human postures. This pose detection can help firefighters prioritize rescue of identified victims by estimating their health condition based on their posture. 

This chapter demonstrates a CNN-based autonomous system capable of generating information that can improve situational awareness for firefighters regarding the environment into which they are deployed.  The information is generated by classifying fire, objects of interest like doors, windows, and people, and other thermal conditions using infrared video that is actively recorded by firefighters on the scene. The CNN system can detect and classify desired targets and relay the information back to firefighters, thereby providing crucial information necessary for informing essential planning decisions. This enables the firefighters and their commanders' access to data collected and processed through an unbiased lens that is comprehensive and reliable. The improved knowledge of local events and changes across the scene allows leadership to thoroughly assess the local critical conditions and make appropriate decisions based on real-time conditions. The improvement in situational awareness provided by the reliable stream of information deduced by the CNN could also assist disoriented firefighters in choosing a safer path in a fire environment by autonomously identifying and alerting firefighters to the presence of objects of interest such as doors, windows, human targets, excessive smoke, etc. that may have been overlooked in their confusion.
\par
Despite the large quantity of work related to the processing of infrared images in fireground or related scenarios, to our knowledge, no work has been published that attempts to construct a system that integrates online detection of targets of interest in these scenarios, including humans, objects, body posture or the presence of fire from a large number of images recorded in real fire training situations by firefighters. Thus, there is a need for effective automatic target detection generated in real-time in a firefighting environment and the associated need for a highly accurate classifier. Our research seeks to address these needs. To do so, we have adapted and enhanced an existing state-of-the-art CNN based automatic classifier system to improve its efficacy in identifying and classifying humans and objects of interest in real-time in a firefighting environment. Also, we have improved upon the creation process of a data set so that it may be used to train the neural network (NN) system effectively. We have also trained the system to detect objects and humans simultaneously. Prior technique capabilities were limited to one or the other. To further assist rescue operations, we also added a pose detection element in which CNN further distinguishes whether a person is in a sitting, crawling, or upright position. This detection series intends to allow rough estimates of the victim's health condition or panic state to be approximated and used in the prioritization of rescue operations.
\par
Thermal imaging cameras are the most widely used cameras to obtain footage and circumvent visibility issues within a fireground. Albuquerque Fire Department(AFD) and the Santa Fe Fire Department keep at least one camera in a truck at all times. The National Fire Protection Association also supports their usage over RGB-based cameras, which are dependant on light sources for the clear depiction of imagery. In active fire scenes, where electricity may be out, and no natural lighting is available, the thermal signature picked up by infrared cameras becomes crucial. Smoke and debris further reduce the clarity of view for an RGB-camera, but thermal cameras can cut through both if a heat source lies beyond the walls of smoke and dust. Compared to cameras operating in the visual spectrum, thermal cameras are advantageous due to their ability to see in total darkness, robustness to illumination variations, and less intrusion on privacy. The characteristics of thermal infrared radiation and imagery pose specific challenges to image analysis algorithms. Unlike visual object tracking, thermal infrared object tracking can track a target object in total darkness. While thermal infrared cameras promise improved bad weather and nighttime robustness compared with standard RGB-cameras, detecting objects, such as persons, in thermal infrared imagery is challenging because image resolution and quality are typically far lower, especially for low-cost sensors. 
\par
However, thermal imaging cameras capable of surviving the temperatures found in buildings ablaze are costly and are not standard issue for every firefighter to carry. Instead, one per fire crew is deployed. Thus, creating a system that can maximize the thermal imagery available, process it, and return accurate, real-time information to every firefighter on the scene would dramatically enhance the responders' combined situational awareness as a unit. Our research is restricted here to thermal imaging, but it can be extended to RGB or UV imaging.

\begin{figure*}
    \centering
    \includegraphics[scale=0.4]{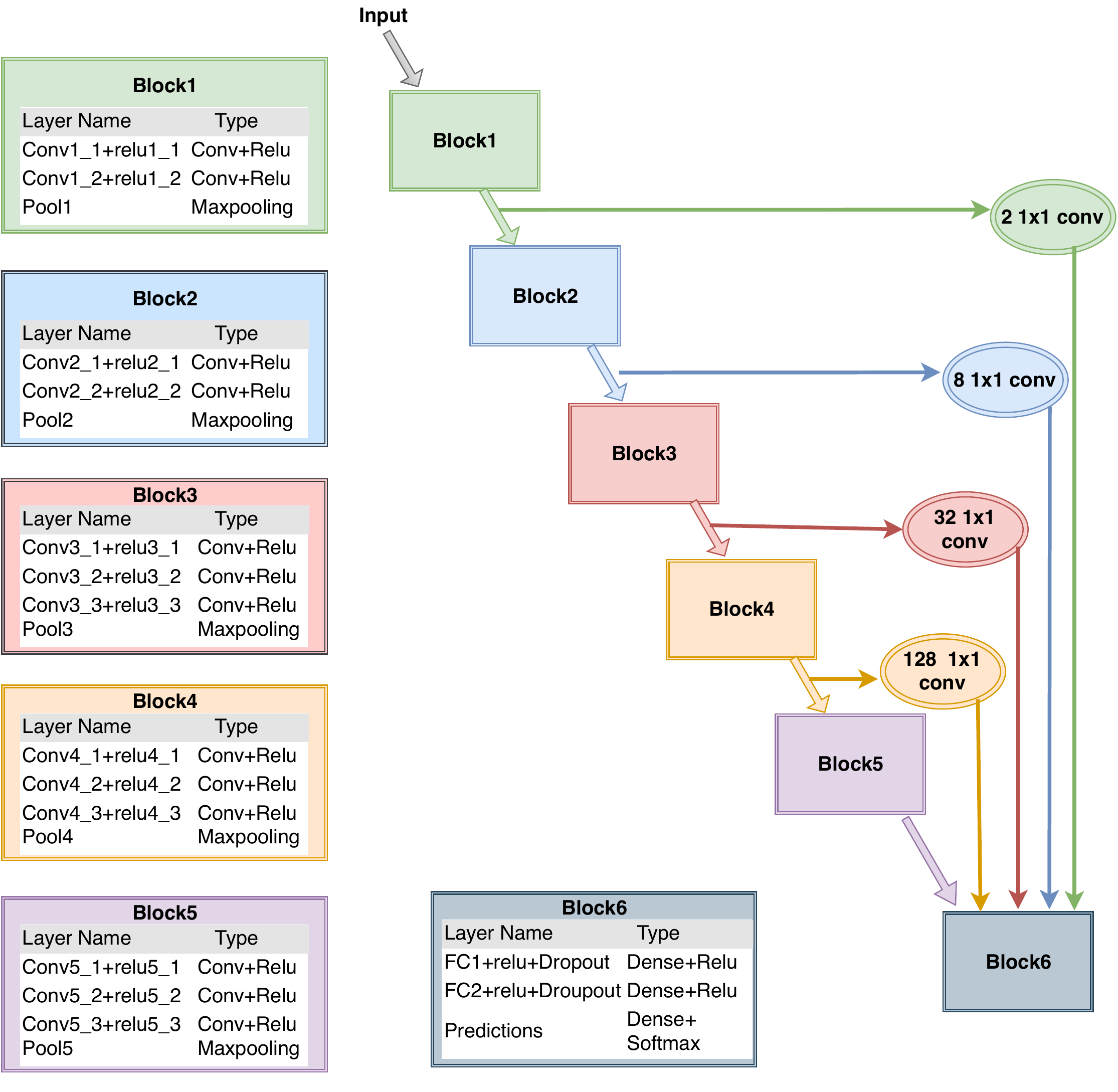} 
    \caption{Block diagram of VGG16 at different depth levels.}\label{fig:1and2}
\end{figure*}

\begin{figure*}
    \centering
    \includegraphics[scale=0.4]{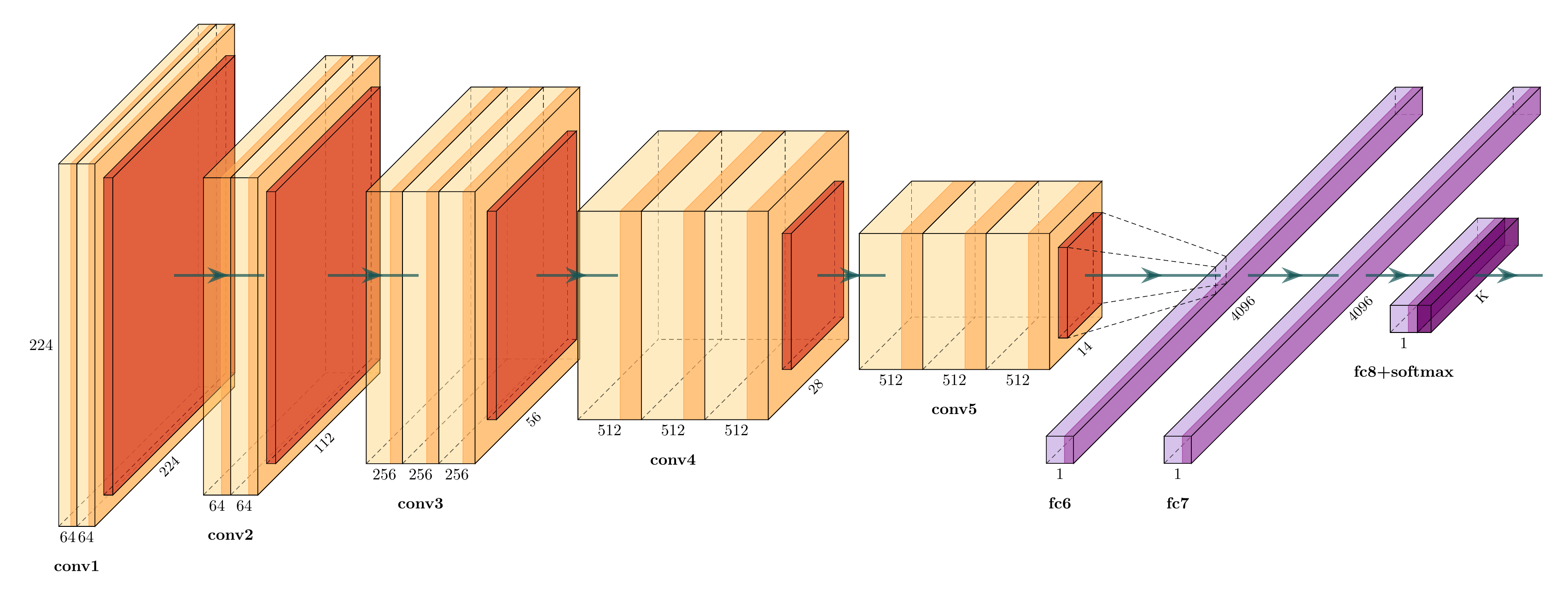} 
    \caption{Architechture of depth-5 VGG16.}\label{fig:archi}
\end{figure*}

\section{System Description}

The convolutional neural network presented is based on the structure of the VGG16 neural network presented in \cite{simonyan2014very}. Several different depths of the neural network have been tested, ranging from 1 to 5 convolutional sections, as shown in figure \ref{fig:1and2}, all of them followed by a fully connected section of three layers. For the depth-1 configuration, CNN's have an input of dimension $224\times 224$. The corresponding IR images are scaled to this size. They are convolved by $3\times3$ filters to produce 64 channels of dimension $224\times 224$. The resulting outputs are then passed through a set of ReLU activations. The process is repeated with an identical convolution, ReLU, and then pooled to produce 64 channels of dimension $112 \times 112$ pixels. Next, the output is passed through a $1 \times 1$ convolution to create two channels of $112\times 112$ pixels. This is then processed through a fully connected layer with 4096 outputs, ReLU activation and dropout, a second identical fully connected layer, and a layer with five outputs and softmax activation that gives the classification scores across five different classes. 

For the depth-2 model, the first convolution layer is identical to the previous model. After the first pool, two more convolutions are added that produce 128 channels of size $112\times 112$, reduced to 128 channels of dimension $56\times 56$. The subsequent layers have identical structure as in the one layer model, where the input to the first fully connected layer has eight channels of dimension $56\times56$. 
The models with 3, 4, and 5 layers are constructed using the same methodology. The architecture for depth-5 is shown in figure \ref{fig:archi}.

The networks have been trained and tested in three different modalities. The first one covers object classification, including people, ladders, windows, doors, and a combination of windows and firefighters (5 classes). The second modality comprises a posture-based classification, including standing, sitting, and crawling (3 classes). The third modality is binary and includes the presence or absence of fire. For the modalities with 5 and 3 classes, the training was performed using a categorical cross-entropy loss. For the modality with two classes, we used binary cross-entropy loss combined with stochastic gradient descent. The learning rate for all training was $10^{-4}$ with a decay of $0.009$. Early stopping with cross-validation was applied to avoid overfitting. 

To avoid overfitting between training and test datasets, images taken from different recordings have been used in both processes. The results shown below contain extensive tests using a database of recordings taken by this work paper's researchers. It is described below.


\section{Training and Test Datasets}

The imagery data set used in this project was recorded at the Santa Fe Firefighting Facility, located in Santa Fe, New Mexico.  Extensive video footage was acquired using an IR MSA 5200HD2TIC Camera. This camera is a multipurpose firefighting tool designed to aid search and rescue efforts in structural firefighting environments. It uses an uncooled microbolometer vanadium oxide(Vox) detector, which comprises 320x240 FPA with the pitch of 38$\mu m$ and a spatial resolution of 7.5 13.5$\mu m$. This resolution is sufficient to capture the necessary features for target detection. It records the image with a 320x240 focal plane array sensor and can record imagery in two different modes, i.e., low and high sensitivity. This device also features high score imagery, generating 76,000 pixels of image detail in low and high sensitivity modes. The dense spectral resolution is (7.5 to 13.5$\mu m$). The output video is in NTSC format with a frame rate of 30 frames per second. The scene temperature has a maximum operating range of 560 degrees Celsius or 1040 degrees Fahrenheit.  

Over 6 hours of recorded video in both open and closed environments was acquired. The recording sessions produced more than  150 infrared video files, each one lasting approximately 2 to 3 minutes. All videos contained some combination of sequences involving the desired targets to be detected.  In some scenarios, single objects of interest are present in a scene, while multiple objects of interest are present simultaneously. This variation requires the CNN image classification system to be capable of multiple simultaneous object detections in the same frame. Other objects or postures of interest outside of what we chose can also be detected if those objects or poses occur in sufficient frequency to train the data.

The objects of interest focused on this research are humans, doors, windows, ladders, and fire. The objective of the above-described structure is to detect all objects of interest present in the scene simultaneously.  
\begin{figure}
    \centering    
    \includegraphics[scale=0.05]{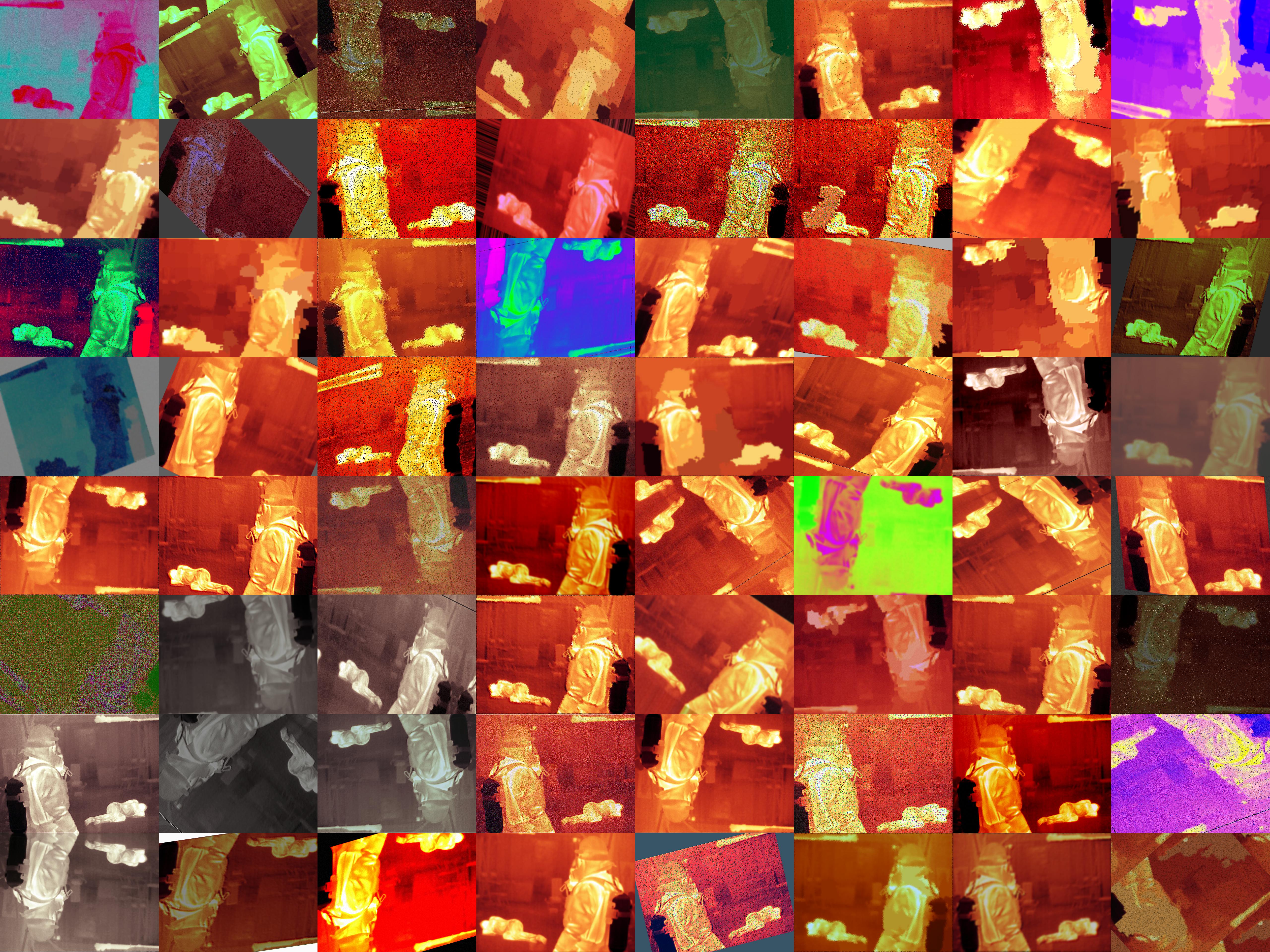} 
    \caption{Demonstration of Image augmentation for the training set.}\label{fig:augment}
\end{figure}
Since data sets of sufficient detail are needed to train the neural network accurately, the videos were used to extract many training and test purposes. The training and test sequences were extracted from different videos to avoid overfitting. The video extracted from the camera was produced in grayscale 8-bit format. To generate the training data set, the images were pre-processed with data augmentation techniques such as skewing, translation, zooming, cropping, and rotation, as shown in figure \ref{fig:augment} (False color for better visualization). 

Tables \ref{tab:data_objects}, \ref{tab:data_fire} and \ref{tab:data_poses} show the total number of images acquired for training and test before the augmentation procedure. Objects of interest contained in images used in the training data set were hand-labeled to assign them to a class. The labeled classes were then grouped into three sets (objects, fire, human poses). To compensate for data asymmetry within the different classes, data augmentation was performed on classes that had lower representation within the original dataset. For example, 7950 images from the original dataset were labeled and added to the "No fire" class within the set "Fire". The "fire class" was augmented to add 347 images to bring the total number of "fire" labeled images available for training up to 7950. 

The primary training was performed using an Alienware Aurora R6 Desktop computer configured with 32GB RAM and a Dual GTX 1080 with 16GB GPU memory.  The cross-validation and hyperparameter tuning portion of the research were performed on the high-performance computer, Xena housed at the UNM Center for Advanced Research Computing(CARC). The machine has 24 single GPU nodes and four dual GPU nodes. These dual GPU nodes have 2 NVIDIA Tesla K40m GPUs with GPU memory of 11GB each and 64GB RAM per node. The computations for cross-validation and hyperparameter tuning utilized the four dual GPU nodes.

\begin{table}
\centering
\begin{subtable}[h]{.3\linewidth}
\centering
\begin{tabular}{|c|c|} 
\hline
Object of & number \\
Interest  & \\
\hline
door & 322 \\
firefighter \& & 4663 \\
window &  \\
firefighter & 15484 \\
ladder & 1589 \\
window & 1620 \\
\hline
\end{tabular}
\caption{}\label{tab:data_objects}
\end{subtable}%
\begin{subtable}[h]{.3\linewidth}
\centering
\begin{tabular}{|c|c|} 
\hline
Object of & number \\
Interest  & \\
\hline
fire & 7603  \\
no fire &7950 \\
\hline
\end{tabular}
\caption{}\label{tab:data_fire}
\end{subtable}%
\begin{subtable}[h]{.3\linewidth}
\centering
\begin{tabular}{|c|c|} 
\hline
Object of & number \\
Interest  & \\
\hline
crawling & 8678 \\
sitting & 1803 \\
standing & 9928 \\
\hline
\end{tabular}
\caption{}\label{tab:data_poses}
\end{subtable}%
\caption{(a) Data quantification for objects classification task(object set), (b) Data quantification for fire classification task(fire set), and (c) Data quantification for poses classification task(poses set)}
\end{table}

The test data consists of 1/10 of the data set. The rest of the data has been used for training and validation purposes. A validation has been performed with a 9-stratified fold procedure with the remaining 9/10ths of data.

\begin{figure}[!ht]
\begin{center}
\includegraphics[scale=0.4]{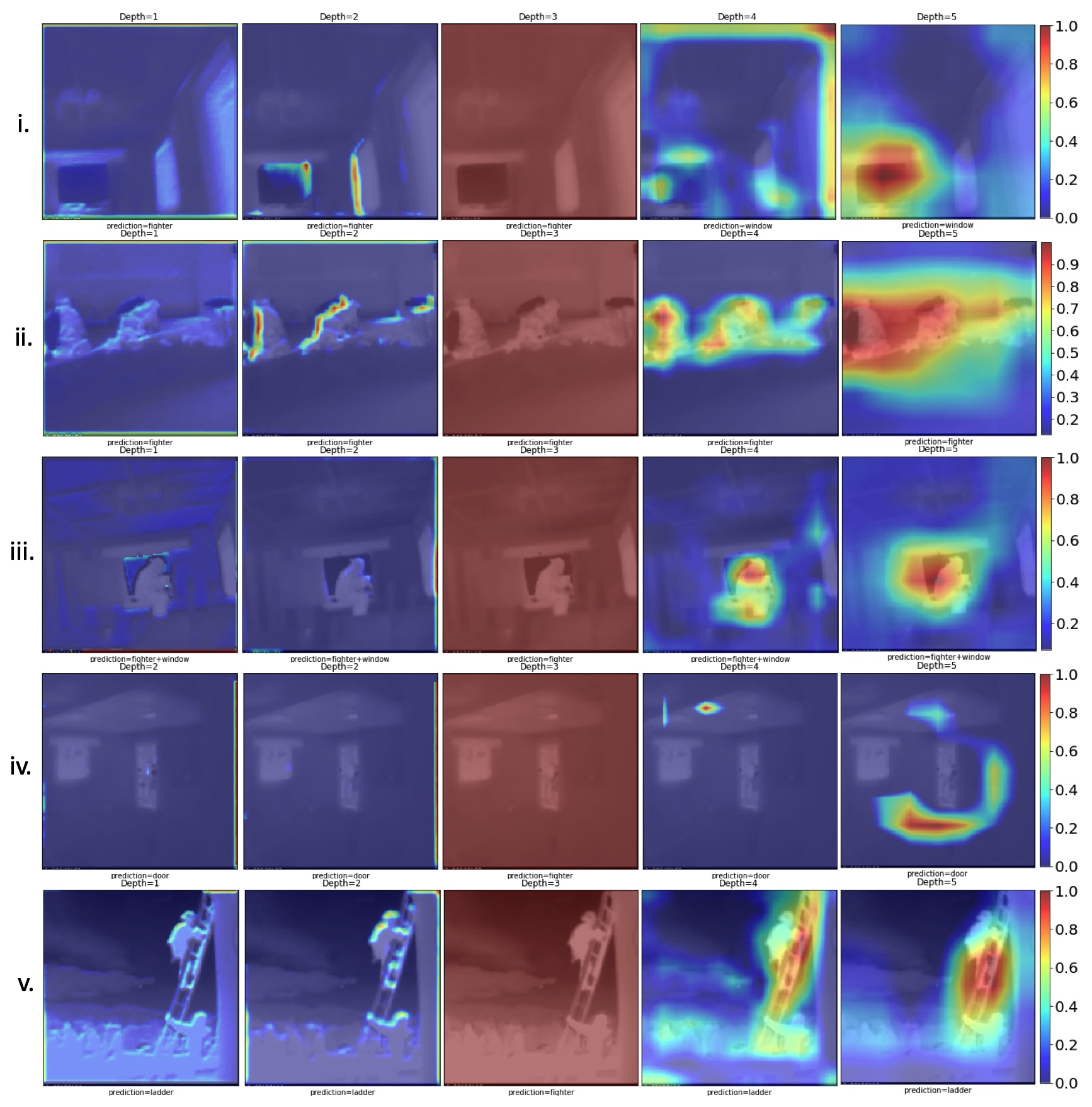}
\caption{Visualization of the features obtained at the last Convolutional layer for Object Classification for different depths for respective classes \textbf{i.} Window, \textbf{ii.} Fighter, \textbf{iii.} Fighter+Window,  \textbf{iv.} Door and  \textbf{v.} Ladder }\label{fig:features_objects}
\end{center}
\end{figure}

\begin{figure}[!ht]
\begin{center}
\includegraphics[scale=0.27]{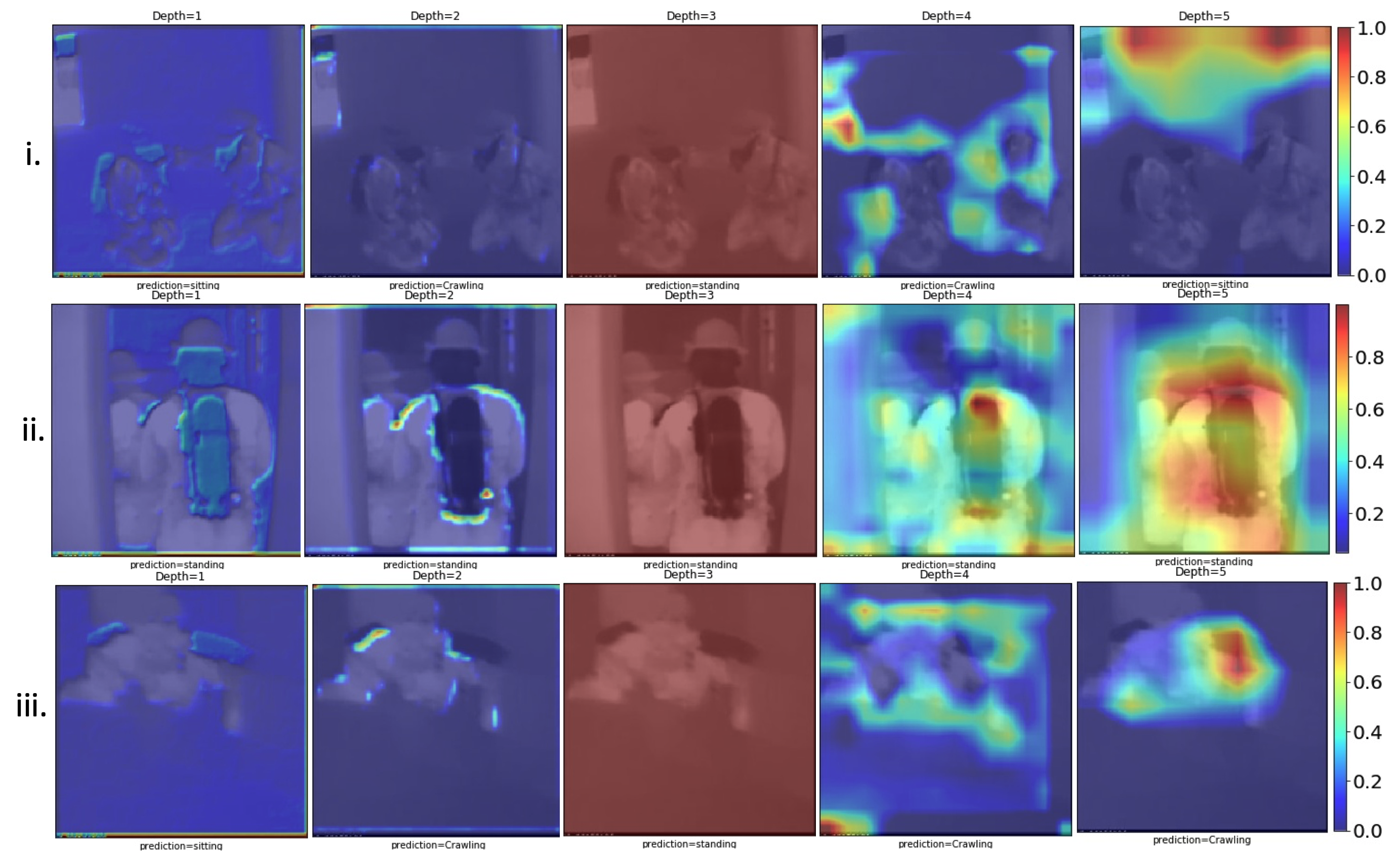}
\caption{Visualization of the features obtained at the last Convolutional layer for pose Classification for different depths for respective classes \textbf{i.} Sitting, \textbf{ii.} Standing and \textbf{iii.} Crawling} \label{fig:features_poses}
\end{center}
\end{figure}

\begin{figure}[!ht]
\begin{center}
\includegraphics[scale=0.27]{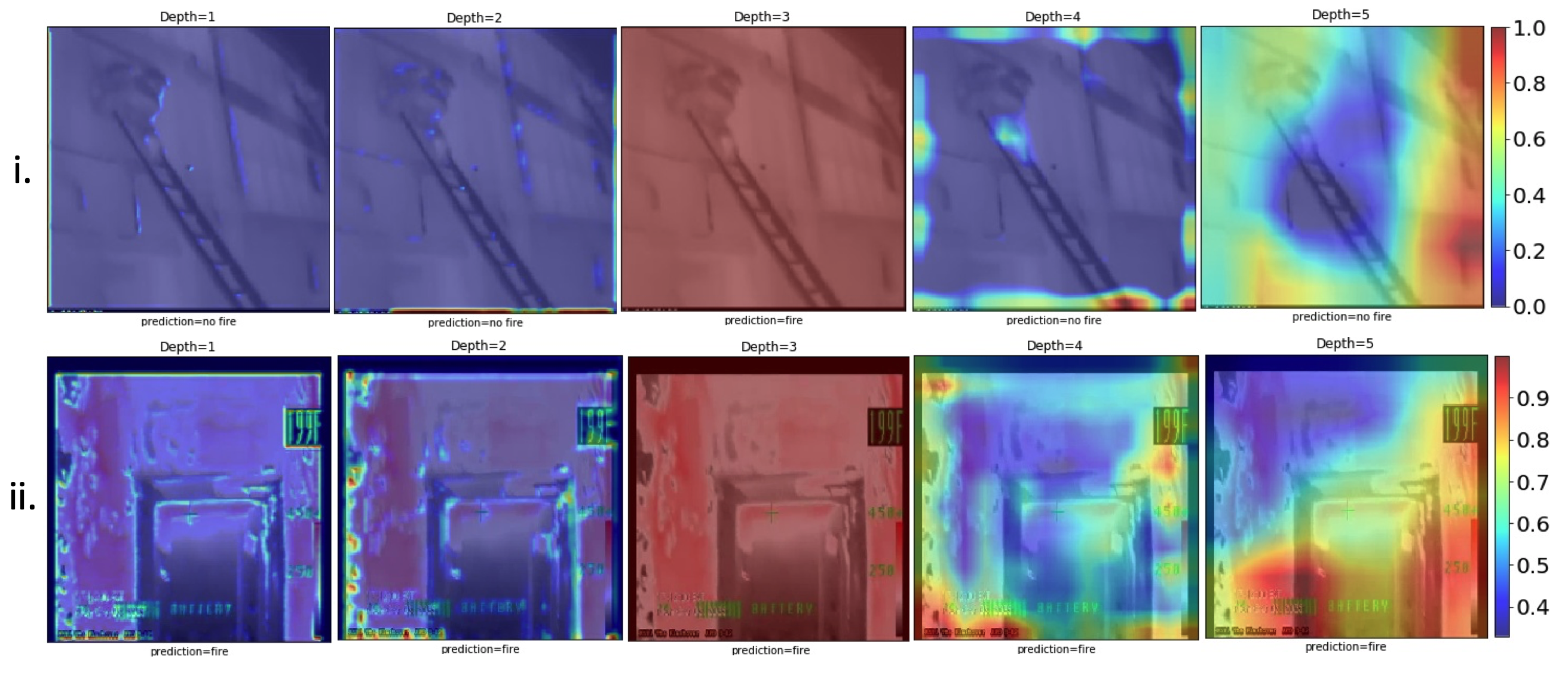}
\caption{Visualization of the features obtained at the last Convolutional layer for fire Classification for different depths for respective classes \textbf{i.} fire and  \textbf{ii.} No Fire} \label{fig:features_fire}
\end{center}
\end{figure}

\begin{figure}[!ht]
\begin{center}
\includegraphics[scale=0.1]{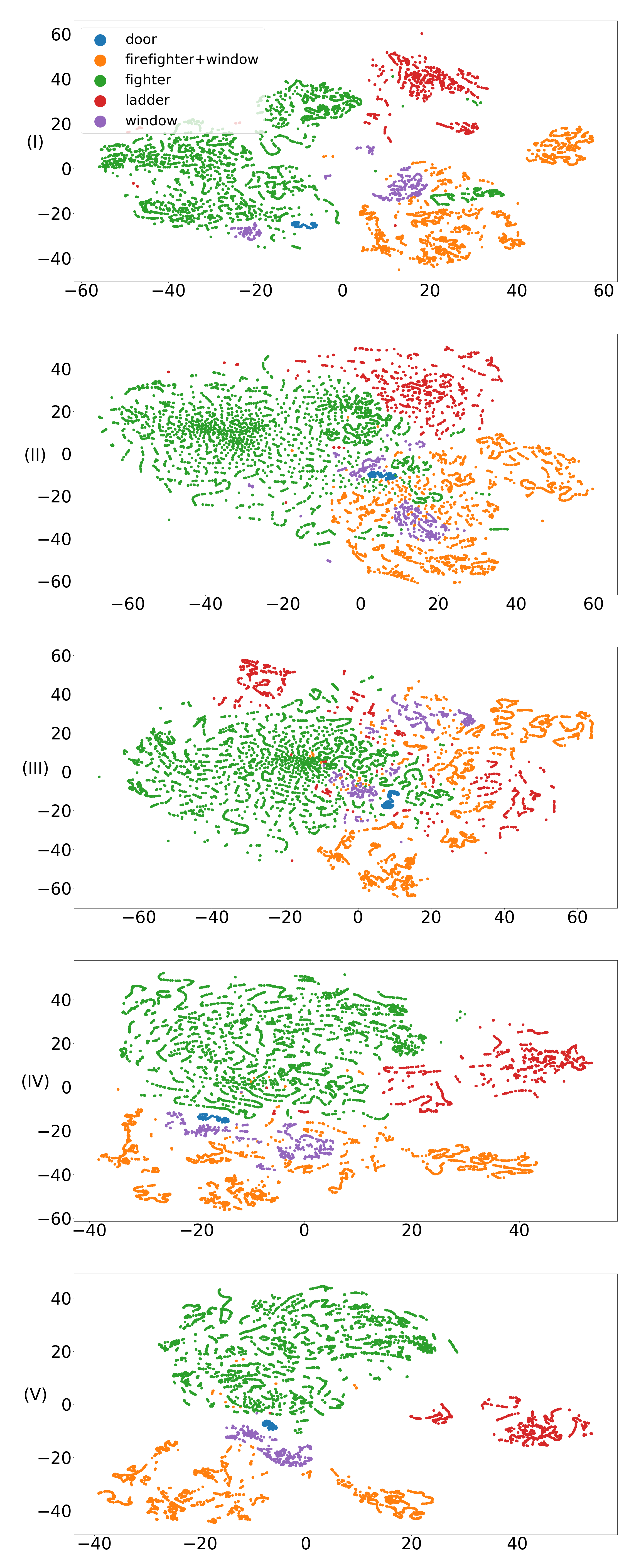}
\caption{t-SNE of the CNN when trained to detect objects. The number of convolutional layers was respectively I) depth-1, II) depth-2, III) depth-3, IV) depth-4 and V) depth-5.}
\label{fig:TSNE}
\end{center}
\end{figure}

\section{Results}

Our network is trained to detect objects, including ladders, doors, windows, people, and fire. The network is also trained to detect and classify different body positions of the detected people, civilian and firefighter alike. We classify three different poses corresponding to standing, sitting, and crawling (prone), which cover the videos' variation. Other important positions can be detected if they are sufficiently available in the image dataset used for training.  The following results present the visualization of the features extracted by the convolutional section of the network, the F1 scores, and achieved an accuracy of the network, confusion matrices, and ROC curves to estimate the false alarm versus the detection probability of the network. For this experiment, we change the detection probability by sweeping the detection threshold of the network.

\subsection{Visualization of the Extracted Features}
All the configurations were tested against the available data. We have used grad-CAM(gradient weighted Class Activation Maps) \cite{selvaraju2017grad} for visualizing attention over input. Grad-CAM uses the  Convolutional layer before the dense layer to utilize spatial information for displaying the saliency maps corresponding to each predicted class. 
The activations correspond to the class with the highest detection score. We can see the classification scores for different classes corresponding to different images at different depths in tables \ref{tab:PredictionObjects}, \ref{tab:PredictionPoses} and \ref{tab:PredictionFire}. Based on the table's classification score, we visualize the activations corresponding to classes with the highest score.  Figures \ref{fig:features_objects}, \ref{fig:features_poses} and \ref{fig:features_fire} show grad-CAM for different depths for each classification task. Each of these figures is comprised of subfigures corresponding to each class depicting the grad-CAM for different depths from left to right. For most of these maps, depth-1 and depth-2 show that activations come from either edges or small regions from the objects of interest. With depths equal to 3, the network cannot produce feature extraction, showing that this configuration is not useful for classification. Nevertheless, at higher abstraction levels (depths equal to 4 and 5), features are extracted that produce exemplary results.

Figure \ref{fig:TSNE} shows the visualization of the output of the neural networks using the t-distributed stochastic neighbor embedding technique(tSNE) presented in \cite{maaten2008visualizing}. This technique allows the user to obtain a low-dimensional representation of the data to understand its distribution and separability better. Figure \ref{fig:TSNE}(I) shows the distribution of the feature space before the network's output layer when the depth is 1. In this configuration, the separability of the data is intuitively fair. Figures \ref{fig:TSNE}(II) and (III) show the distribution for depth 2 and 3. These representations show a high level of overlapping of the different classes, which is consistent with the poor extraction of features, as shown in figures \ref{fig:features_objects}, \ref{fig:features_poses} and \ref{fig:features_fire}. The representation for a depth of 4 and 5, as shown in figures \ref{fig:TSNE} (IV) and (V), is highly improved, as the overlapping is dramatically decreased compared to the 2 and 3 depth networks. 

The green dots correspond to the images of firefighters. Orange dots correspond to images containing firefighters and windows. These two classes comprise most of the data set. Still, they show a low overlap between them, indicating high accuracy in the classifier's ability to distinguish between firefighters in the presence or absence of a window. The blue dots correspond to images with doors. Door shapes vary due to the angle of the camera. However, the classifier can extract all door features regardless of these differences in perspective-induced shape, and all features appear clustered in a small area. At lower depths, however, they are highly overlapped with the images containing firefighters. The overlap decreases with a higher level of abstraction (4 and 5 depths). The violet spots represent windows, which appear to be overlapping with the images of firefighters and windows. The accuracy and classification rates are in high agreement with this visualization.

\begin{table}[!ht]
\centering
\caption{Prediction score for object classification for different depths (green and yellow cells are the top 2 prediction scores)}
\label{tab:PredictionObjects}
\begin{tabular}{|l|l|r|r|r|r|r|}
\hline
\cellcolor[HTML]{C0C0C0} Figure & \cellcolor[HTML]{C0C0C0} Depth & \cellcolor[HTML]{C0C0C0} Door & \cellcolor[HTML]{C0C0C0} F/W & \cellcolor[HTML]{C0C0C0} Ladder & \cellcolor[HTML]{C0C0C0} Window & \cellcolor[HTML]{C0C0C0} Fighter \\ \hline
\ref{fig:features_objects} i.    &1&    0.003&    0.053&    0&    \cellcolor{yellow!50}0.408     & \cellcolor{green!25}0.536  \\ \hline
    &2&    0&    \cellcolor {yellow!50}0.43&    0&    0.01    & \cellcolor{green!25}0.56  \\ \hline
    &3&    0&    0&    0&    0&     \cellcolor{green!25}1  \\ \hline
    &4&    0.002&    \cellcolor {yellow!50}0.037&    0&     \cellcolor{green!25}0.961&    0  \\ \hline
    &5&    0.001&    \cellcolor {yellow!50}0.004&    0&     \cellcolor{green!25}0.996&    0  \\ \hline \hline
\ref{fig:features_objects} ii.    &1& \cellcolor {yellow!50}0.006&    0&    0.001&    0.001&     \cellcolor{green!25}0.992  \\ \hline
    &2&    0&    0&    0&    0&     \cellcolor{green!25}1  \\ \hline
    &3&    0&    0&    0&    0&     \cellcolor{green!25}1  \\ \hline
    &4&    0&    0&    0&    0&     \cellcolor{green!25}1  \\ \hline
    &5&    0&    0&    \cellcolor {yellow!50}0.001&    0&     \cellcolor{green!25}0.9999  \\ \hline \hline
    
\ref{fig:features_objects} iii.    &1&    0&     \cellcolor{green!25}0.997&    0&    \cellcolor {yellow!50}0.003&    0  \\ \hline
    &2&    0&     \cellcolor{green!25}1&    0&    0&    0  \\ \hline
    &3&    0&    0&    0&    0&     \cellcolor{green!25}1  \\ \hline
    &4&    0&     \cellcolor{green!25}0.974&    0&    \cellcolor {yellow!50}0.025&    0  \\ \hline
    &5&    0&     \cellcolor{green!25}0.999&    0&    \cellcolor {yellow!50}0.001&    0  \\ \hline \hline

\ref{fig:features_objects} iv.    &1&     \cellcolor{green!25}0.988&    0.001&    0&    0.001&    \cellcolor {yellow!50}0.01  \\ \hline
    &2&     \cellcolor{green!25}0.996&    0&    0&    0.001&    \cellcolor {yellow!50}0.003  \\ \hline
    &3&    0&    0&    0&    0&     \cellcolor{green!25}1  \\ \hline
    &4&     \cellcolor{green!25}0.95&    \cellcolor {yellow!50}0.029&    0&    0.02&    0.001  \\ \hline
    &5&     \cellcolor{green!25}0.805&    0.01&    0&    \cellcolor {yellow!50}0.181&    0.004  \\ \hline \hline

\ref{fig:features_objects} v.    &1&    0.007&    0.001&     \cellcolor{green!25}0.976&    0.001&    \cellcolor {yellow!50}0.015  \\ \hline
    &2&    0&    0&     \cellcolor{green!25}0.999&    0&    \cellcolor {yellow!50}0.001  \\ \hline
    &3&    0&    0&    0&    0&     \cellcolor{green!25}1  \\ \hline
    &4&    0&    0&     \cellcolor{green!25}0.998&    0&    \cellcolor {yellow!50}0.002  \\ \hline
    &5&    0&    0&    \cellcolor{green!25}0.9&    0&     \cellcolor{yellow!50}0.1  \\ \hline 

\end{tabular}
\end{table}

\begin{table}[!ht]
\centering
\caption{Prediction score for Pose classification for different depths}
\label{tab:PredictionPoses}

\begin{tabular}{|l|l|r|r|r|}
\hline
\cellcolor[HTML]{C0C0C0} Figure & \cellcolor[HTML]{C0C0C0} Depth & \cellcolor[HTML]{C0C0C0} Crawling & \cellcolor[HTML]{C0C0C0}Standing & \cellcolor[HTML]{C0C0C0} Sitting  \\ \hline
\ref{fig:features_poses} i.&    1&    \cellcolor {yellow!50}0.33& 0&\cellcolor{green!25}    0.67  \\ \hline
    &2&     \cellcolor{green!25}0.83&    0.001&\cellcolor {yellow!50}    0.169  \\ \hline
    &3&    0&     \cellcolor{green!25}1&    0  \\ \hline
    &4&    \cellcolor{green!25}0.63&    0&\cellcolor {yellow!50}    0.37 \\ \hline
    &5&\cellcolor {yellow!50}0.20& 0&\cellcolor{green!25}    0.80  \\  \hline \hline
\ref{fig:features_poses} ii.&    1&    0& \cellcolor{green!25}    1&    0  \\ \hline
    &2&    \cellcolor {yellow!50}0.01& \cellcolor{green!25}    0.99&    0.001  \\ \hline
    &3&    0& \cellcolor{green!25}    1&    0  \\ \hline
    &4&    0& \cellcolor{green!25}    1&    0  \\ \hline
    &5&    0& \cellcolor{green!25}    1&    0  \\ \hline \hline
\ref{fig:features_poses} iii.    &1& \cellcolor{yellow!50}0.33&    0.02& \cellcolor{green!25}    0.65   \\ \hline
    &2&     \cellcolor{green!25}0.536&    \cellcolor {yellow!50}0.311&    0.153  \\ \hline
    &3&    0& \cellcolor{green!25}    1&    0  \\ \hline
    &4&     \cellcolor{green!25}0.62&    0.097&\cellcolor {yellow!50}    0.283   \\ \hline
    &5&     \cellcolor{green!25}.8&    0&\cellcolor {yellow!50}    0.2   \\ \hline 
\end{tabular}
\end{table}

\begin{table}[!ht]
\centering
\caption{Prediction score for Fire classification for different depths}
\label{tab:PredictionFire}

\begin{tabular}{|l|l|r|r|}
\hline
\cellcolor[HTML]{C0C0C0} Figure & \cellcolor[HTML]{C0C0C0} Depth & \cellcolor[HTML]{C0C0C0} Fire & \cellcolor[HTML]{C0C0C0} No Fire  \\ \hline
\ref{fig:features_fire} i.&    1&    0&     \cellcolor{green!25}1  \\ \hline
    &2&    0.001&  \cellcolor{green!25}    0.999  \\ \hline
    &3&     \cellcolor{green!25}1&    0  \\ \hline
    &4&    0&     \cellcolor{green!25}1  \\ \hline
    &5&    0&     \cellcolor{green!25}1  \\ \hline \hline
\ref{fig:features_fire} ii.&    1& \cellcolor{green!25}    1& 0  \\ \hline
    &2&     \cellcolor{green!25}1&    0  \\ \hline
    &3&     \cellcolor{green!25}1&    0  \\ \hline
    &4&     \cellcolor{green!25}1&    0  \\ \hline
    &5&     \cellcolor{green!25}0.999&    0.001  \\ \hline
\end{tabular}
\end{table}
\subsection{Accuracy and Precision}

\begin{figure}[!ht]
\begin{center}
\includegraphics[scale=0.1]{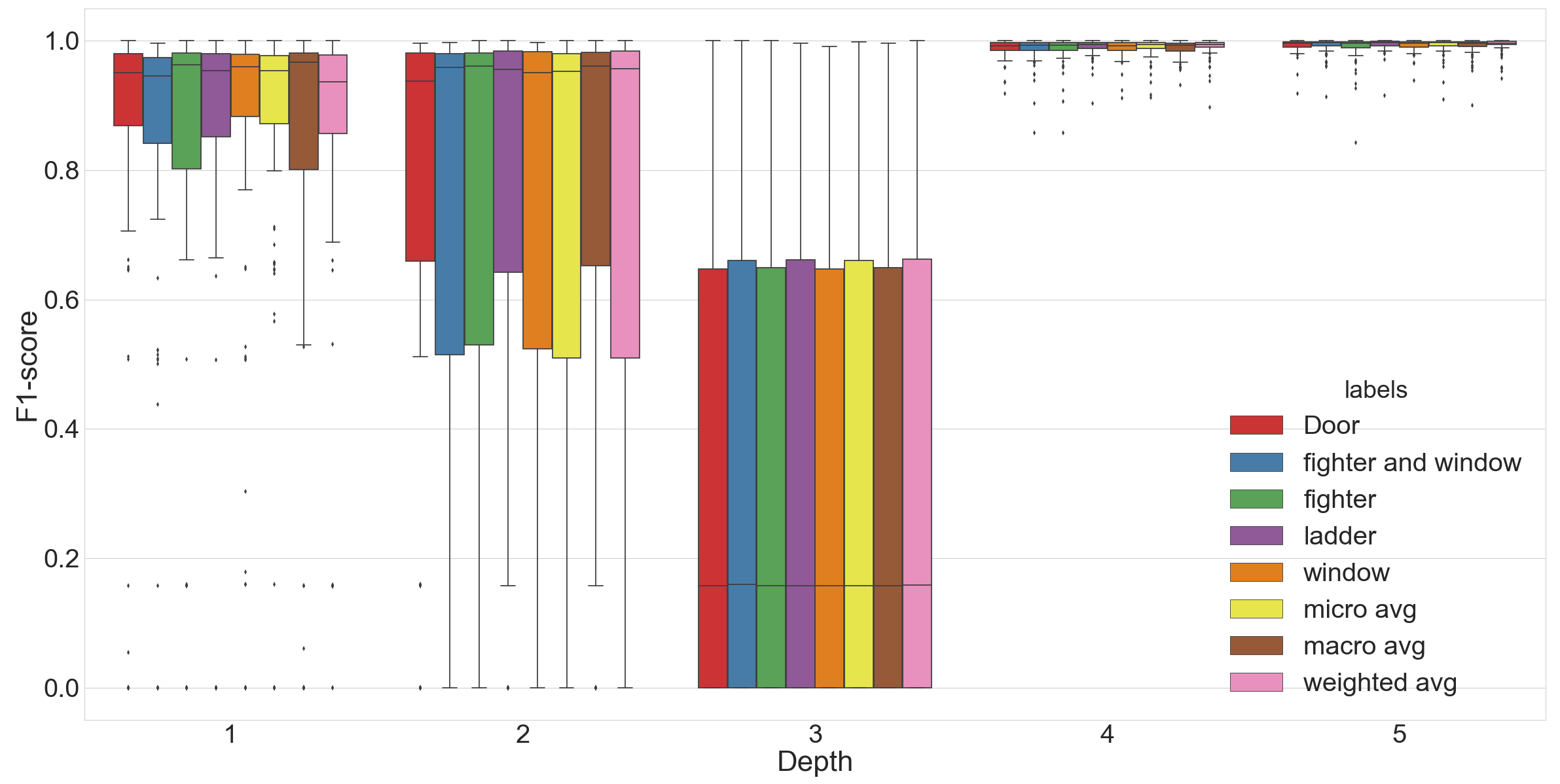}
\caption{F1 scores for the classification of objects with CNNs of different depths.}\label{fig:F1Scores_object}
\end{center}
\end{figure}

\begin{figure}[!ht]
\begin{center}
\includegraphics[scale=0.1]{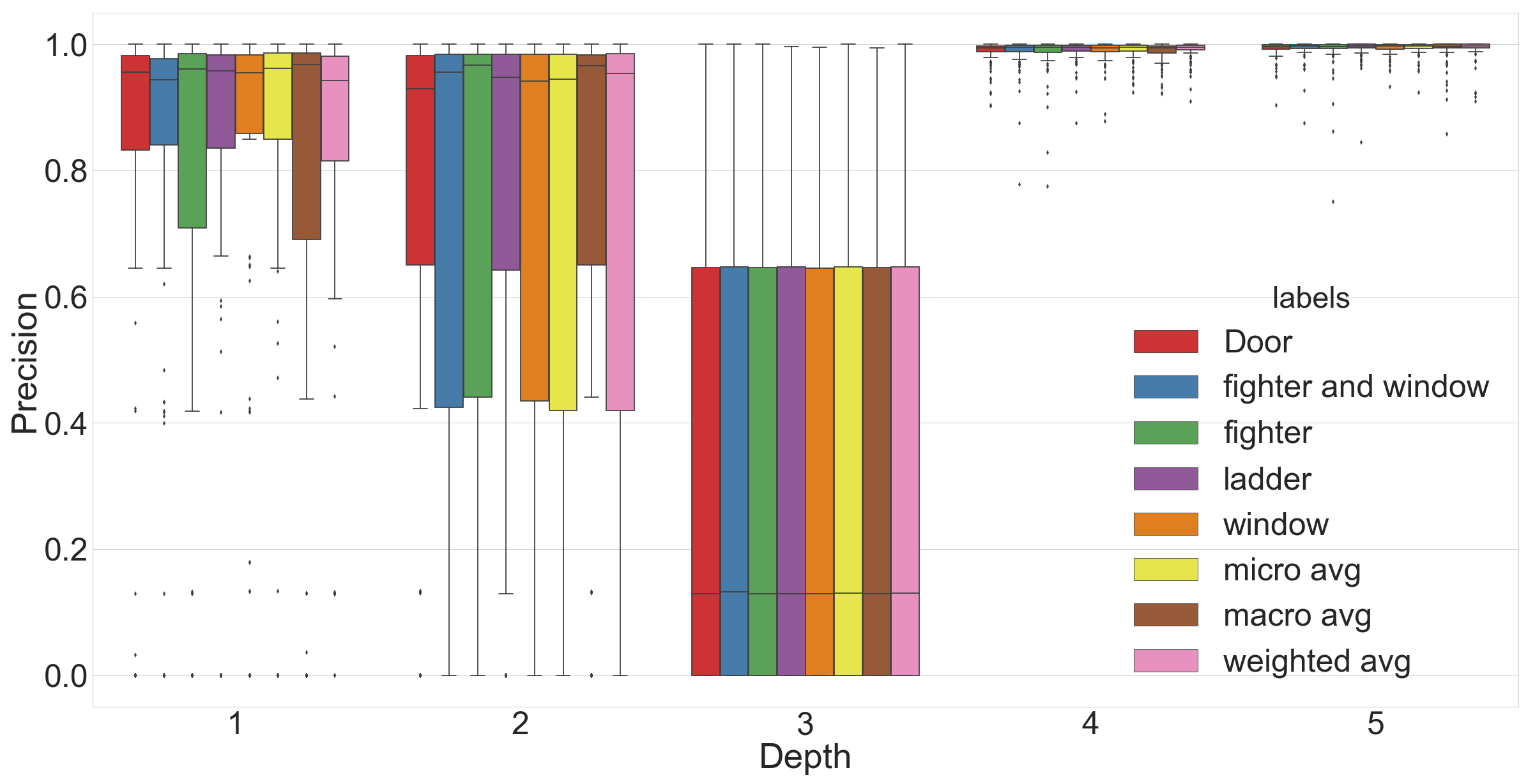}
\caption{Precision curves for object detection with CNNs of different depths.}\label{fig:precision_object}
\end{center}
\end{figure}

The accuracy of the neural network for different depths is consistent with visualizing the features and the t-SNE. Figures \ref{fig:F1Scores_object} and \ref{fig:precision_object} show the F1 scores and the precision of the network at depths 1 through 5. For this research, the networks have been trained to detect objects pertinent to fire navigation and rescue, including doors, people, ladders, windows, and the combination of firefighters and windows.  The network with one layer(depth 1) has a reasonable accuracy close to 75\% and shows a high variance in the results. The depth one framework requires a low computational burden in training and test. Computational burden increases with the growth in computational complexity. Each additional depth increases computational complexity due to the additional number of Floating Points(FLOPs) operations. Interestingly, the use of 2 and 3 depths produces an unacceptable performance but dramatically improves when using 4 and 5 depths. The use of  depth-5 is not necessary since its performance is almost identical to the run with four layers, but the computational burden is significantly higher. 

\begin{figure}[!ht]
\begin{center}
\includegraphics[scale=0.1]{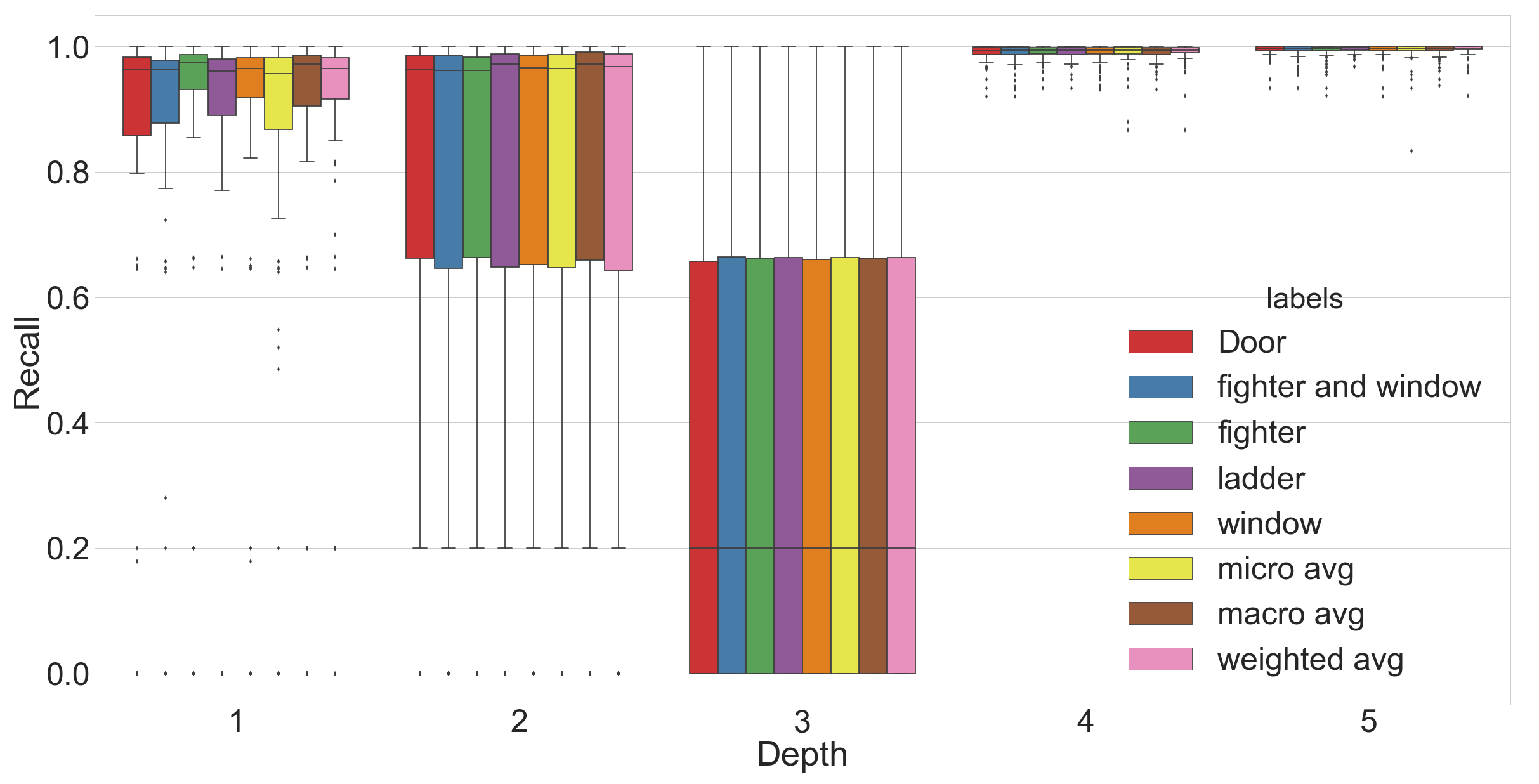}
\caption{Recall curves for object detection with CNNs of different depths.}\label{fig:recall_object}
\end{center}
\end{figure}

\begin{figure}[!ht]
\begin{center}
\includegraphics[scale=0.1]{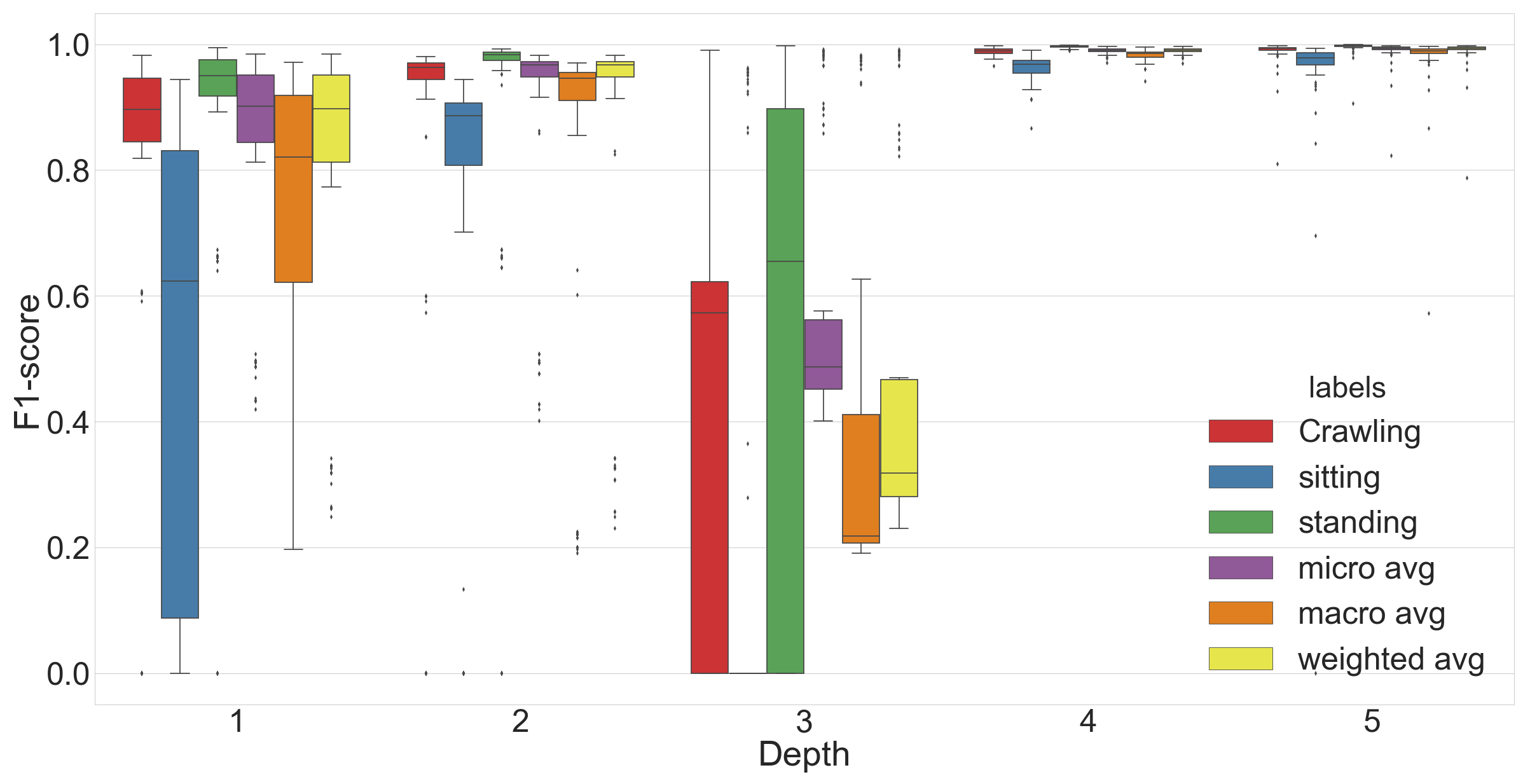}
\caption{F1 scores for the classification of poses with CNNs of different depths.}\label{fig:F1Scores_poses}
\end{center}
\end{figure}

\begin{figure}[!ht]
\begin{center}
\includegraphics[scale=0.1]{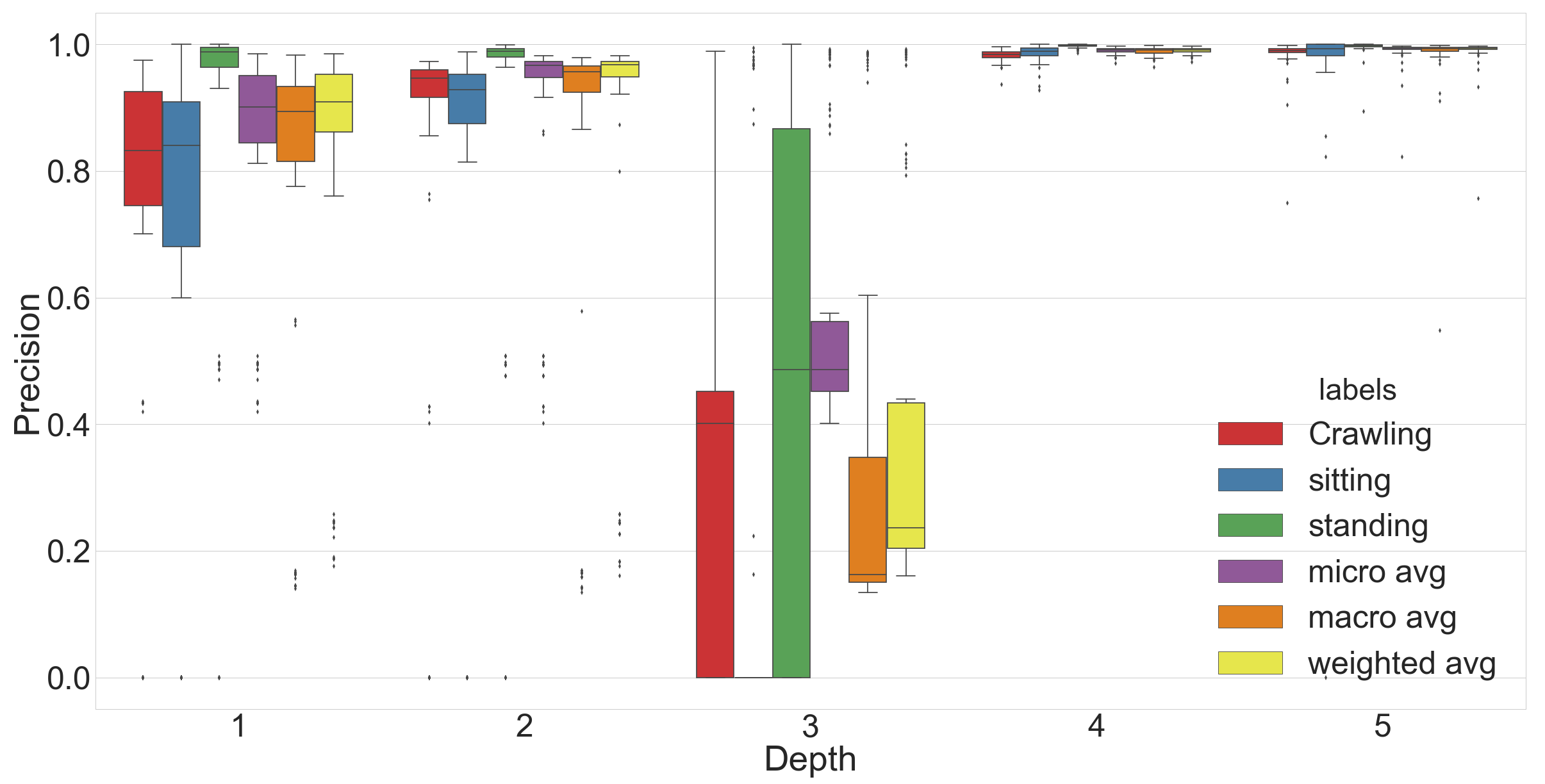}
\caption{Precision curves for poses detection with CNNs of different depths.}\label{fig:precision_poses}
\end{center}
\end{figure}

\begin{figure}[!ht]
\begin{center}
\includegraphics[scale=0.1]{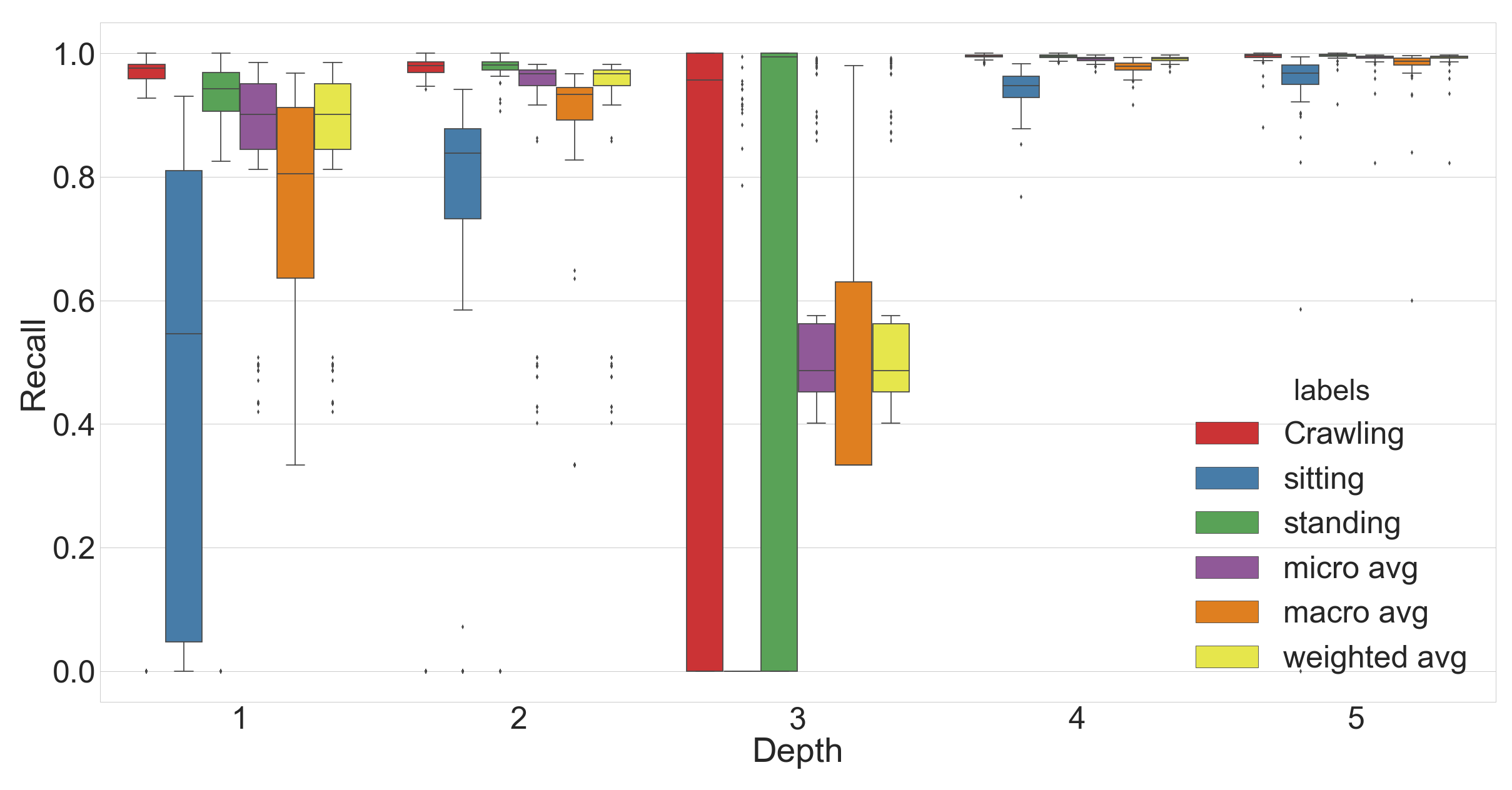}
\caption{Recall curves for poses detection with CNNs of different depths.}\label{fig:recall_poses}
\end{center}
\end{figure}

\begin{figure}[!ht]
\begin{center}
\includegraphics[scale=0.08]{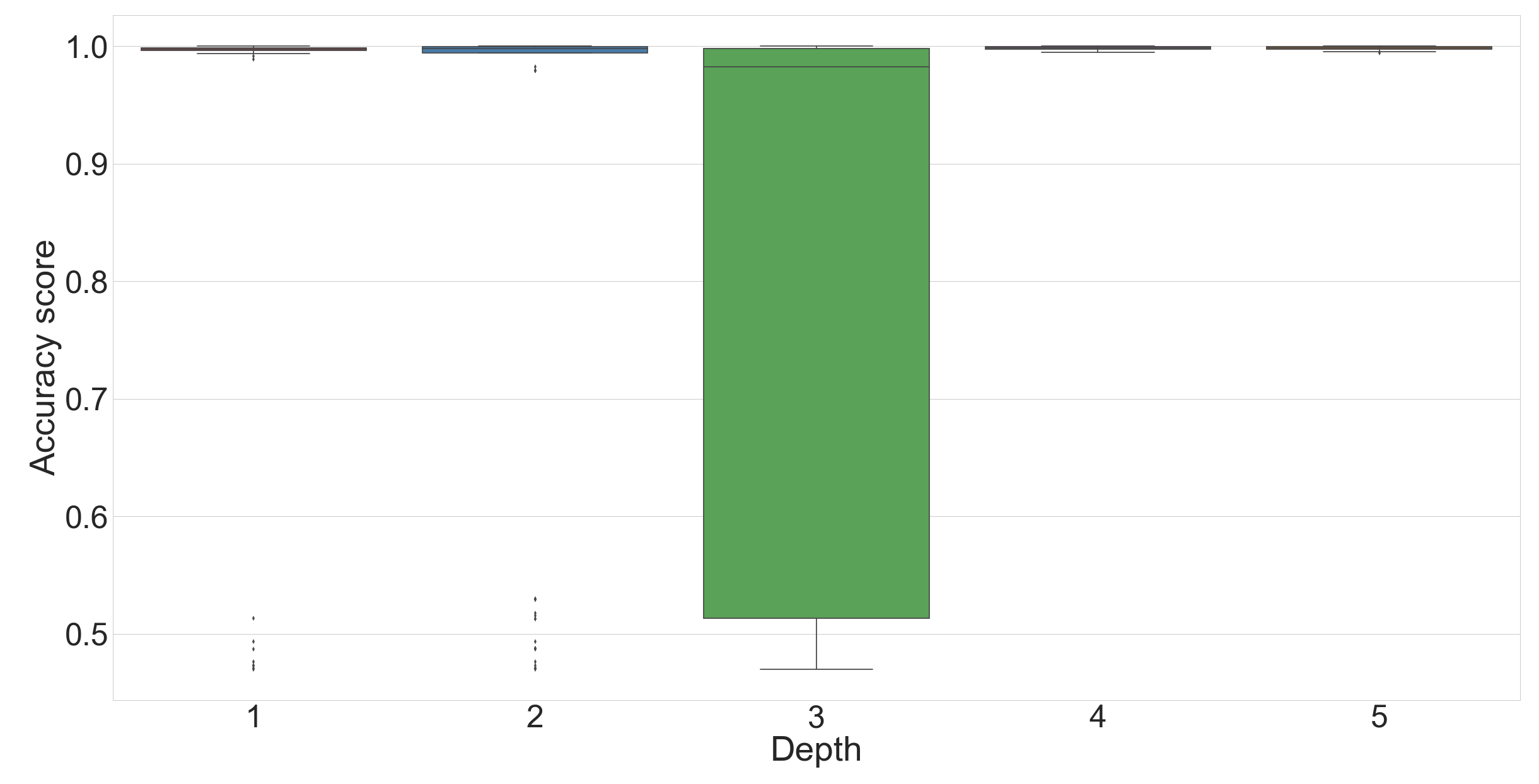}
\caption{Test accuracy in fire detection with CNNs of different depths.}\label{fig:fireAccuracy}
\end{center}
\end{figure}

\begin{figure}[!ht]
\begin{center}
\includegraphics[scale=0.08]{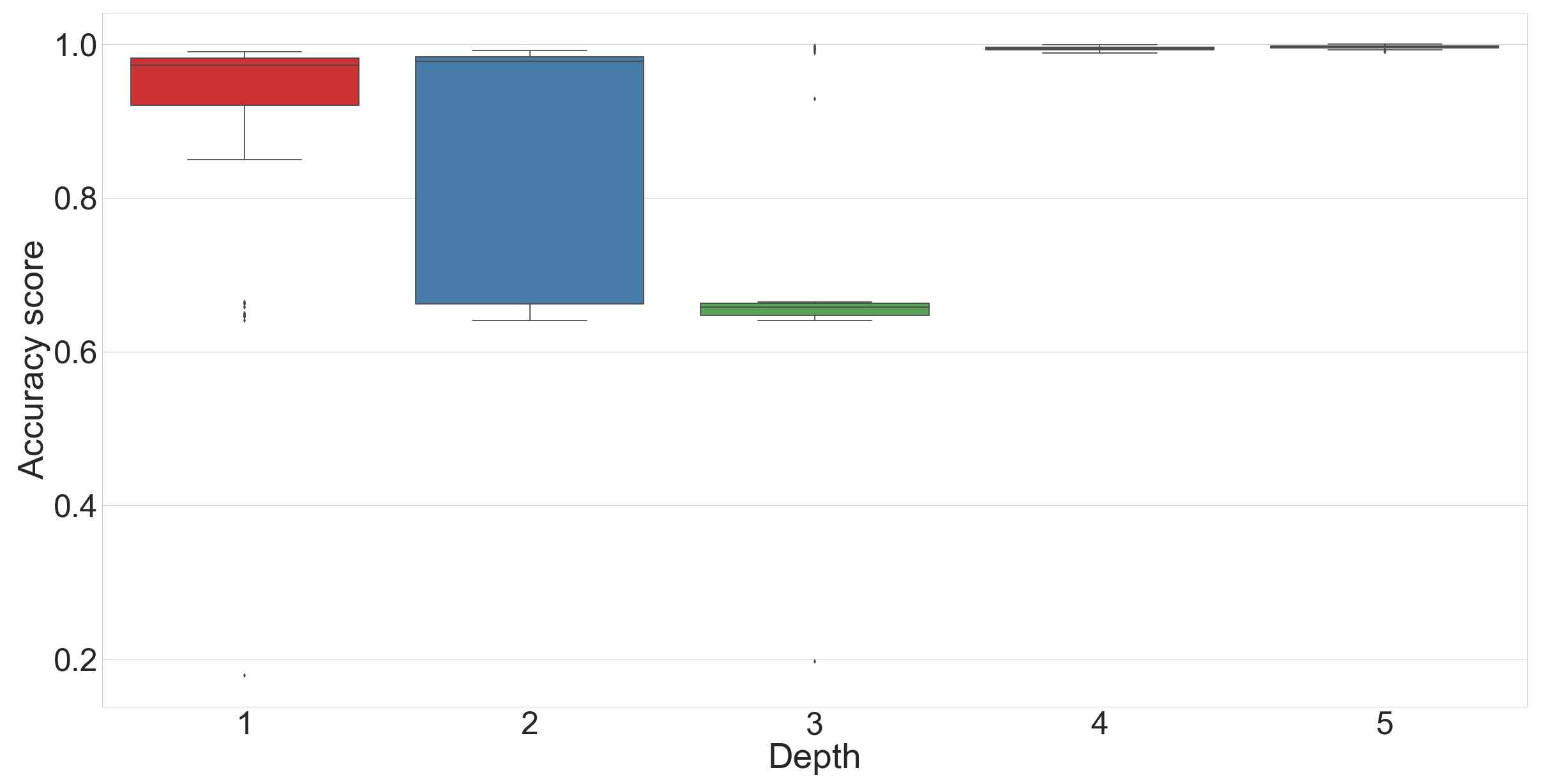}
\caption{Test accuracy in object detection with CNNs of different depths.}\label{fig:objectAccuracy}
\end{center}
\end{figure}

Figure \ref{fig:fireAccuracy} shows the accuracy of the network when detecting fire. In this case, it is only necessary to run the network with one layer. As in the previous experiments, the network with three layers produces a poor performance, while the other depth tests show an almost identical accuracy. The average test accuracy for all objects is depicted in figure \ref{fig:objectAccuracy}. From this graph, we can conclude that a good trade-off between computational burden and accuracy is obtained with a network of only one convolutional layer. In contrast, the highest level of accuracy is obtained using a network with four layers, which can achieve more than 97\%. 

\begin{figure}[!ht]
\begin{center}
\includegraphics[scale=0.09]{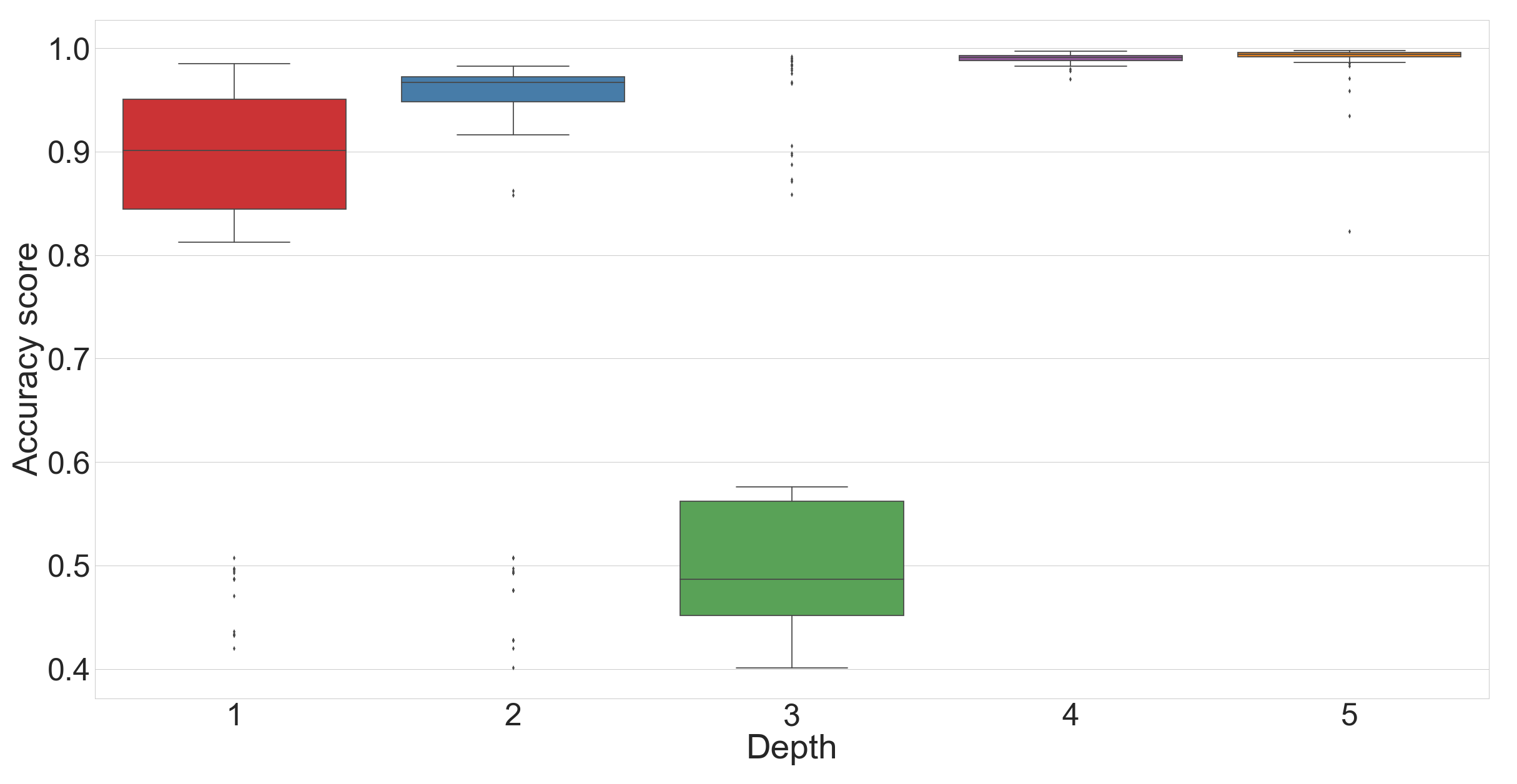}
\caption{Test accuracy in pose detection with CNNs of different depths.}\label{fig:poseAccuracy}
\end{center}
\end{figure}

The network's ability to accurately distinguish between different human positions and classify them accordingly presents new and exciting opportunities. With pose recognition, body position can be used to assist in making health inferences. For example, the presence of a person lying down is very important in a fire scenario, as it may be a significant indicator of a person who has succumbed to smoke inhalation and is in desperate need of rescue. The results can aid in alerting rescue teams to the possible health condition of victims and prioritize evacuation. In our tests, we trained the network to detect persons standing, sitting, or crawling from labeled images. The network shows a similar performance with an accuracy of 95\% on average.

\subsection{Confusion Matrices}
The confusion matrices for all detection modalities have been computed. Tables \ref{tab:confusionObjects1} to \ref{tab:confusionObjects5} shows the confusion matrices for the task of object and human detection. As stated before, the network with just one depth of convolutional blocks performs reasonably well without excess computational burden. The detection probabilities are around 75\%. However, we note that at lower depths(i.e., 1 and 2), the network is challenged to make clear distinctions between classes that tend to occur together. For example, both windows and ladders tend to occur in a scene with one or more firefighters. The network has a high confusion rate in determining windows and ladders as individual classes, as there is a confusion of about 19\% with firefighters in the classification of these elements. This is likely due to a significantly higher number of unaugmented images in the firefighter class than the others in the "object of interest" set. When the number of depths is increased to 2, the confusion is even worse, reaching 35\%. However, the accuracy of the detection of firefighters as a class is between 97 and 99\%. If the number of layers is increased to 4, the confusion matrices show confusion between 0 and 1.7\% for all objects being classified. As stated above, there are no significant differences between 4 and 5 depths. Note that the classification performance with these number of depths ranges between 97 and 99.8\%.

\begin{table}[!ht]
\centering
\caption{Confusion matrix for object detection, depth=1}
\label{tab:confusionObjects1}
\begin{tabular}{|l|r|r|r|r|r|}
\hline
Door & \cellcolor[HTML]{C0C0C0}73.2 & 1.1 & 23.5 & 0.0 & 2.2 \\ \hline
F/W & 0.0 & \cellcolor[HTML]{C0C0C0}77.7 & 19.0 & 0.0 & 3.3 \\ \hline
Fighter & 0.3 & 1.8 & \cellcolor[HTML]{C0C0C0}97.0 & 0.1 & 0.8 \\ \hline
Ladder & 0.0 & 1.2 & 19.1 & \cellcolor[HTML]{C0C0C0}79.6 & 0.1 \\ \hline
Window & 0.3 & 4.3 & 19.9 & 0.0 & \cellcolor[HTML]{C0C0C0}75.5 \\ \hline
 & \multicolumn{1}{l|}{Door} & \multicolumn{1}{l|}{F/W} & \multicolumn{1}{l|}{Fighter} & \multicolumn{1}{l|}{Ladder} & \multicolumn{1}{l|}{Window} \\ \hline
\end{tabular}
\end{table}

\begin{table}[!ht]
\centering
\caption{Confusion matrix for object detection, depth=2}
\label{tab:confusionObjects2}
\begin{tabular}{|l|l|l|l|l|l|}
\hline
Door & \cellcolor[HTML]{C0C0C0}61.3 & 0.0 & 38.2 & 0.0 & 0.5 \\ \hline
F/W & 0.0 & \cellcolor[HTML]{C0C0C0}62.4 & 36.0 & 0.0 & 1.7 \\ \hline
Fighter & 0.1 & 0.2 & \cellcolor[HTML]{C0C0C0}99.4 & 0.1 & 0.2 \\ \hline
Ladder & 0.0 & 0.0 & 34.3 & \cellcolor[HTML]{C0C0C0}65.7 & 0 \\ \hline
Window & 0.1 & 1.0 & 35.9 & 0 & \cellcolor[HTML]{C0C0C0}63.0 \\ \hline
 & Door & F/W & Fighter & Ladder & Window \\ \hline
\end{tabular}
\end{table}

\begin{table}[!ht]
\centering
\caption{Confusion matrix for object detection, depth=4}
\label{tab:confusionObjects4}
\begin{tabular}{|l|l|l|l|l|l|}
\hline
Door & \cellcolor[HTML]{C0C0C0}97.1 & 0.2 & 0.9 & 0.0 & 1.7 \\ \hline
F/W & 0.0 & \cellcolor[HTML]{C0C0C0}98.3 & 1.0 & 0.0 & 0.7 \\ \hline
Fighter & 0.1 & 0.1 & \cellcolor[HTML]{C0C0C0}99.8 & 0.0 & 0.0 \\ \hline
Ladder & 0.0 & 0.0 & 0.2 & \cellcolor[HTML]{C0C0C0}99.8 & 0.0 \\ \hline
Window & 0.3 & 0.5 & 0.3 & 0.0 & \cellcolor[HTML]{C0C0C0}98.9 \\ \hline
 & Door & F/W & Fighter & Ladder & Window \\ \hline
\end{tabular}
\end{table}

\begin{table}[!ht]
\centering
\caption{Confusion matrix for object detection, depth=5}
\label{tab:confusionObjects5}
\begin{tabular}{|l|l|l|l|l|l|}
\hline
Door & \cellcolor[HTML]{C0C0C0}98.0 & 0.0 & 0.7 & 0.0 & 1.3 \\ \hline
F/W & 0.0 & \cellcolor[HTML]{C0C0C0}99.2 & 0.5 & 0.0 & 0.3 \\ \hline
Fighter & 0.0 & 0.1 & \cellcolor[HTML]{C0C0C0}99.9 & 0.0 & 0.0 \\ \hline
Ladder & 0.0 & 0.0 & 0.1 & \cellcolor[HTML]{C0C0C0}99.9 & 0.0 \\ \hline
Window & 0.3 & 0.6 & 0.2 & 0.0 & \cellcolor[HTML]{C0C0C0}98.9 \\ \hline
 & Door & F/W & Fighter & Ladder & Window \\ \hline
\end{tabular}
\end{table}

A similar trend in detection accuracy related to the number of depths can be seen in detecting poses. Results at depths 1 and 2 perform poorly compared to 4 and 5 as shown in Tables \ref{tab:confusionPoses1},  \ref{tab:confusionPoses2}, \ref{tab:confusionPoses4} and  \ref{tab:confusionPoses5} . At depths 1 and 2, the network was mainly challenged in distinguishing the differences between sitting and crawling. This could be explained by the firefighters' relative positions in these two poses are similar but rotated. The accuracy increases significantly with four depths (Table \ref{tab:confusionPoses4}), where the confusion decreases to a range between 0 and 0.4 in all cases except for the detection of the sitting pose, which stands at a confusion rate of 5.7 (in the network with four depths) and 5.1\% with the crawling pose with five depths (Table \ref{tab:confusionPoses5}). Again we found no significant differences between the results of these two network configurations.   

\begin{table}[!ht]
\centering
\caption{Confusion matrix for pose detection, depth=1}
\label{tab:confusionPoses1}
\begin{tabular}{|l|r|r|r|}
\hline
Crawling & \cellcolor[HTML]{C0C0C0}85.6 & 1.2 & 13.2 \\ \hline
Sitting & 41.4 & \cellcolor[HTML]{C0C0C0}45.9 & 12.7 \\ \hline
Standing & 10.8 & 0.4 & \cellcolor[HTML]{C0C0C0}88.8 \\ \hline
 & \multicolumn{1}{l|}{Crawling} & \multicolumn{1}{l|}{Sitting} & \multicolumn{1}{l|}{Standing} \\ \hline
\end{tabular}
\end{table}

\begin{table}[!ht]
\centering
\caption{Confusion matrix for pose detection, depth=2}
\label{tab:confusionPoses2}
\begin{tabular}{|l|l|l|l|}
\hline
Crawling & \cellcolor[HTML]{C0C0C0}86.3 & 0.8 & 12.9 \\ \hline
Sitting & 19.3 & \cellcolor[HTML]{C0C0C0}67.8 & 12.9 \\ \hline
Standing & 7.5 & 0.2 & \cellcolor[HTML]{C0C0C0}92.3 \\ \hline
 & Crawling & Sitting & Standing \\ \hline
\end{tabular}
\end{table}

\begin{table}[!ht]
\centering
\caption{Confusion matrix for pose detection, depth=4}
\label{tab:confusionPoses4}
\begin{tabular}{|l|l|l|l|}
\hline
Crawling & \cellcolor[HTML]{C0C0C0}99.5 & 0.2 & 0.3 \\ \hline
Sitting & 5.7 & \cellcolor[HTML]{C0C0C0}94.1 & 0.2 \\ \hline
Standing & 0.5 & 0.0 & \cellcolor[HTML]{C0C0C0}99.5 \\ \hline
 & Crawling & Sitting & Standing \\ \hline
\end{tabular}
\end{table}

\begin{table}[!ht]
\centering
\caption{Confusion matrix for pose detection, depth=5}
\label{tab:confusionPoses5}
\begin{tabular}{|l|l|l|l|}
\hline
Crawling & \cellcolor[HTML]{C0C0C0}99.4 & 0.2 & 0.4 \\ \hline
Sitting & 5.1 & \cellcolor[HTML]{C0C0C0}94.7 & 0.2 \\ \hline
Standing & 0.5 & 0.0 & \cellcolor[HTML]{C0C0C0}99.5 \\ \hline
 & Crawling & Sitting & Standing \\ \hline
\end{tabular}
\end{table}

\subsection{Detection  Versus False Alarm Probabilities}

\begin{figure}[!ht]
\centering
\begin{subfigure}[b]{0.3\textwidth}
\centering
\includegraphics[scale=0.11]{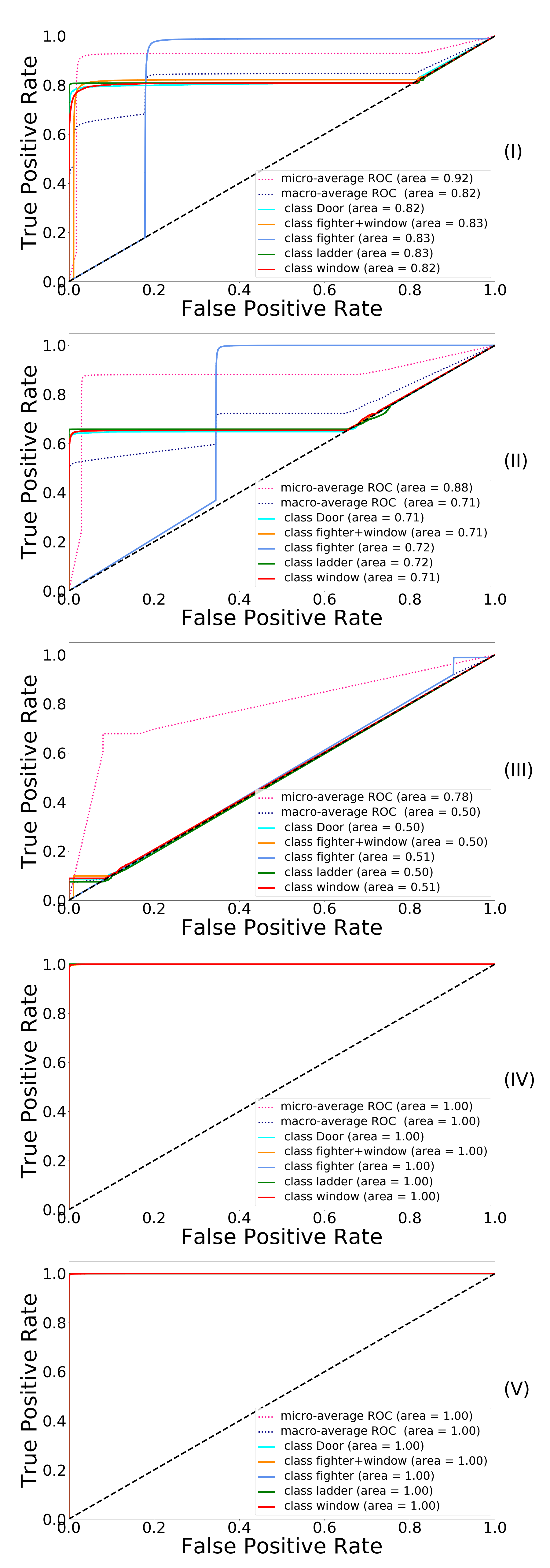}
\caption{ROC curve for object detection for I) Depth-1, II) Depth-2, III) Depth-3, IV) Depth-4 and V) Depth-5 architectures. }\label{fig:ROC_object}
\end{subfigure}
\hfill
\begin{subfigure}[b]{0.3\textwidth}
\centering
\includegraphics[scale=0.11]{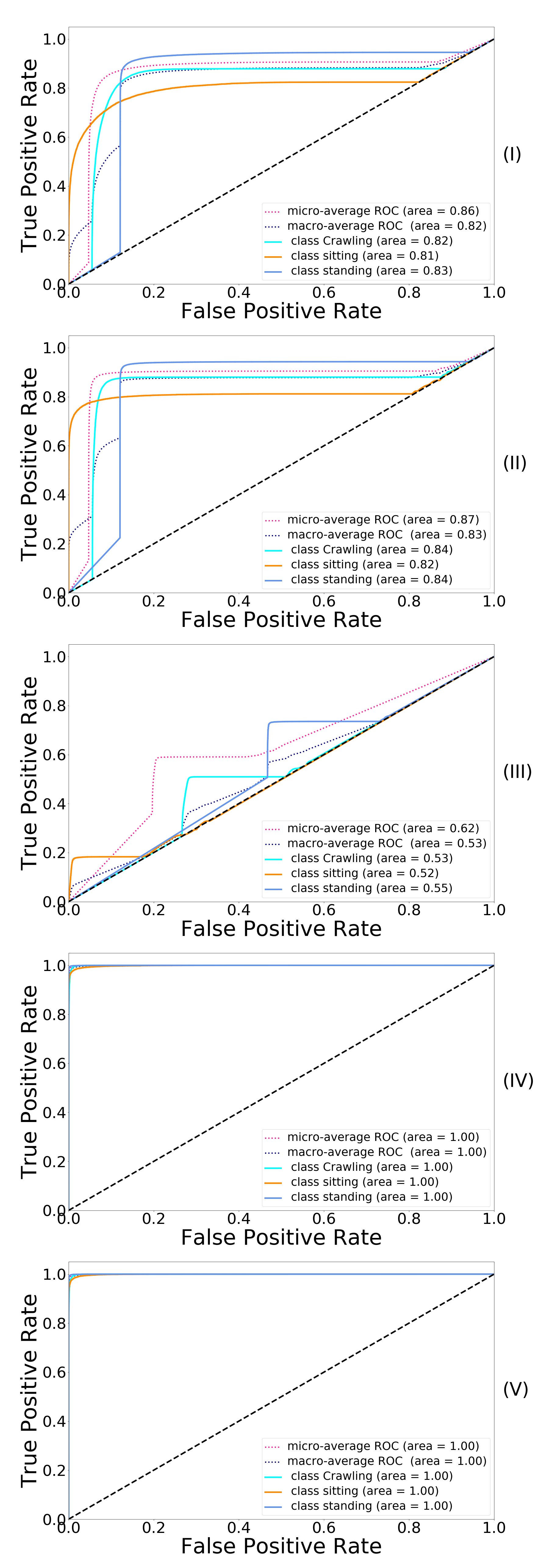}
\caption{ROC curve for pose detection for I) Depth-1, II) Depth-2, III) Depth-3, IV) Depth-4 and V) Depth-5 architectures .}\label{fig:ROC_pose}
\end{subfigure}
\hfill
\begin{subfigure}[b]{0.3\textwidth}
\centering
\includegraphics[scale=0.11]{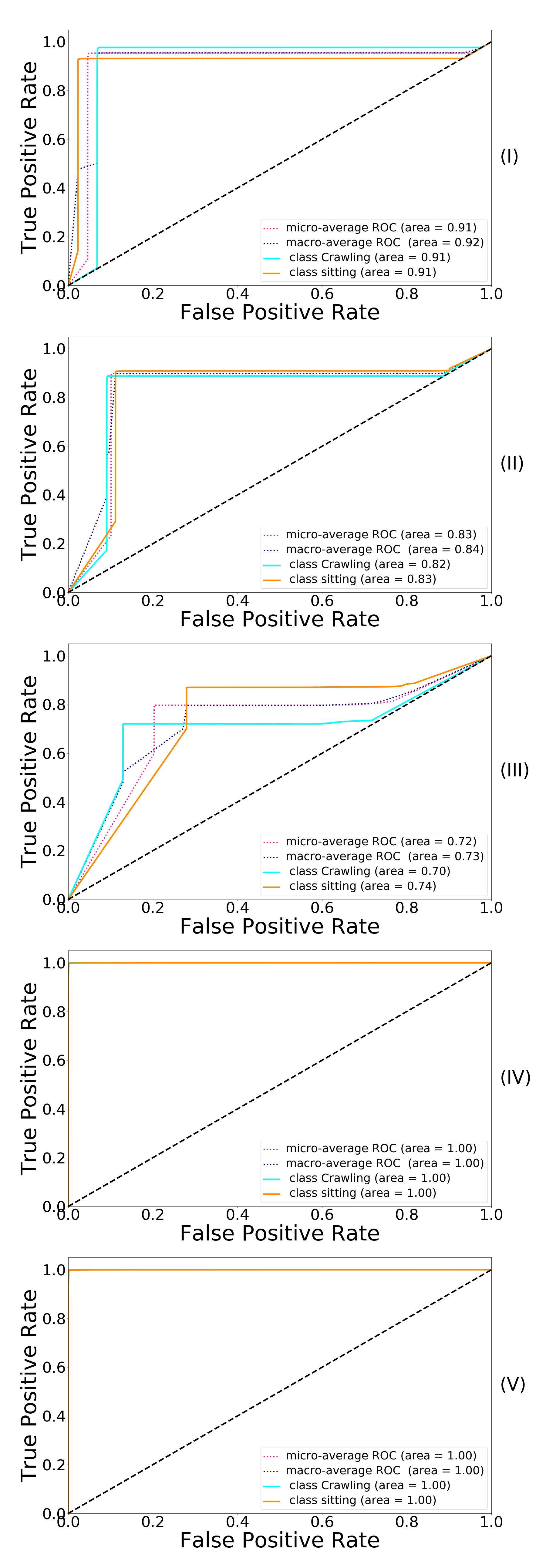}
\caption{ROC curve for fire detection for I) Depth-1, II) Depth-2, III) Depth-3, IV) Depth-4 and V) Depth-5 architectures .}\label{fig:ROC_fire}
\end{subfigure}
\end{figure}

Figures \ref{fig:ROC_object}, \ref{fig:ROC_pose} and \ref{fig:ROC_fire}  show the probability of detection versus false alarm for the classification of objects and humans, poses and fire. The graphs have been obtained by sweeping the detection threshold from 0 to 1. In all cases, it is observed that the probability of false alarm is negligible when the networks have a depth of 4 or 5. As expected, given the classifier's poor performance at depths 2 and 3, the results showed high false alarm probabilities. The results show flaws in the probability of detection versus probability of false alarm in the depth one network. A detection rate of about 82\% is achieved with a false alarm rate of about 18\% (figure \ref{fig:ROC_object})for object detection as per the micro/macro average ROC area. The optimal false alarm rates in individual class object detection range from 15-20\%, particularly in the detection of humans. Similar results are obtained in pose detection. Therefore, although depth one networks can be used with a lower computational burden, they should only be used when such false alarm rates can be tolerated. False alarm probability can be decreased if the detection is performed on a sequence of images chronologically, and then a voting procedure is applied to all detections. This would be done at the cost of additional computational time in the object or pose detection procedures.

\chapter{Object Detection,Tracking and Segmentation in Firefighting Environments}

\section{Motivation}
The research objective discussed in the following chapter is to create an automated system that is capable of real-time, intelligent object detection, tracking, and segmentation and facilitates the improved situational awareness of firefighters during an emergency response. We have explored state-of-the-art machine/deep learning techniques to achieve this objective. This work aims to enhance the situational awareness of firefighters by effectively exploiting the infrared video that is actively recorded by firefighters on the scene. To accomplish this, we use a trained deep Convolutional Neural Network (CNN) system to detect and track objects of interest from thermal imagery in real-time. Amid those critical circumstances created by a structure fire, this system can accurately inform firefighters’ decision-making process with up-to-date scene information by extracting, processing, and analyzing crucial information. Utilizing the new information produced by the framework, firefighters can make more informed decisions as they progress in fighting the fire at hand and continue their search and rescue efforts through hazardous and potentially catastrophic environments.
\par
Despite the large quantity of work related to the processing of infrared images in fireground or related scenarios, to our knowledge, no work has been published that attempts to construct a system that integrates online detection and tracking of targets of interest in these scenarios, including humans, door, windows, and fire. Thus, there is a need for effective automatic target detection generated in real-time in a firefighting environment and the associated need for a highly accurate classifier.
Our research seeks to address these needs. Such processed information can then be fed to firefighter’s augmented reality for an informed situational awareness to aid in navigation through the fire environment. 

\par
To achieve the goal, we have implemented a novel mask RCNN framework combining the capabilities of detection, tracking, and segmentation concurrently. The detection and tracking are first performed by the Faster R-CNN built on top of the CNN framework from Chapter2. Further, the Faster R-CNN is extended to Mask R-CNN by adding a 1x1 convolutional layer for performing object segmentation to achieve more precise localization of objects of interest. 

\section{System Description}

In the deep learning community, Region-Based CNN(R-CNNs) are very popular for object detection. They function to precisely locate an object in an image and enable tracking of the image across multiple sequential frames.  This process involves a region proposal generation and classification process. 
Instead of using an exhaustive selective search at the input stage of CNN as R-CNN and fast R-CNN, Faster R-CNN speeds up the detection process by introducing a region-proposal network(RPN) at the end of the feature extraction network. The RPN module processes the feature map and produces region proposals used by classification and regression heads for bounding box and label generation. 
\par
Prior to the development of deep learning models, various feature engineering techniques such as Histogram of the Oriented Gradient(HOG)\cite{dalal2005histograms}, Scale Invariant Feature Transform(SIFT)\cite{lowe1999object}, optical flow\cite{horn1981determining}, Fischer vectors\cite{sanchez2013image} and machine learning techniques such as Support vector machines(SVM)\cite{suykens1999least}, logistic regression\cite{hosmer2013applied}, random forest\cite{liaw2002classification} and decision trees \cite{quinlan1987simplifying} were applied to the extracted features for classification and detection tasks. Despite the optimal hyper-parameter search, the methods performed poorly. This is due to the limited color, texture, and contour information present in infrared imagery. The shallow machines could also not reach the level of complexity needed for the classification and detection tasks. Besides, the tracking and segmentation frameworks are based on the classification and regression blocks, which are needed to be trained separately, making their integration difficult in a combination of subsystems like the one in \ref{fig:structure} \par

\begin{figure*}
\centering
\begin{subfigure}[b]{0.4\linewidth}
	\includegraphics[width=\linewidth]{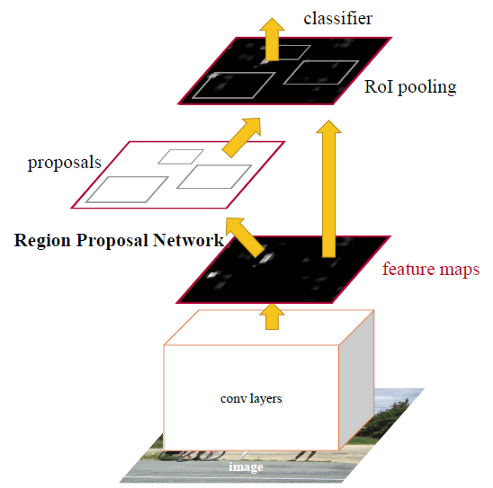}
	\caption{i.}
\end{subfigure}
\begin{subfigure}[b]{0.4\linewidth}
	\includegraphics[width=\linewidth]{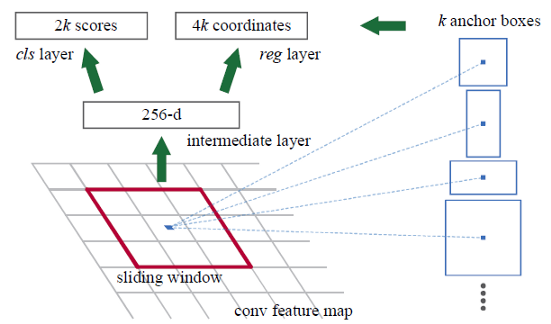}
	\caption{ii.}
\end{subfigure}
\caption{i) Overview of RPN framework and ii) Overview of Regression and classification head}
\label{fig:frcnn}
\end{figure*}

\begin{figure}[!ht]
    \centering
    \includegraphics[scale=0.5]{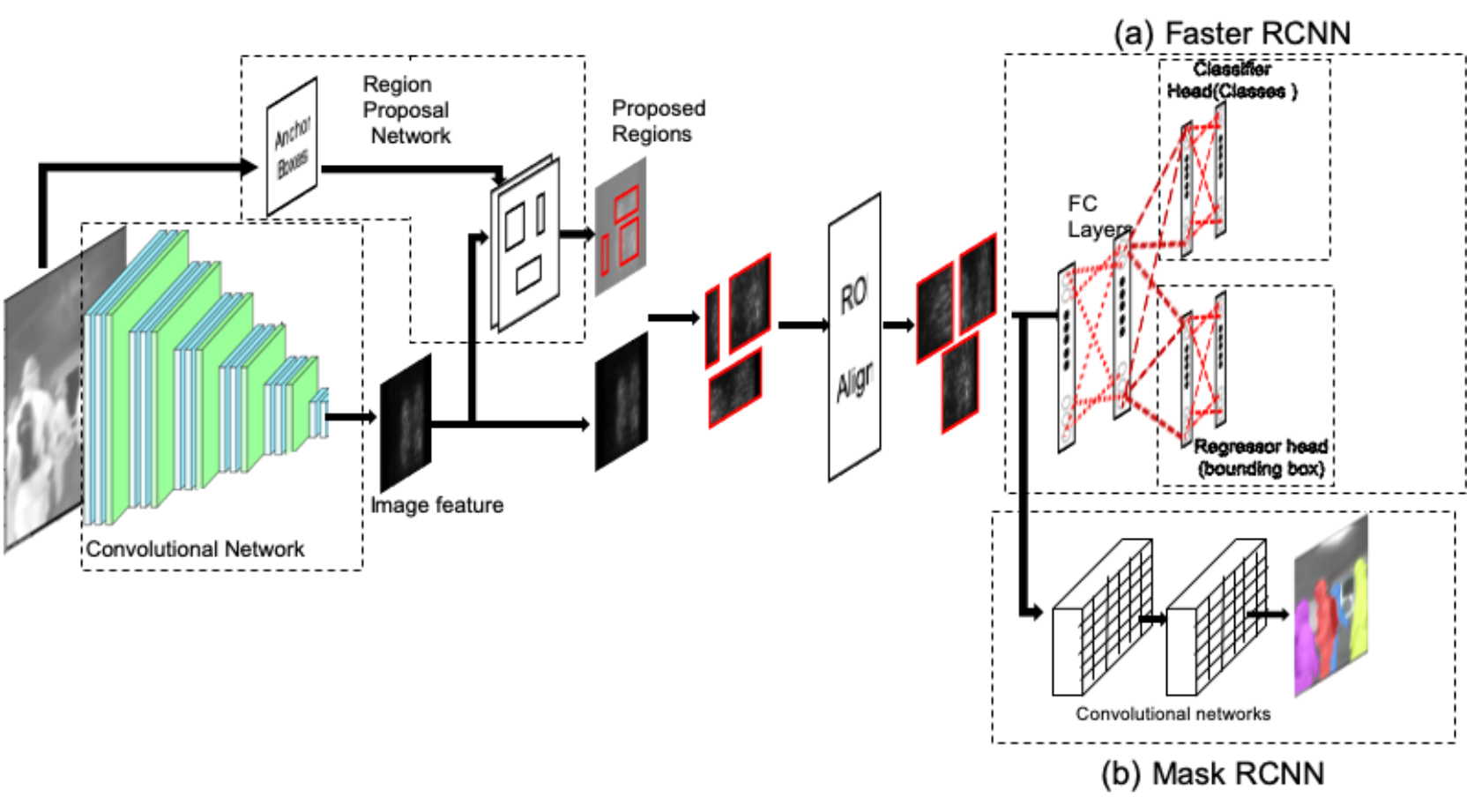}
    \caption{Both Mask R-CNN and Faster R-CNN share a common Region Proposal Network(RPN) for significant feature extraction. For object tracking, the RPN’s processed output is fed to a classifier and a regressor to output class scores and bounding box coordinates, respectively. This is performed by Faster RCNN, as shown in (a). For image segmentation, RPN’s same processed output is fed to a Convolution Network that generates a mask when overlayed with the original image, resulting in semantic segmentation as its output.}
    \label{fig:structure}
\end{figure}

Region Proposal Network is a fully convolutional network that operates on the extracted features with different anchor boxes generating scores, as shown in \ref{fig:frcnn}. Anchor boxes are a sliding window with a variable size that act on CNN feature maps and for each window, output k potential bounding boxes and scores that measure the amount of accuracy of each box, i.e., bounding box coordinates and scores that indicate how likely the image in the given bounding box would be an object of interest. These bounding boxes are further processed by the classification heads and regression heads to generate classification and tighten the bounding boxes.  The objective training function for the Faster RCNN is the total loss, considering the classification and bounding box generation. 
Mask-RCNN is an extension to the Faster-RCNN for pixel-level segmentation. Faster RCNNs can only produce the bounding boxes for each object and become confused when the objects of interest are highly overlapped in an image. Mask R-CNN addresses this issue by locating exact pixels in each object. 
Mask R-CNN adds a fully convolutional network (FCN) branch to Faster R-CNN that processes the CNN feature map and outputs a binary mask indicating whether it belongs to the object in question or not.  There is a small crucial correction on the mask RCNN when the convolutional block is added on top of Faster RCNN called RoiAlign, which stands for Realigning RoIPool. This step fixes the misalignment of feature maps selected by RoIPool compared to regions of the original image. This is achieved by the bilinear implementation for a lower resolution feature map to align the segmentation with the original image. Having generated these masks, these are combined with classification and bounding boxes from faster R-CNN to generate precise segmentation maps.  The training loss comprises classification error, bounding box regression error, and mask generation error. For this reason, training a Mask RCNN is slower compared to the faster RCNN due to the additional objective of minimizing error due to overlapping objects.

The tasks of object recognition and tracking have the structure depicted in Figure \ref{fig:structure}.  
The components for the object detection and tracking framework(i.e., Faster RCNN are discussed as follows: 
\begin{itemize}
    \item \textbf{Feature Extractor :} 
Using a conventional CNN, the role of this part is to extract the features from input frames. The framework uses one of the CNN blocks such as VGG, Inception, ResNet, or other existing models. The input to this block is image data, and the output is the k h$\times$w feature maps. The elements can be fine-tuned in the training process.
\item \textbf{Proposal Generation with Region Proposal Network:} 
In addition to a simple classifier involved in a deep CNN system, the proposed system involves a regressor. The tracking modality’s key element
is using a regressor, which takes in the feature map generated by the feature extractor and outputs a bounding box (x,y,w,h) and class probability.
Here x,y stands for the bounding box’s initial reference position, and
w,h corresponds to the extent of the box along with horizontal and vertical directions, respectively. To localize the object with the regression, we need to
train the system with an image and a ground truth and use L2 distance
to calculate the loss between the predicted bounding box and the ground
truth. Fig. 15 shows the overall outlook corresponding to our proposed
RCNN system which involves the classification head and regression head
\end{itemize}

The overall detection and tracking framework has been summarized below:
\begin{itemize}
    \item 	First, the picture goes through convolutional layers, and feature maps are extracted.
    \item Then, a sliding window is used in RPN for each location over the feature map.
    \item For each location, k (k=9) anchor boxes are used (3 scales of 128, 256, and 512, and 3 aspect ratios of 1:1, 1:2, 2:1) for generating region proposals.
    \item A classifier layer outputs 2k scores depending on whether there is an object or not for k boxes
    \item A regressor layer outputs 4k for the coordinates (box center coordinates, width, and height) of k boxes.
    \item With the size of the W×H feature map, there are WHk anchors in total.
\end{itemize}
A detailed overview of the framework is shown in \ref{fig:frcnn}. 

The RPN loss function is computed as 

\begin{equation}
    L(p_i,t_i)=1/N_{cls} \sum_{i} L_{cls}(p_i,p_i^{*}) + \lambda * 1/N_{reg} \sum_{i} p_{i}^{*} L_{reg}(t_{i},t_{i}^{*})
\end{equation}
 
The first term is the classification loss over two classes (presence or absence of an object). The second term is the regression loss of bounding boxes only when there is an object (i.e., $p_{i}^{*}=1)$.

Mask R-CNN (regional convolutional neural network) is a two-stage framework: the first stage scans the image and generates proposals(areas likely to contain an object). And the second stage classifies the proposals and generates bounding boxes and masks.
The backbone of a Mask RCNN is a Feature Pyramid Network(FPN) style deep neural network. FPN improves the standard feature extraction pyramid by adding a second pyramid that takes the high-level features from the first pyramid and passes them down to lower layers. Doing so allows features at every level to have access to both lower and higher-level features.
The Mask RCNN framework consists of a bottom-up pathway, a top-bottom pathway, and lateral connections. Bottom-up pathways comprise standard convolutional neural nets such as ResNet or VGG, which extract features from raw images. The top-bottom pathway generates a feature pyramid map that is similar in size to the bottom-up pathway. Lateral connections are convolution and add operations between two corresponding levels of the two pathways. The advantage of adding an FPN compared to ConvNets is that it maintains semantically robust features at various resolution scales. \par
\begin{figure}[!ht]
	\centering
	\includegraphics[width=.7\textwidth]{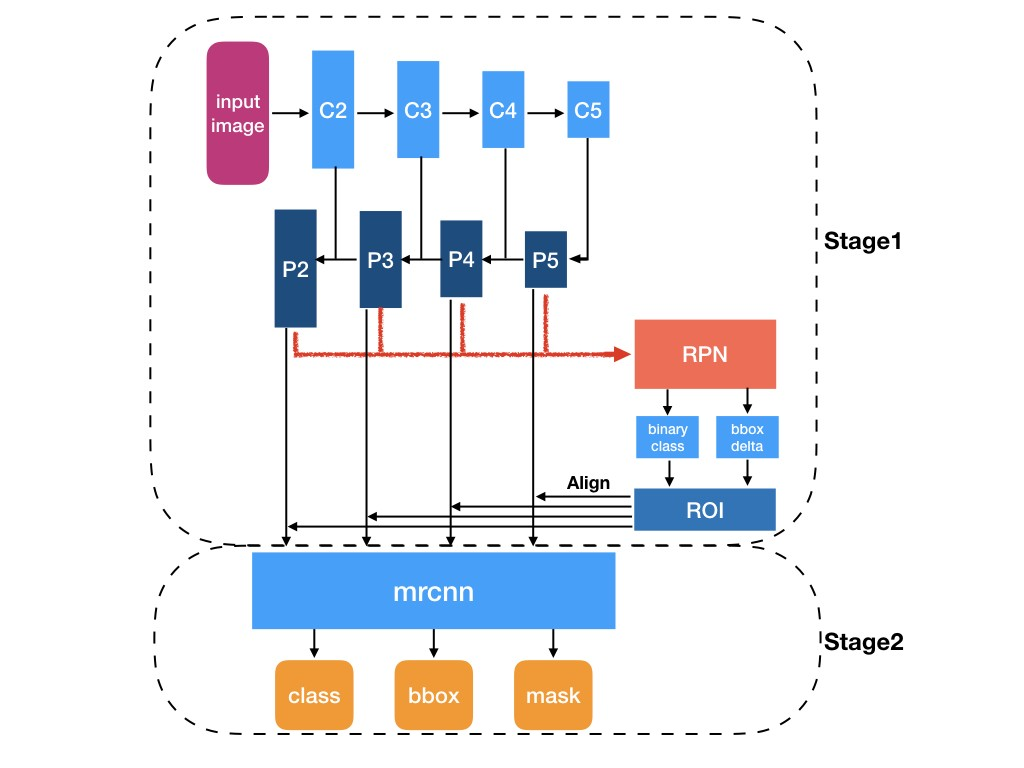}
	\caption{A detailed Overview of Mask RCNN with FPN framework}
	\label{mrcnn2}
\end{figure}
At the first stage, a lightweight neural network called Region proposal network(RPN) scans all FPN top-bottom pathways (hereafter referred to as feature map) and proposes regions which may contain objects. While scanning the feature map is an efficient way, anchors bind features to the raw image location. Anchors are a set of boxes with predefined locations and scales relative to images. Ground-truth classes( only object or background binary classified at this stage) and bounding boxes are assigned to individual anchors according to some Intersection of Unions(IoU) value. As anchors with different scales bind to different levels of feature maps, RPN uses these anchors to figure out where the feature map ‘should’ get an object and what size its bounding box should be. After convolving, downsampling and upsampling would keep features staying in the same relative locations as the objects in the original image. \par
At the second stage, another neural network takes proposed regions by the first stage and assigns them to several specific areas of a feature map level, scans these areas, and generates objects classes (multi-categorical classified), bounding boxes, and masks. The procedure looks similar to RPN. Differences are that without the help of anchors, stage-two used a trick called ROIAlign to locate the feature map’s relevant areas, and there is a branch generating a mask for each object at the pixel level. 

In Mask RCNN, different layers in a neural network learn features with different scales, just like the anchors and ROIAlign. 

The multi-task loss function of Mask R-CNN combines the loss of classification, localization and segmentation mask: $L=L_{cls}+L_{bbox}+L_{mask}$, where $L_{cls}$ and $L_{bbox}$ are the same as in Faster R-CNN.
The mask branch generates a mask of dimension $m \times m$ for each RoI and each class; K classes in total. Thus, the total output is of size $K \times m^2$
Because the model is trying to learn a mask for each class, there is no competition among classes for generating masks.
$L_{mask}$ is defined as the average binary cross-entropy loss, only including k-th mask if the region is associated with the ground truth class k.  
\begin{equation}
    L_{mask} = -1/m^2 \sum_{1\leq i,j \leq m} [ y_{ij} log(\hat{y}_{ij}^k) + (1-y_{ij})log(1-\hat{y}_{ij}^k)]
\end{equation}
where $y_{ij}$ is the label of a cell $(i, j)$ in the true mask for the region of size $m \times m$; $y^k_{ij}$ is the predicted value of the same cell in the mask learned for the ground-truth class k.

\section{Dataset}
The dataset used in this research was recorded at the Santa Fe Firefighting Facility, located in Santa Fe, New Mexico, USA.  Extensive video data were acquired using an IR 5200HD2TIC Camera. This camera is a multipurpose firefighting tool designed for search and rescue and structural firefighting. It uses an uncooled microbolometer vanadium oxide (VOX) detector, which comprises of 320x240 FPA with a pitch of 38um and spatial resolution of 7.5 to 13.5um. This resolution is sufficient to capture necessary features for target detection corresponding to this project.  It records the image with a 320x240 focal plane array sensor. It can record imagery in two different modes, i.e., low and high sensitivity modes. This device also features high score imagery generating 76000 pixels of image detail in low and high sensitivity modes. The dense spectral resolution is (7.5 to 13.5 um), and the output video is in NTSC format with a frame rate of 30 fps and a scene temperature range of 560 degrees Celsius/ 1040 degrees Fahrenheit.  
The recording of the video was performed in a closed and open environment for more than 6 hours. We have more than 400 infrared video files, each 2-3 min long, where each video has the subjects to be detected. There are single and multiple objects of interest to be detected in the same frame.
The objective is to detect humans and targets such as doors, windows, and fire in a firefighting environment. We have used an Alienware Aurora R6 Desktop with configurations of 32GB RAM, Dual GTX 1080 with 16GB GPU memory.  

To create a bounding box of objects and poses, we used a well-known VATIC\cite{vatic} toolbox, which stands for Video Annotation Tool from Irvine, California. This toolset’s output is an XML file that bears information about the bounding box corresponding to all objects of interest present
in the video frames. Also, to generate the masks for semantic segmentation, we used the VGG image annotation toolbox \cite{dutta2016via} developed by the Visual Geometry Group (VGG) at Oxford University. This application saves the annotation labels in an XML file in standard COCO format. In both toolboxes, the objects of interest for annotation were people, doors, and windows. 

\section{Experiments}
For training the Faster RCNN based Mask RCNN, We used 2 GPUs to train the model with two images per GPU.  Training Steps per epoch was set to 1000, whereas for the test, it was set to 50. We used the pre-trained VGG16 on the thermal dataset as our backbone model, and strides of 2,4,8,16 were set for each layer for the FPN pyramid. The fully connected layer was set to a size of 1024.The RPN anchor scales were 8,16,32 and 64 and ration were 0.5,1,2. The RPN Non-maximum suppression threshold was set to 0.7. This threshold was varied to observe the different generated numbers of proposals and filter them as required. Also, the RPN training anchors per image were set to be 256. For the detection, minimum confidence of 0.6 was used, and an NMS threshold of 0.3 was used. We used a learning rate of 0.001, the momentum of 0.9, weight decay regularization of 0.0001. Also, the loss factors i) RPN class loss, ii) RPN bounding box loss, iii) MRCNN class loss, iv) MRCNN bounding box loss, and v) MRCNN mask loss were used. 
Similar configurations were used for Faster RCNN, where the loss functions are instead i) RPN class loss, ii) RPN bounding box loss, iii) FRCNN class loss, iv) FRCNN bounding box loss.

In addition, we also implemented a Simple Online and Realtime Tracking (SORT) \cite{sort} algorithm, which can track multiple objects in real-time by merely associating already detected objects across different frames based on the coordinates of the detection result. SORT takes the output of the off-the-shelf model for object detection and matches detected objects across frames. The SORT algorithm is independent of object labels for tracking and performs mathematical heuristics, such as maximizing the IOU (intersection-over-union) metrics between bounding boxes in neighboring frames.

\begin{figure}[!ht]
\begin{subfigure}[b] {0.9\textwidth}
\begin{center}
   \includegraphics[width=0.9\linewidth]{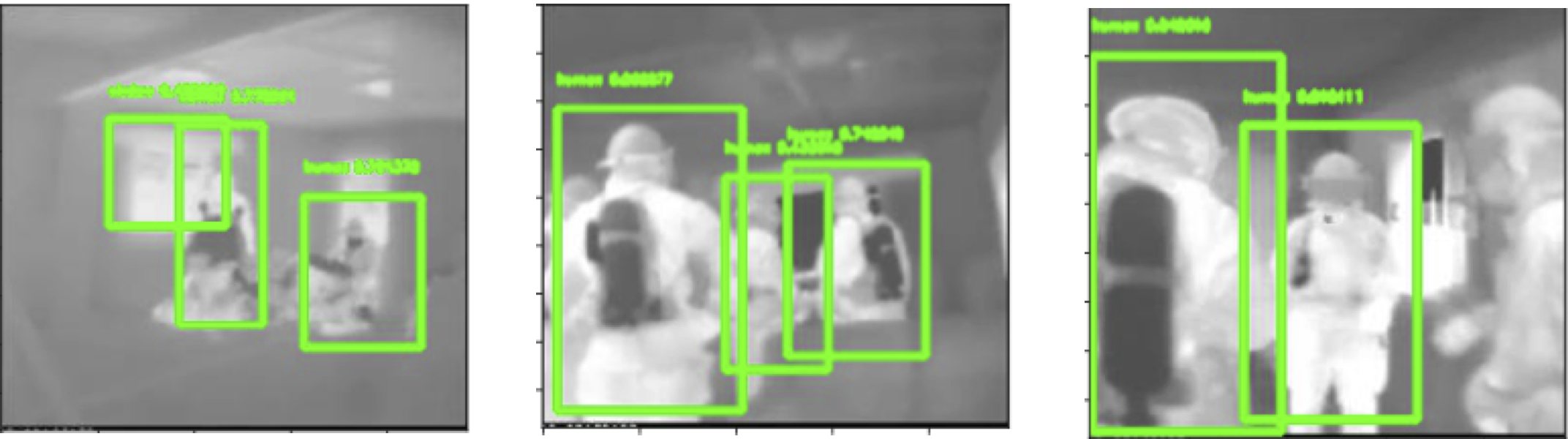}  \end{center}
   \caption{Demonstration of  object detection and tracking implemented with Faster R-CNN. The green boxes bind the object of interest. The label on top infers corresponding object and detection probability. Selected objects of interest include firefighters and civilians, doors, windows, and ladders.}
 
   \label{fig:FasterRCNNTracking} 
\end{subfigure}

\begin{subfigure}[b]{0.9\textwidth}
\begin{center}
   \includegraphics[width=0.9\linewidth]{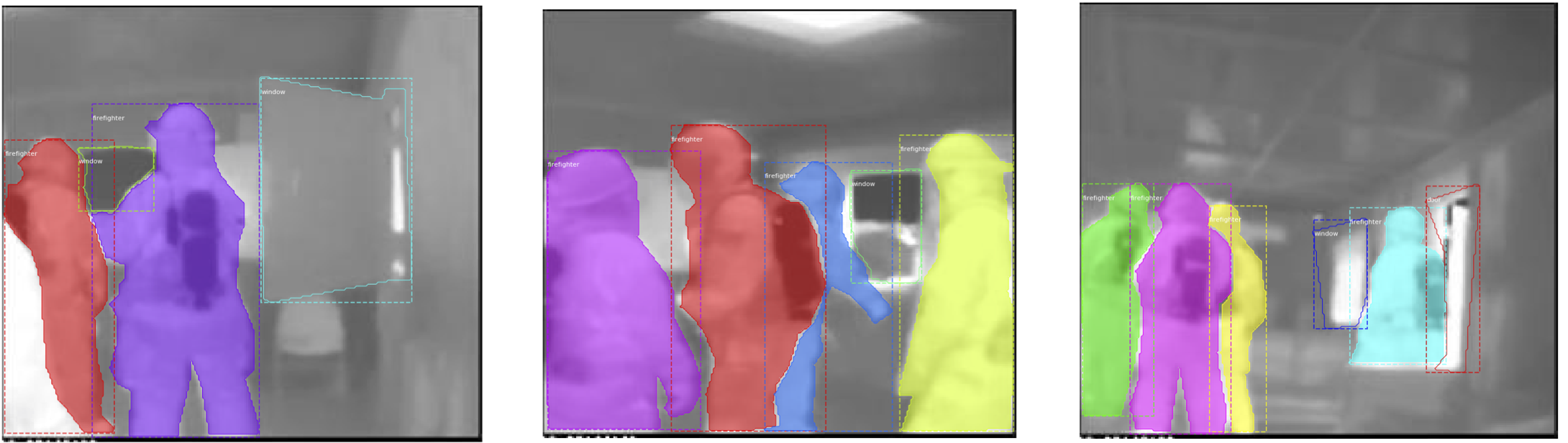}   \end{center}
   \caption{Demonstration of  object instance segmentation implemented with Mask R-CNN. Every instance of significant object is enclosed with different mask . The instances for firefighter are shaded with distinct colors. Doors and windows are shaded with instance boundaries. }

   \label{fig:MaskRCNNTracking}
\end{subfigure}
\caption{Results of the proposed framework } \label{fig:Tracking}
\end{figure}
\section{Results}
To quantify the quality of detection and segmentation, we used the following metrics
\begin{itemize}
    \item \textbf{IoU (Intersection over Union)}: IoU measures the  intersection between the predicted bounding box and actual bounding box divided by their union. A prediction is considered to be True Positive if IoU $>$ threshold, and False Positive if IoU $<$ threshold.
    \item \textbf{Precision and Recall}:

 The recall is the true Positive Rate, i.e., the ratio of True positives predictions to the actual number of positives. Precision is the positive predictive value, i.e., the proportion of true positives to all the positive predictions. 
 
 Recall = $\frac{TP}{TP+FN}$ \& Precision = $\frac{TP}{TP+FP}$
 \item \textbf{mean Avergage Precision(mAP)}
 The mean Avergage Precision(mAP) for object detection is the average of the AP calculated for all the classes. Here we compute the AP for IOU = 0.5 ie. (IoU of BBs need to be $\>$ 0.5)
\end{itemize}

Starting with object detection, we compared our faster RCNN model with other detection models such as Single Shot Detector(SSD) \cite{liu2016ssd} and YOLO \cite{redmon2016you}. Despite their efficacy on RGB datasets due to the right information in the dataset, they did not perform well on the infrared dataset. With YOLO and SSD, we achieved an mAP score of 0.48 and 0.44, respectively. With our proposed framework, we gained an mAP of 0.58. And introducing the SORT algorithm, the detection performance improved to 0.62.

Based on the results, we can infer that the detection and tracking applied to infrared datasets are incredibly challenging as the original image features are too few. They get diminished when passing through CNN. So the classifier and regressor barely receive sufficient feature parameters during training, so the test accuracy is low. It would be better to fuse depth or RGB data information with this infrared dataset, increasing the feature sets in the input image and enhancing the tracking and classification accuracy.

\subsection{Challenges}
The presence of fire causes a target scene’s temperature to vary with time, which might pose restrictions/constraints in the usual identification. There is a need for the algorithm to be invariant to such parameters to give the robust identification of the object under different temperature conditions. These challenges can be overcome by designing an attention-based framework that utilizes the correlation along spatial and temporal dimensions to achieve invariance to temperature changes. The absence of a sufficient database to train the neural network for object recognition and the lack of adequate features to apply standard classification methods are additional challenges for this framework. One possible solution is to incorporate a self-supervised learning framework to learn latent manifold with unlabelled datasets apriori and thus minimize overfitting. 
With advanced techniques such as domain adaptation and domain generalization, we could inherit the feature space information from other thermal datasets such as  FLIR Thermal Dataset for Algorithm Training, an open-source RGB-thermal driving dataset. One of the great illustrations of domain generalization for diagrams recognition and retrieval are shown in \cite{bhattarai2020diagram} and \cite{castorena2020learning}. 

\chapter{Scene Description through Caption Generation in Firefighting Environments}
\section{Motivation} 

The scenes of active fire can be characterized by abysmal visibility due to smoke and debris, making them difficult to navigate. This often leads to a lack of visibility and impacts firefighters’ situational awareness. One possible remedy to address this issue is to keep the firefighters informed about the scene and getting them prepared in extreme conditions. This supplemental situational awareness can be achieved by an intelligent scene description system that combines object recognition and a language model to interpret the scene. Despite the firefighters’ lack of visibility, the system can describe the scene via audio feedback like SIRI or ALEXA. Such tools can help them in decision making by providing necessary information when the firefighter cannot interpret such information directly with their senses and do not have any direct contact with or instruction from the commander.
\par
Deep Convolutional neural nets have demonstrated the state of 
 art performance in different computer vision applications such as classification, prediction, tracking, segmentation, synthesis, etc. These CNNs can be further combined with another modality such as graphs, language models, physics models, etc., to learn a joint representation to accomplish a complicated objective. Such multimodal fusion aims to align the embedding space across different data domains in feature space. One such multimodal fusion application can be a scene description where the objective is to combine a CNN model with a language model, where the visual and word features can be learned jointly as a probability function, and this knowledge can be learned in the form of the two neural networks weight parameters. For the language model, a variant of Recurrent Neural Network(RNN) called Long-Short Term Memory(LSTM) addresses the vanishing and exploding gradient problems. The widely used models for caption generation are usually an encoder-decoder model where the encoder is a
CNN model that generates features corresponding to the input image and the decoder is the language LSTM model that inputs the visual features and generates output sequence predicting a word at a time. Furthermore, an attention model is added as an extension of the encoder-decoder modality to visualize the correspondence of generated captions with the input image. This gives an understanding of the visual regions that the neural net sees for describing the scene.
The contributions are:
\begin{itemize}
    \item We construct a first scene description dataset for a firefighting scenario, where each infrared image is described with 3-4 sentences.
    \item We demonstrate a first caption generation module combining a CNN and LSTM model for scene description in a firefighting scenario.
\end{itemize}
Further, An attention module can also be introduced, which aids in visualizing the image regions contributing to scene description. The saliency map thus generated allows us to understand infrared imagery capabilities for other complex tasks.

\section{System Description}
Our proposed scene description framework is based on the feature extraction framework utilized in chapters 2 and 3. The deep learning-based architecture can recognize the image and subsequently describe the image’s aggregate details for captioning.  The objective is to learn a single joint model that takes image I as input and train to maximize the likelihood $p(S|I)$ where S is the target sequence of words $S={S_1, S_2,…..S_N}$ for each $S_k \in V$ where V is the dictionary or vocabulary which is the set of words used for describing the image adequately. Recurrent Neural Nets(RNNs) first evolved as a translation model widely used for language translation and text to speech/speech to text converters. These models were based on an encoder-decoder structure where the encoder receives the source sequence, and the decoder translates to the target sequence. The encoder-decoder model based on RNN was able to learn a joint embedding space where mapping was achieved between source and target word features in the form of vector representation.  LSTMs were later used as a replacement for RNN to address the vanishing and exploding gradient problem. The state of art sequence to sequence translation is based on transformer models, which use attention models to enrich the representations. Here, the objective is to replace the language encoder with a visual encoder, which extracts the visual semantic information from an image to a  fixed-length vector. This CNN framework has been pre-trained for image classification/ object recognition tasks. The CNN framework is then pruned by trimming the last prediction layers, and only the fixed-length feature vectors are used as encoder outputs. The last layer of this CNN is then fed as input to the LSTM decoder to generate the neural image caption model sequence. \par
Considering $\theta$ as the parameters of the framework, for an image $I$ and target description $S$, the objective is to maximize the probability of correct description given as 
\begin{equation}
\theta^* = argmax_{\theta} \sum_{(I,S)} log p(S|I;\theta) 
\end{equation}

Where, $log p(S|I;\theta) = \sum_{t=0}^N log p(S_t|I, S_0,…,S_{t-1};\theta)$ based on chain rule on the joint probability over $S_0,….S_N$. 
 The probability $ p(S_t|I, S_0,…,S_{t-1};\theta)$ is modeled with an RNN as the current predication and is based on a history of words.  The variant of RNN called LSTM is used to estimate the probability distribution. 


\clearpage
\begin{figure}[!ht]
\centering
\begin{subfigure}[b]{0.5\linewidth}
	\includegraphics[width=\linewidth]{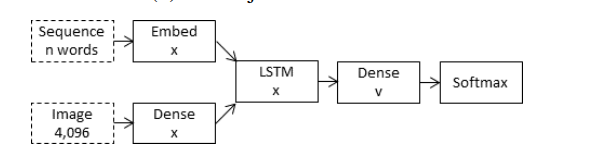}
	\caption{i.}
\end{subfigure}
\begin{subfigure}[b]{0.5\linewidth}
	\includegraphics[width=\linewidth]{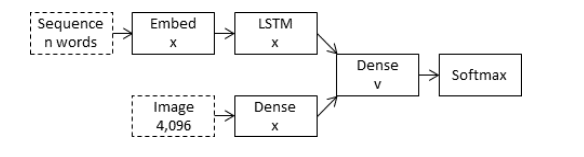}
	\caption{ii.}
\end{subfigure}
\caption{i) Inject and ii) Merge Captioning framework}
\label{fig:captions_frame}
\end{figure}
The caption generation process can be achieved through either of two approaches 1) Merge and 2) Inject approach as shown in \ref{fig:captions_frame}. We have tried both the approaches for this project and have found the merge architecture provides the better performance.
\begin{figure}[!ht]
    \centering
    \includegraphics[width=.8\linewidth]{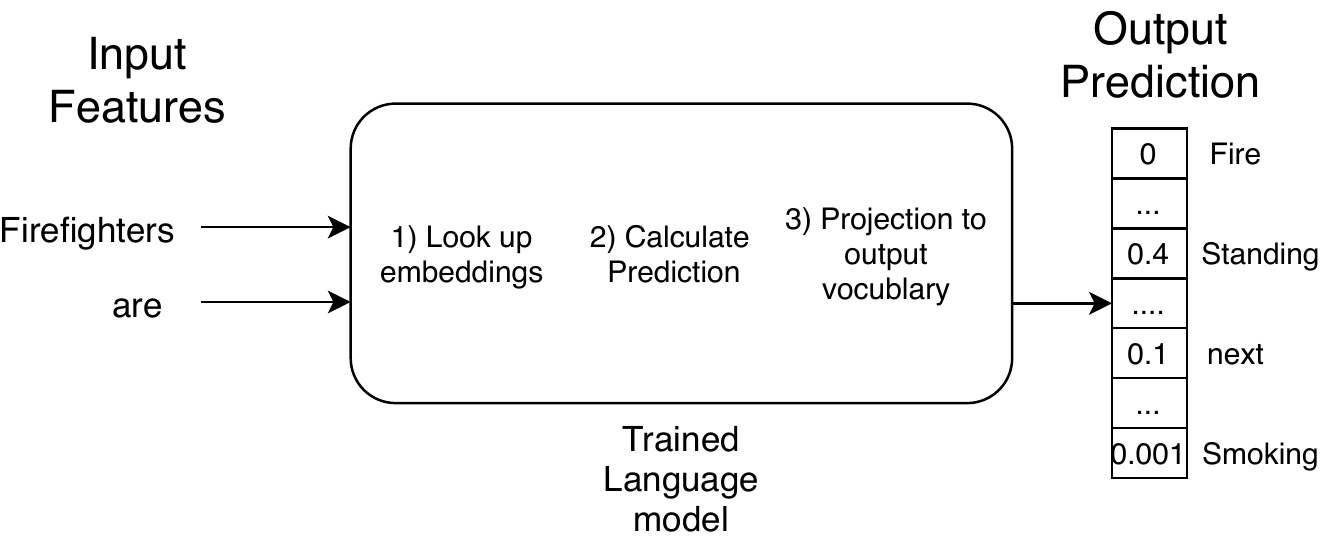}
    \caption{Illustration of word prediciton}
    \label{fig:predict_model}
\end{figure}

\begin{figure}[!ht]
	\centering
	\includegraphics[width=.8\textwidth]{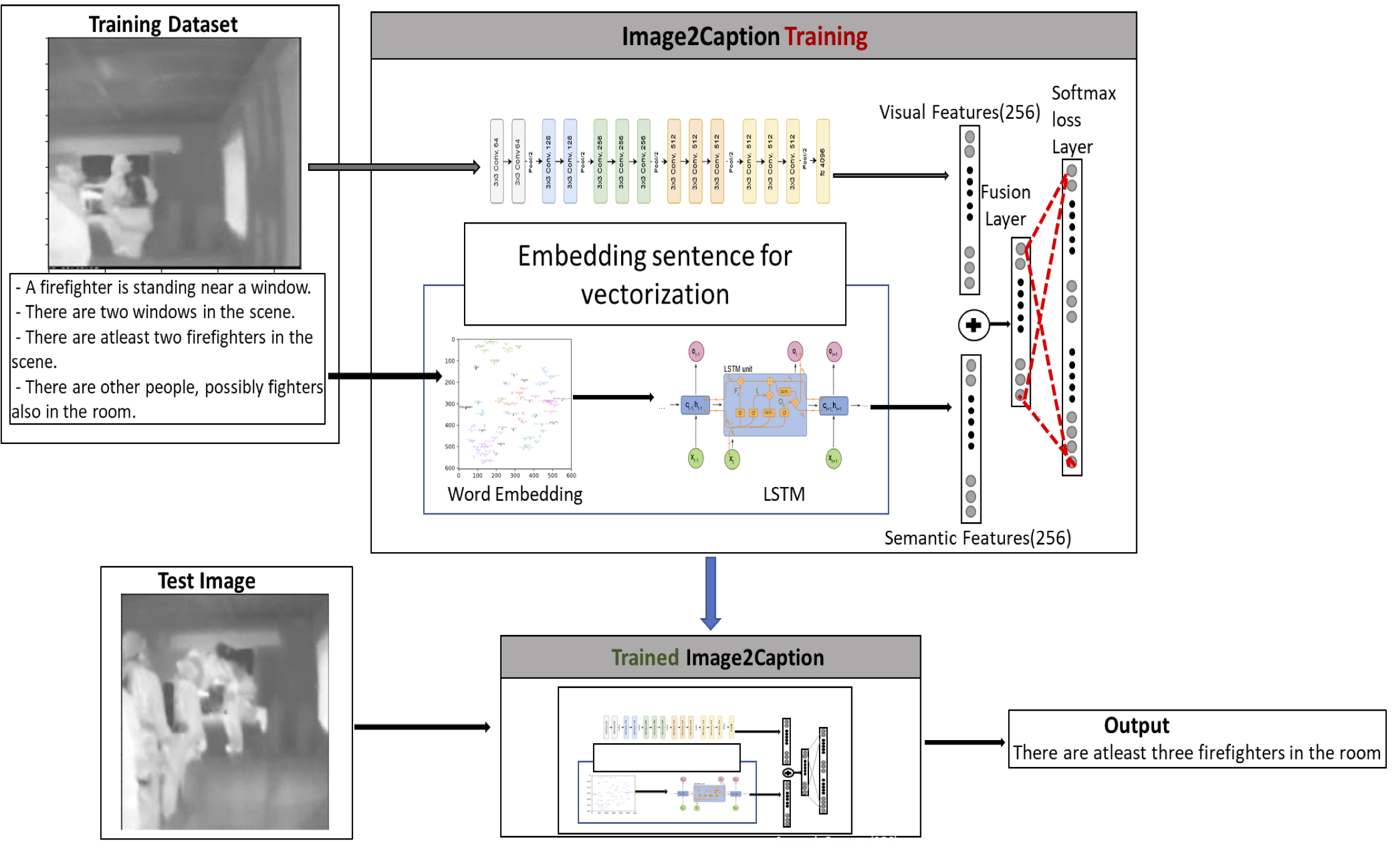}
	\caption{The system comprises a CNN block for visual feature extraction. The inputs of the CNN block are image datasets. The system has a subsystem for embedding scene descriptor sentences for vectorization. The subsystem performs the word2vec embedding which maps input word sequences into equivalent numeric vectors. The vectors are then fed to a variant of Recurrent Neural Net(RNN) called Long Short Term Memory(LSTM). LSTM learns the sentence structure from the caption dataset during training. During the testing phase, it is able to generate the vectors corresponding to the most likely sequence of words which is added with visual description features and then fed to a Neural network to output the probability of correctness associated with the caption, }
	\label{fig:sceneunderstanding}
\end{figure}
We constructed a basic neural language model consisting of a word embedding matrix, a basic LSTM (Hochreiter and Schmidhuber, 1997), and a softmax layer. The LSTM is defined as follows: 
\begin{gather*}
\begin{aligned}
   i_n = \sigma(x_n W_{xi} + s_{n−1} W_{si} + b_i) \\
 f_n = \sigma(x_n W_{xf} + s_{n−1} W_{sf} + b_f )\\
  o_n = \sigma(x_n W_{xo} + s_{n−1} W_{so} + b_o) \\
 g_n = \tanh(x_n W_{xc} + s_{n−1} W_{sc} + b_c)\\
 c_n = f_n  \circ c_{n−1} + i_n \circ g_n\\
  s_n = o_n \circ tanh(c_n) 
\end{aligned}
\end{gather*}

where $x_n$ is the $n_{th}$ input, $s_n$ is the hidden state after n inputs, $s_0$ is the all-zeros vector, $c_n$ is the cell state after n inputs, $c_0$ is the all-zeros vector, $i_n$ is the input gate after n inputs, $f_n$ is the forget gate after n inputs, $o_n$ is the output gate after n inputs, $g_n$ is the modified input used to calculate $c_n$ after n inputs, $W_{\alpha \beta}$ is the weight matrix between $\alpha$ and $\beta$, $b_{\alpha}$ is the bias vector for $\alpha$,  $\circ$ is the element-wise vector multiplication operator, and $\sigma$ refers to the sigmoid function. The hidden state and the cell state always have the same size. 
\par
Word embedding is mapping words into vectors of real numbers using the neural network, probabilistic model, or dimension reduction on a word co-occurrence matrix. Various word embedding techniques are widely used where the major ones are word2vec from Google, Glove from Stanford, and fastest by Facebook. To categorize similar words together, word embedding can establish a proper cluster of word semantics. Word embedding enables document clustering, text classification, sequence to sequence translation, and various other NLP tasks. In word2vec, the words are represented in the form of vectors with one-shot encoding representation. The words with similar semantic contexts tend to appear closer than the ones with different contexts separated apart. Previously bag of words(BOG) was one of the popular word embedding methods, which incorporates a  sparse vector representation where the dimension is equal to the vocabulary size. Due to the lack of ability to attribute the context of words, BOG standalone is not used as a word embedding method. To address that, Word2Vec was introduced as it learns word context by predicting the surrounding context. 
Word2vec architecture uses two techniques -CBOW(Continuous bag of words) and the Skip-gram model. CBOW predicts a word’s probability given a context where the current word is predicted using a window of surrounding context windows. Skip-Gram operates inversely to CBOW and predicts the given sequence or context from the word.

\section{Experimental Details}
Since the language model plays a vital role in the scene description, it is essential to set up the experiment and the network model parameters for the training and validation. First, based on the training description set, we construct a vocabulary by considering all the unique words using the description. Before training, two matrices are first created - an Embedding matrix and a Context matrix, where these two matrices have an embedding for each word in the vocabulary. Both of the matrices have a size of vocabulary size times the embedding size. These matrices are initialized randomly at the beginning of the training process and learn them through neural net training. Cosine similarity is taken to measure the similarity between the input embedding with each of the context embeddings, followed by a softmax to estimate the probability associated with the prediction. The hyperparameters used for training the framework were the window size and number of negative samples. The Window size is the size of context vectors that contribute to the next prediction, and negative samples are the samples of words that are not neighbors in the embedding space. 

In the experiments, this basic neural language model is used as a part of two different architectures: 
\par
	In the inject architecture, the image vector is concatenated with each of the word vectors in a caption. In the merge LSTM architecture, it is only concatenated with the final LSTM state. The layer sizes of the embedding, LSTM state, and projected image vector were also varied in the experiments to measure the effect of increasing the networks' capacity. The layer sizes used are 128, 256, and 512. The details of the architectures used in the experiments are illustrated in Figure 3. The training was performed using the Adam optimization algorithm (Kingma and Ba, 2014) with default hyperparameters and a minibatch size of 50 captions. The cost function used was sum cross-entropy. The training was carried out with an early stopping criterion, which terminated training as soon as the validation data’s performance deteriorated (validation performance is measured after each training epoch). Initialization of weights was done using Xavier initialization (Glorot and Bengio, 2010), and biases were set to zero. Each architecture was trained three separate times; the results reported below average over these three different runs.

In our merge structure, it consists of two different deep learning substructures. The upper structure is a CNN that has been previously trained with the training thermal images described before, and the corresponding targets were the labels of the objects in the scene. We use a  VGG16 framework for feature extraction with the last classification/softmax layer stripped off to produce 4096 features. We precompute the visual features with this model and save them to 4096 array vectors. This reduces memory and computational burden on the training system. 

The lower structure is an LSTM network trained with the sentences previously transformed by a word embedding. Prior to the embedding, a standard preprocessing is applied to the descriptions, including stop word and words with numbers removal, conversion to lowercase, and punctuation removal. A dictionary of $4000$ words is used that assigns an index to each possible word among a given set. Then, the description sequence is coded with these indices before its introduction in the LSTM. The structure shown in Figure \ref{fig:sceneunderstanding} contains 256 memory units. They constitute the semantic features. During the training, the descriptions are fed with a sequence of words sequentially, and the corresponding target is the next word of the given sequence. The sequences are started with a “start sequence” code, and they finish with an “end sequence” code. 

The visual and semantic features are added together to form a vector of 256 mixed features that input a fully connected neural network with two layers. The output layer has as many nodes as possible words (4000), and their activation is a standard softmax. It estimates the probability of each word of the dictionary given the input sequence and image. During the test, the predicted words are fed back into the LSTM structure.

To generate the captions for image captioning, three individuals were assigned the task to provide at-least three captions for each image. Then the aggregated captions from each labeler were used to train the image captioning framework. 

\section{Dataset}
The data used for the architecture training consists of 1000 images that were first labeled with the objects in them and then with four different short sentences describing various aspects of the scenario’s interest. Volunteers have freely carried out the descriptions. For example, in the image shown in Figure 
\ref{fig:sceneunderstanding}, two of the sentences are “there are two windows in the scene” or “a firefighter is standing next to a window.” For all our experiments, we used a fixed vocabulary size of 500. The dataset was split into train-valid-test splits in the ratio of 60:20:20. To avoid biased performance, the partitions were sampled 20 times, and the average performance is reported. 

\section{Results}

We report results with the frequently used BLEU \cite{papineni2002bleu}  metric, which is the standard in the caption generation literature.  BLEU stands for bilingual evaluation understudy. It is an algorithm for evaluating the text’s quality, which has been machine-translated from one natural language to another to that of humans. BLEU is language independent, easy to understand, and easy to compute. The standard BLEU ranges [0,1], where higher values correspond to a better quality caption. Individual text segments are compared with a set of referenced texts, and scores are computed for each of them. In estimating the overall quality of the generated text, the computed scores are averaged. The approach works by counting matching n-grams in the candidate translation to n-grams in the reference text, where 1-gram or unigram would be each token, and a bigram comparison would be each word pair. Here larger n results in lower performance as the chances for match decrease with an increase in the word count. 
\clearpage
\begin{figure}[!ht]
	\centering
	\includegraphics[width=.8\textwidth]{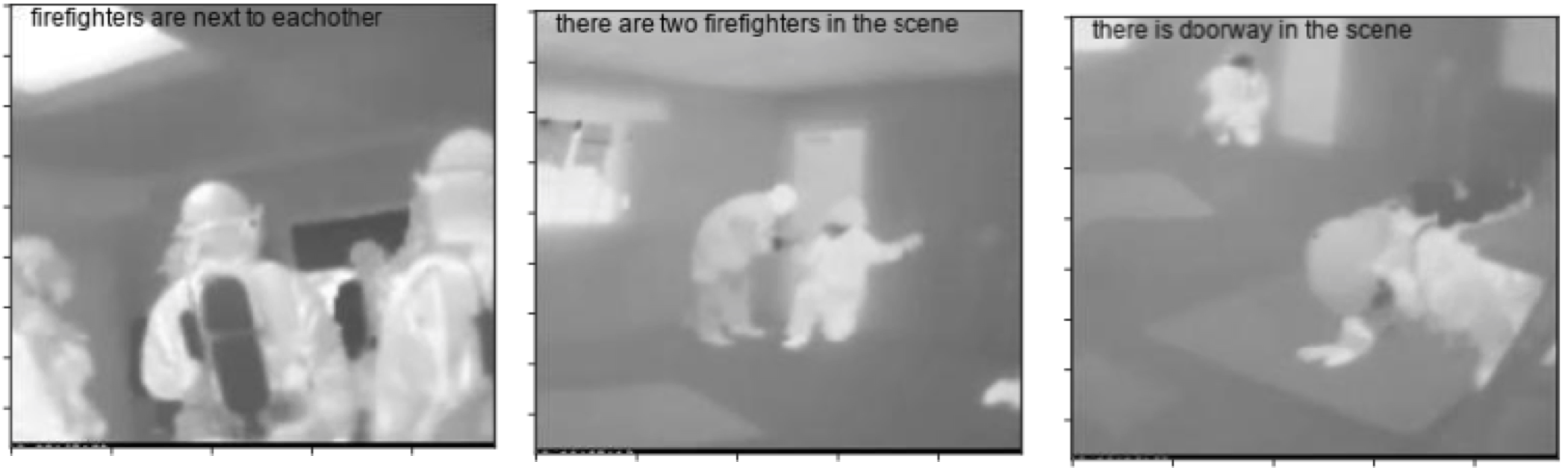}
	\caption{Demonstration of image caption system. Every scene is described by a relevant sentence, as shown in the images. }
	\label{fig:Imagecaption}
\end{figure}

\par
We achieved a maximum BLEU score of 80 for 1-gram, 68 for 2-gram, 52 for 3-gram, and 45 for 4-gram. 

\subsection{Drawbacks}
The proposed method is based on a holistic approach where the captioning is done based on a global image without considering any semantic relationship between objects in the scene. So, suppose the captioning system is designed based on the object detection and segmentation framework instead of the classification framework. In that case, the captioning system can encode better semantic relationships and describe each object/region of the image separately, generating a detailed description of the scene. 

\chapter{Embedded Deep Learning Implementation}
\section{Motivation}

Firefighting is an inherently dangerous and potentially fatal activity. Even the most experienced firefighters can be affected by stress and anxiety during a fire, leading to disorientation and the corresponding impairment of situational awareness. Furthermore, these stressors are only exacerbated by the smoke, high temperatures, and low visibility at the scene, further compromising a firefighter's ability to respond to the situation effectively. Situational awareness, or the real-time knowledge of continuously changing conditions at the scene, is essential to accurate decision-making. A firefighter's ability to maintain situational awareness in these conditions is likewise critical.

The procedure for object and human detection can be used to describe
the scene by using human language, prioritize the presentation of images to the
commander depending on the contents of the image, or to find different scenarios
on demand (by presenting those scenes that contain the fire, procumbent humans, or other
situations of interest). Another interesting application is to produce augmented
the reality for the firefighter in the scene. The situations of interest where this implementation can be applied are diverse. For instance, firefighters may find themselves in a scenario where their vision is impaired due to dense smoke or dust particles or where there is no visible light. Hence, the scene has to be illuminated with a handheld or helmet flashlight. In general, in these situations, thermal cameras can help improve the visual clarity of the scenario. Nevertheless, thermal images are more difficult to interpret than visible light-based images. A way to enhance the firefighters' perceptual experience is to use the detected images to reconstruct the scenario using augmented reality glasses. We have developed preliminary prototypes of an
augmented reality system that projects the images detected and segmented by a
neural network to the Hololens augmented reality glasses, commercialized by Microsoft. They project the image from a computer
over a semitransparent screen to see the projected images and through them. The images are
projected, so their virtual focal point is about 10 inches away from the user's eyes,
thus producing the illusion of seeing a semitransparent screen floating over the
user. Their see-through effect allows the user to interact with the environment without being impeded or disoriented by a virtual experience.
The headset has an incorporated inertial measurement unit (IMU) used to
determine the head's position. This information is useful to maintain the position of the image when the user rotates their head.

This work aims to demonstrate the hardware deployment of the firefighter aid system. This proposed system integrates FLIR One G2 and Intel RealSense D435i thermal and depth cameras with an NVIDIA Jetson embedded Graphics Processing Unit (GPU) platform that deploys already developed machine and deep learning models \cite{bhattarai2020deep} to process captured datasets and then wirelessly stream the augmented images in real-time to a secure local network router based on a mesh node communication topology \cite{hamke2019mesh}. The augmented images are then relayed to a Microsoft HoloLens augmented reality platform incorporated into a firefighter's PPE as part of a system that is insusceptible to environmental stressors while also being able to detect objects that can affect safe navigation through a fire environment with a greater than 98\% test accuracy rate \cite{bhattarai2020deep}. This relay, interpretation process, and return of interpreted and projected data is real-time.  \par
\begin{figure}[!ht]
\centering
\includegraphics[scale=0.8]{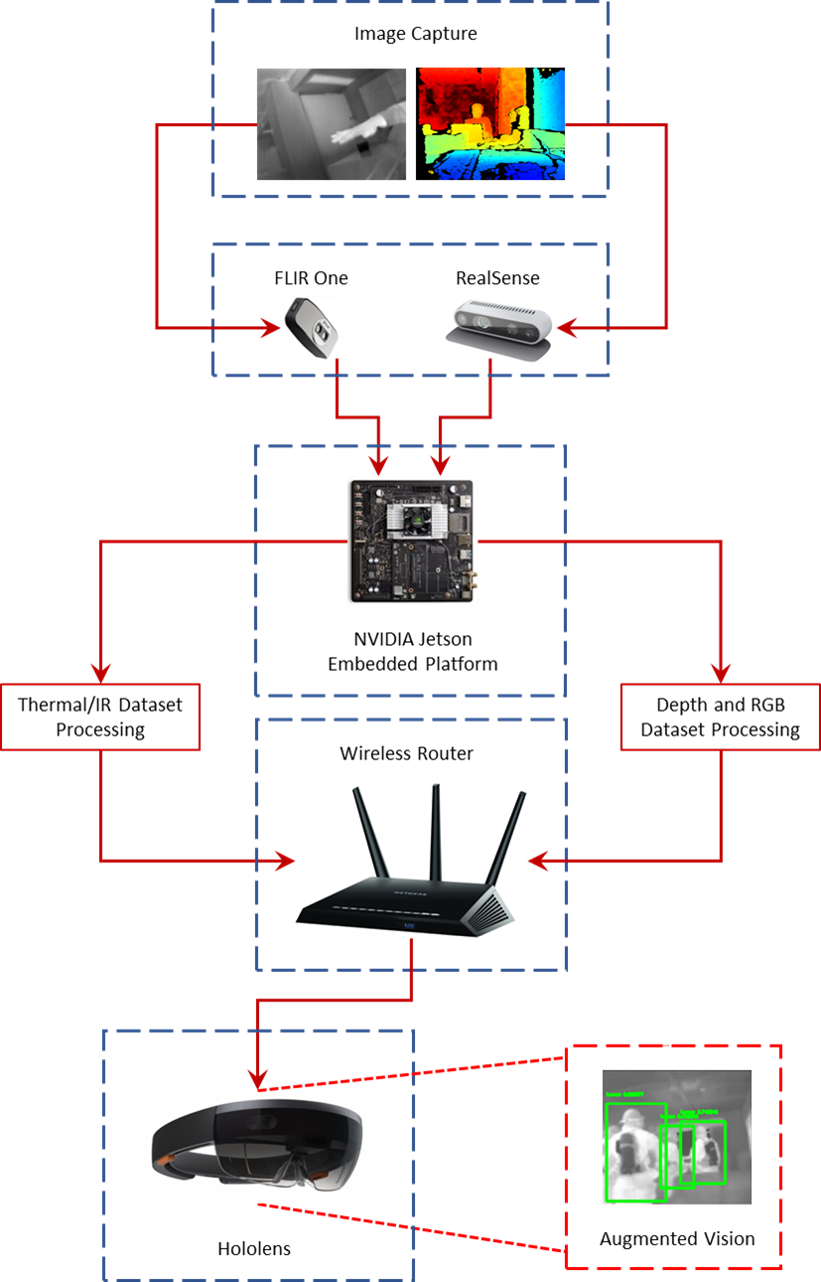}
\caption{Overview of relationships between system components}
\label{overview}
\end{figure}
The contributions to the field delivered by this work are targeted towards application-oriented research. We present a system that demonstrates the effective combination of existing technologies to cyber-human systems applied to firefighters to improve on-duty firefighter safety. From a technical point of view, our contribution consists of designing a unified tracking and segmentation framework for infrared imagery and performing an online computation in an embedded platform. \par

\begin{figure}[!ht]
\centering
\includegraphics[width=.98\linewidth]{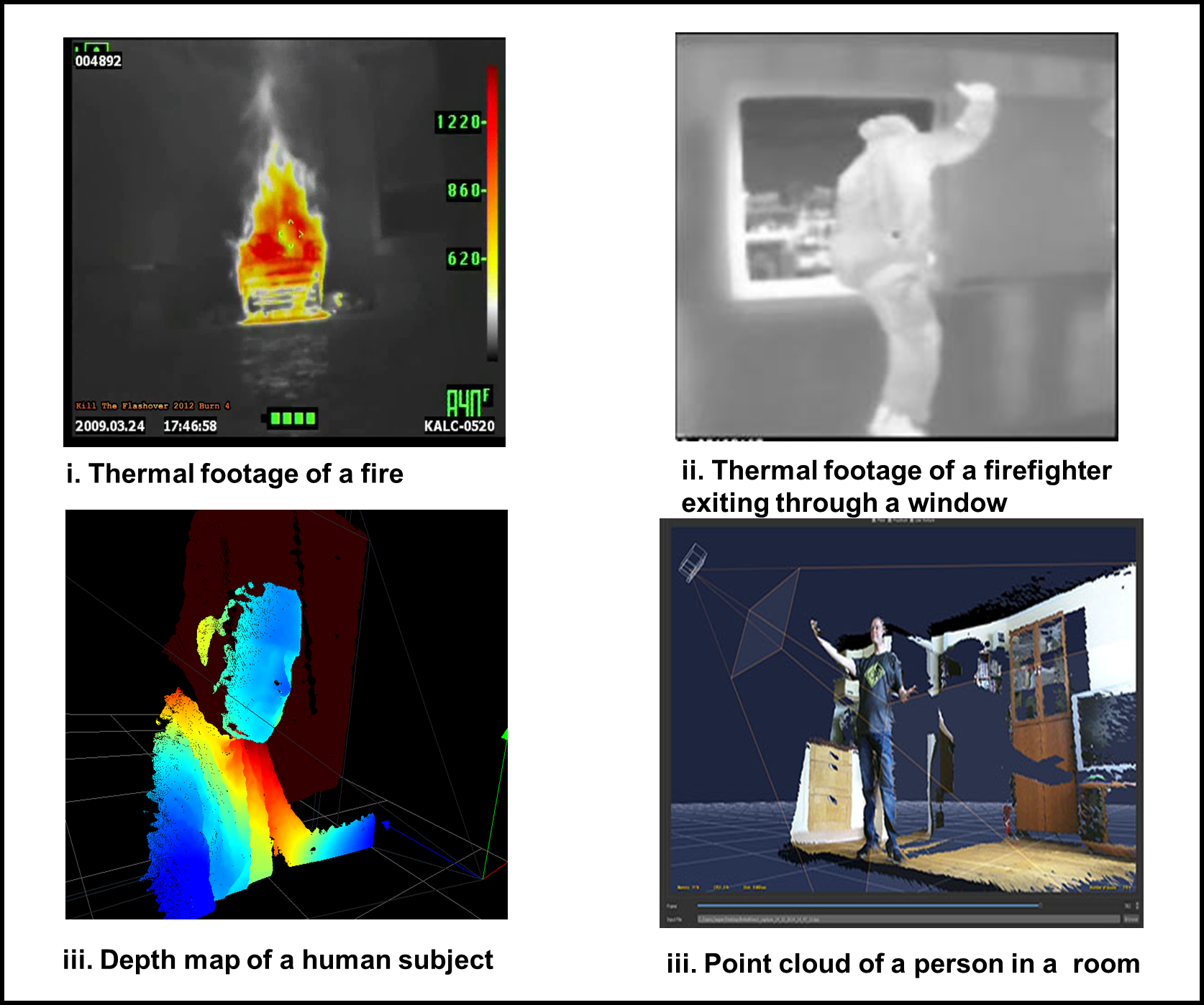}
\caption{Types of datasets acquired by different cameras}
\label{datasets}
\end{figure}

\section{ Deep CNN Model}
This research deploys a trained deep Convolutional Neural Network (CNN) based autonomous system to identify and track objects of interest from thermal images captured at the fireground.
While most applications of CNN systems to date are found in the related fields of surveillance and defense, some recent work has explored applications in unmanned aerial vehicles (UAVs) to detect pedestrians \cite{de2018using} and avalanche victims \cite{rodin2018object} from the air. Other recent work has demonstrated the use of CNN systems on embedded NVIDIA Jetson platforms, such as the TX1 and TX2, for traffic sign identification (with applications for autonomous vehicles) \cite{han2017traffic} and autonomous image enhancement \cite{nokovic2019image}. Other work has deployed fully-Convolutional Neural Networks (FCNNs) on the TX2 to provide single-camera real-time, vision-based depth reconstruction \cite{bokovoy2019real} and demonstrated an attention-based CNN model that can learn to identify and describe image content that can then be generated as image captions \cite{xu2015show}. \par

\clearpage

\begin{figure}[!ht]
    \begin{subfigure}[b]{0.5\linewidth}
        \centering\includegraphics[width=\linewidth]{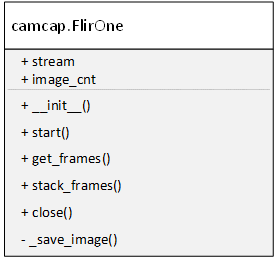}
        \caption{FlirOne class}\label{fig:flir_uml}
    \end{subfigure}%
    \begin{subfigure}[b]{0.5\linewidth}
        \centering\includegraphics[width=\linewidth]{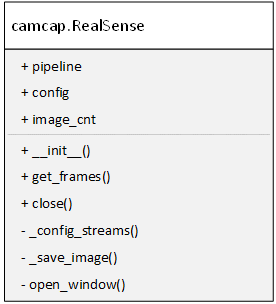}
        \caption{RealSense class}\label{fig:rs_uml}
    \end{subfigure}
\caption{UML class diagrams}\label{fig:uml}
\end{figure}


The embedded system described in this work deploys a deep CNN system but is a novel approach that, to the best of our knowledge, is the first of its kind \cite{bhattarai2020deep}. To implement the augmented reality system, we constructed a lightweight  Mask R-CNN~\cite{he2017mask} with a Faster R-CNN~\cite{girshick2015fast}  backbone, as shown in Figure~\ref{fig:structure}. Objects are identified and tracked in images from the thermal imaging dataset using Faster R-CNN, and then instances of significant objects are masked and shaded using Mask R-CNN. This framework is built on top of a VGG-16 like model\cite{bhattarai2020deep}.  We extended the VGG model framework with proposal networks to construct a  region-based CNNs (RCNNs) to accomplish the object segmentation and tracking tasks. These models can localize each one of the objects in the scene and continuously track them. The model was further modified with a 1x1 Convolutional layer constructing a Mask R-CNN to achieve better detection and produce, at the same time, semantic segmentation of the scene. The system was trained with many images recorded in real fire training situations by firefighters to identify and track objects in real-time. 
\par
  The detection and segmentation framework is based on the Faster R-CNN backbone. The detection and tracking process involves region proposal generation and classification process. Faster R-CNN uses a region-proposal network(RPN) at the end of the feature extraction network to process the feature maps generated and produce region proposals used by classification and regression heads for bounding box and label generation.
The region proposal network is a fully convolutional network which operates on the extracted features with different anchor boxes generating scores. Anchor boxes are a sliding window with a variable size that act on CNN feature maps. For each window, output k potential bounding boxes and scores measure the amount of accuracy of each box, i.e., bounding box coordinates and score indicating how likely the image in the given bounding box will be an object of interest. These bounding boxes are further processed by the classification heads and regression heads to generate classification and tighten the bounding boxes. 
Mask R-CNN is an extension to the Faster R-CNN for pixel-level segmentation. Faster R-CNN can only produce the bounding boxes for each object and might be confusing when the objects of interest are highly overlapped in an image. Mask R-CNN addresses this issue by locating the exact pixels of each object. Mask R-CNN adds a fully convolutional network (FCN) branch to Faster R-CNN that processes the CNN feature map and outputs a binary mask indicating whether the pixel in question belongs to the object or not.   Mask R-CNN uses RoIPool to generate the masks,  
combined with classification and bounding boxes, from faster R-CNN to generate precise segmentation maps.  The training loss comprises classification error, bounding box regression error, and mask generation error.
\par
Furthermore, the backbone VGG16-based system \cite{bhattarai2020deep} is capable of performing human recognition and posture detection to deduce a victim's condition and guide firefighters accordingly in prioritizing victims by urgency. The processed information can be relayed back to the firefighter via an augmented reality platform that enables them to visualize the analyzed inputs' results and draws their attention to objects of interest, such as doors and windows otherwise invisible through smoke and flames. This visualization also provides localized information related to those objects.
\section{ Embedded System}

\subsection{System Overview}
The embedded system consists of the following commercial off-the-shelf (COTS) hardware components:
\vspace{\baselineskip}
\begin{itemize}
\item{NVIDIA Jetson TX2 Development Kit}
\item{FLIR One G2 for Android thermal camera}
\item{Intel RealSense D435i depth camera}
\item{Wireless network router}
\end{itemize}
\vspace{\baselineskip}
An overview of the interactions between these components and their place within the smart and connected firefighting system is illustrated in Figure~\ref{overview}. The components can be further categorized into those that acquire data (FLIR One and RealSense cameras), those that process data (NVIDIA Jetson), those that communicate data (NVIDIA Jetson and wireless router), and those that ultimately augment data (HoloLens). Each of these categories is described here in detail.


\subsection{System Architecture}
\subsubsection{Data Acquisition}
The system acquires data utilizing the live streaming of thermal, RGB, and depth image datasets captured by both the FLIR One and Intel RealSense cameras. Examples of such datasets are shown in Figure~\ref{datasets}. Each camera can adjust the frame rate, but for consistency, all tests were conducted at a rate of 30 frames per second (fps). Both cameras are attached to Jetson platforms via USB cables, and the acquired datasets are transmitted from the camera sensors to the embedded Jetson platforms through the same USB connections.

\subsubsection{Data Processing}
The camera-acquired datasets, including thermal, RGB, and depth images, are processed by the deep learning models deployed on the Jetson GPU platforms. The algorithms process each captured frame, and the rendered images are then streamed to a secure wireless network router, along with the original raw images. The images are also saved locally on the Jetson's onboard flash memory; available storage on the embedded platform is limited; however, stored images are only stored until a limit is reached (up to 100 captured frames). At this point, the saved frames are removed, and a new stored set begins.

The specific information processed by the deep learning model is object detection, object tracking, instance segmentation, and scene description.

\subsubsection{Data Communication}
The processed and raw images are wrapped in HTML code and made available for access by connected devices on the secure local network through a wireless router as a video stream. This allows access to both the firefighter whose sensors captured the data and commanding officers overseeing and guiding the response effort and any other firefighters at the scene.

\subsubsection{Data Augmentation}
The processed data is relayed in real-time through the network for visualization on the Microsoft HoloLens, thus providing the firefighter with the visualized augmented information to assist navigation and enhance situational awareness.

\subsection{System Implementation}
\subsubsection*{Camera Sensor Implementation}
Both the FLIR One and the RealSense D435i cameras are available commercially; however, neither camera was designed for the ARM64 architecture of the NVIDIA Jetson platforms, which run a version of embedded Ubuntu Linux OS developed specifically for the Jetson's Tegra processor (a system on a chip, or SoC). Therefore, it was first necessary for the system to recognize and capture data from the USB connected cameras to build each camera its kernel driver module for the Tegra Ubuntu kernel. These modules are variations of the open-source v4l2loopback Ubuntu Linux kernel module, modified for the individual camera.

\begin{figure}[!ht]
\centering
\includegraphics[width=.98\linewidth]{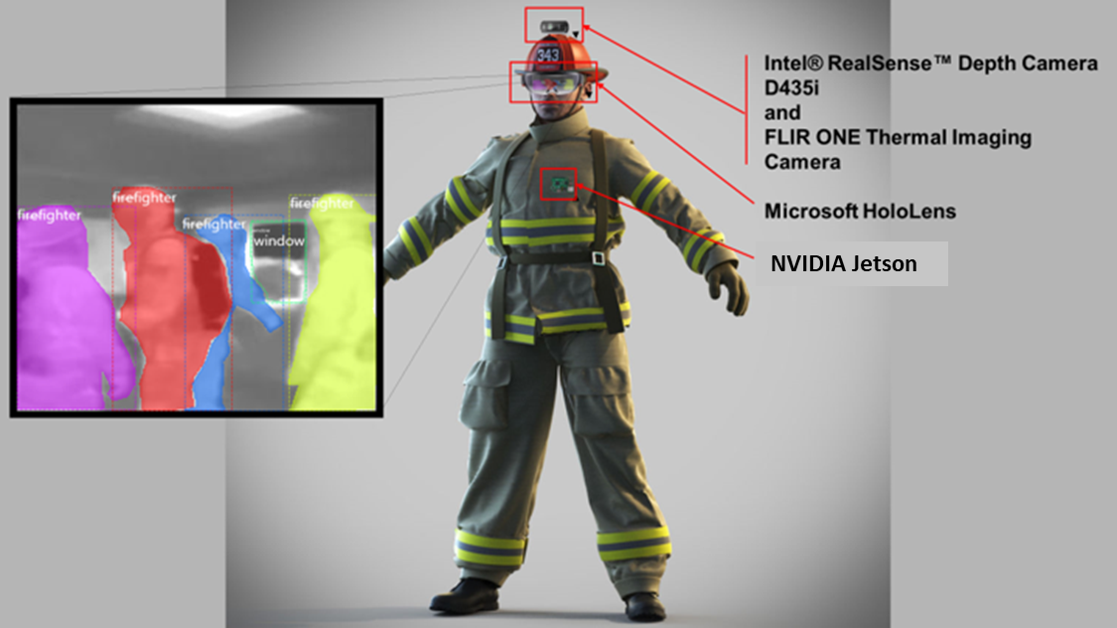}
\caption{Embedded system deployment}
\label{deployment}
\end{figure}

The cameras are controlled at the user level by the Python language wrapper programs for the FLIR One and Realsense cameras. These programs implement the FLIR One and RealSense classes, where each class provides the necessary user-level controls to initialize the cameras and acquire and save frames. These individual camera wrapper programs also initialize the cameras and implement OpenCV to process the individual captured frames with the trained deep CNN system using the Jetson's onboard GPU.

\begin{figure}
\centering
\includegraphics[width=.98\linewidth]{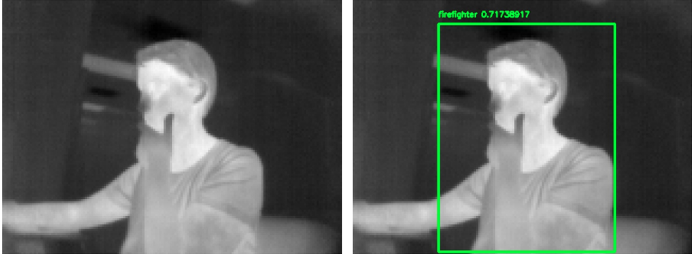}
\caption{FlirOne video feed raw and processed output}
\label{fig:flir_feed}
\end{figure}

\begin{figure}
\centering
\includegraphics[width=.98\linewidth]{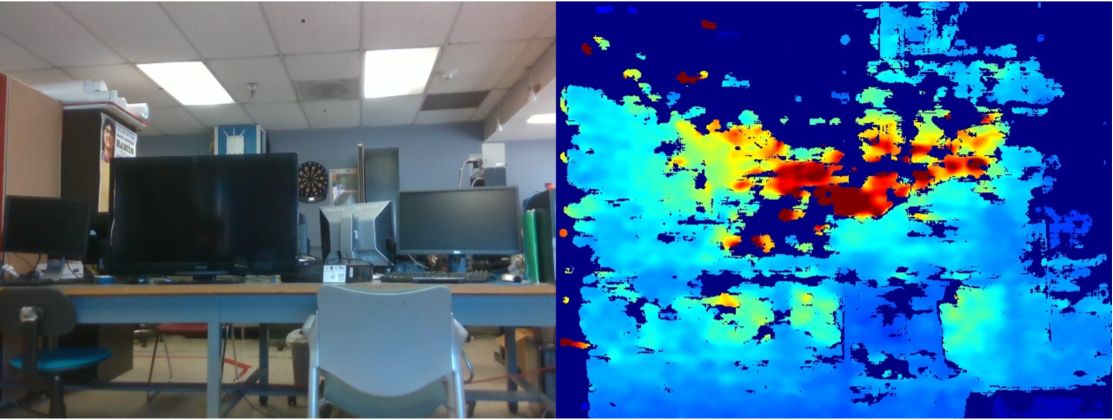}
\caption{RealSense RGB and depth colormap video feed output}
\label{fig:rs_feed}
\end{figure}
\subsection{System Deployment}

The data acquisition, data processing, and augmented data hardware components of the embedded system seen in Figure ~\ref{overview} will be integrated with the firefighter's personal protective equipment (PPE) as shown in Figure \ref{deployment} to provide the firefighter the real-time information processing and augmentation that will enhance situational awareness and assist in minimizing the number of poor decisions made due to stress, anxiety, and disorientation. The system is lightweight and can easily be integrated with the firefighter suit. 

\section{Discussion and Results}

 The resulting live stream output comprising both raw and processed thermal images is illustrated in Figure~\ref{fig:flir_feed}. Similarly, the live stream output comprising the RGB color and color mapped depth images captured by the RealSense camera is presented in Figure~\ref{fig:rs_feed}. The thermal captured feed algorithm results in firefighting scenarios are shown in Figure~\ref{fig:FasterRCNNTracking} and Figure~\ref{fig:MaskRCNNTracking}.
 
 Figure \ref{fig:FasterRCNNTracking}  shows the result of the object detection and tracking, where each object
is confined in a bounding box with a detection probability. This is performed by the Faster R-CNN described in section II. Using our modified R-CNNs, the objects
are segmented in different colors and tracked in real-time, as shown in Figure~\ref{fig:MaskRCNNTracking}. The sequences of segmented images are represented in the Hololens. The raw depth maps and RGB feeds presented in Figure \ref{fig:rs_feed} are combined to construct the 3D map of the scene via Intel Realsense built in a 3D reconstruction algorithm. We are currently working to combine these feeds with the thermal feed to construct a better 3D map in low light conditions.  \par

The current system's latency is about 0.1 seconds, which includes the computational time plus communications time. The lower communication and communication time is due to the low resolution of the data and the fact that the number of floating-point operations in the test is relatively low to be executed with the used GPU in a short time. In any case, the latency is noticeable by the user, and the use of more powerful processors can only reduce it. Therefore, there is a technological limitation in this aspect that will only be overcome as computer technology advances. The connection between the Hololens and the processor is local. It is performed through a personal area network, which is then assumed to be on at all times since the signals do not need to reach more than one foot (the distance between antennas in the firefighter gear). In separate research, the team proposed prototypes for nodal communications based on nodes carried by each of the firefighters, whose redundant nature makes communications robust in inconsistent indoor communications channels \cite{hamke2019mesh}.  \par

The work presented here lays a foundation for developing a real-time situational map of the structural configuration of a building that is actively built and updated via the live thermal imagery being recorded by firefighters moving through the scene. An initial demonstration of indoor positioning and path planning is presented in \cite{vadlamani2020novel}, which is based on the estimation of camera movement through assessment of the relative orientation with SIFT and Optical flow.  This map, which is updated in real-time, could be used by firefighters to safely navigate the burning structure and improve the situational awareness necessary in decision making by tracking exits that may become blocked and finding alternatives. Utilizing the features detected via the approach presented in this work, a robust localization and tracking system to track objects of interest in sequences of frames can be built. The visual features from this framework have also been coupled with a Natural Language Processing (NLP) system for scene description and allow the framework to autonomously make human-understandable descriptions of the environment to help firefighters improve their understanding of the immediate surroundings and assist them when anxiety levels are heightened. Our future work seeks to join these two components with a deep reinforcement learning (Q-learning) algorithm that utilizes the continuously updated state map to assist in path planning. The navigation steps computed by this algorithm can be vocalized through the NLP system. The deep Q-learning based approach provides a navigation system that actively avoids hazardous paths. These three components, built on the backbone of the research presented here, can be fused to accomplish the ultimate goal of providing an artificially intelligent solution capable of guiding firefighters to safety in worst-case scenarios. Further, to compress the neural net parameters into an embedded platform for real-time performance, we would like to explore the tensor train networks\cite{novikov2015tensorizing} and distributed implementation \cite{bhattarai2020distributed} for large scale data.  In moments of panic, our system may help firefighters regain a sense of their surroundings and the best path they can take to exit the burning structure.


During the initial development of the embedded system, some challenges related to implementing the COTS hardware were encountered. This included the lack of existing Tegra kernel driver modules for the thermal and depth cameras and compatibility between the Intel RealSense D435i camera and the NVIDIA Jetson TX2 used for initial development. The TX2 did not provide the USB-C 3.0 connectivity required by the camera and could not capture the RealSense RGB feed. Therefore, a second embedded GPU platform was required for prototype testing, the NVIDIA Jetson AGX Xavier, while the FLIR One thermal camera development was conducted on the TX2.

\begin{figure}[!t]
\centering
\includegraphics[width=3.49in]{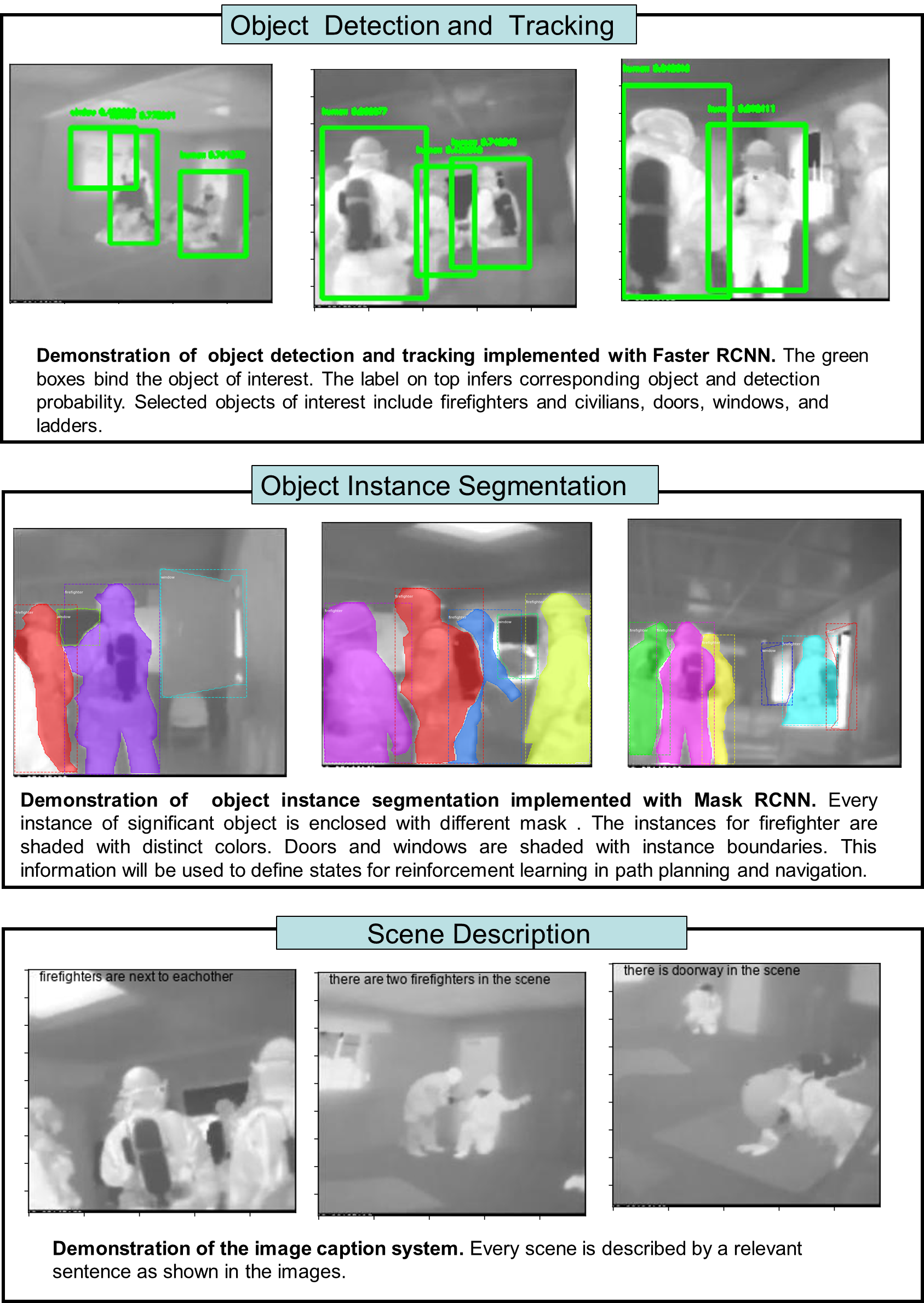}
\caption{Processed datasets}
\label{dataset_results}
\end{figure}


\subsection{Drawbacks}
\begin{figure}
\centering
\includegraphics[width=3.49in]{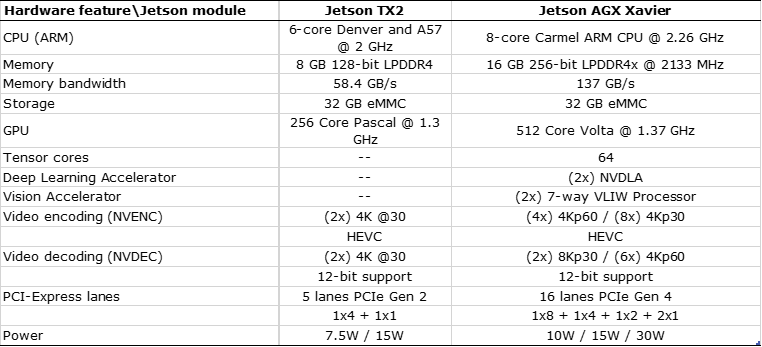}
\caption{Comparison between Jetson modules}
\label{fig:jetson}
\end{figure}
Although the separation of the camera development over two platforms provides the additional benefit of a dedicated GPU for image processing, this architecture is impractical for firefighters at the fireground. Additionally, while the TX2 is a more cost-effective embedded development platform, the AGX Xavier provides greater processing capabilities (see Figure~\ref{fig:jetson}). Furthermore, the Jetson AGX Xavier does not natively provide WiFi connectivity capabilities, so that additional hardware would be required for field implementation. Future work will need to address these hardware issues.
\chapter{Deep RL for Path Planning and Navigation}

\section{Motivation}


Near-zero visibility, unknown hallways, deadly heat and flame, and people in dire need. These are the challenges firefighters face with every structure fire they respond to. Firefighters endure extreme external conditions and the internal hazards of stress, panic, and disorientation as part of their daily job. Their primary weapon against internal and external dangers is their training on maintaining situational awareness or understanding the activities and circumstances occurring in their immediate vicinity. Maintaining situational awareness is key to a firefighter's quick and apt response to an ever-changing environment and is critical to accurate decision-making. Situational awareness can be heavily impacted by external hazards related to fire and the first responders' corresponding internal stresses. Loss of situational awareness is one of the leading causes of firefighters' loss of life on the scene. Firefighters must make prompt decisions in high-stress environments, continually assessing the situation, planning their next set of actions, and coordinating with other colleagues, often with an incomplete picture of the problem. Situational awareness is the foundation of further decisions on how to coordinate both rescue operations and fire suppression.  Firefighters on-scene pass their scene interpretations via portable radio devices to field commanders for further assistance in decision-making. The passing along of an inaccurate understanding of current conditions can prove disastrous. The limitation of this decision-making system is well reflected in the annual statistics by the US Fire Administration on the loss of human life\footnote{\href{https://www.usfa.fema.gov/downloads/pdf/publications/ff_fat17.pdf}{https://www.usfa.fema.gov/downloads/pdf/publications/ff\_fat17.pdf}}. Existing fire fighting protocols present an excellent use case for the institution of state-of-the-art communication and information technologies to improve search, rescue, and fire suppression activities through enhanced utilization of the data already being collected on-scene. 
\par
Firefighters often carry various sensors in their equipment, including a thermal camera, gas sensors, and a microphone to maintain their situational awareness. Still, this data is currently used only in real-time by the firefighter holding the instruments.  Such data has great potential for improving the fire teams' capability on the ground if the data produced by these devices could be processed with relevant information extracted and returned to all on-scene first responders quickly, efficiently, and in real-time the form of an augmented situational awareness.  
The loss of situational awareness is at the core of disorientation and poor decision making. Advancement in computing technologies, small, cheap, wearable sensors paired with wireless networks combined with advanced computing methodologies such as machine learning (ML) algorithms that can perform all data processing and predicting utilizing mobile computing devices makes it not only possible but quite feasible to create AI systems that can assist firefighters in understanding their surroundings to combat such disorientation and its consequences. This research presents a theoretical approach that can serve as the backbone upon which such an AI system can be built by demonstrating the power of deep Q-learning in creating a path planning and navigation assistant capable of tracking scene changes and offering firefighters alternative routes in dynamically changing fire environments.

\par
AI planning is a paradigm that specializes in design algorithms to solve planning problems. This is accomplished by finding a sequence of actions and addressing the needs and constraints to drive an agent from a specified initial state to a final state, satisfying several specified goals. We utilize these paradigms to build a framework that teaches the agent about fire avoidance and deploys a reactive decision process to successfully guide the agent through simulated spaces that are as dynamic as those encountered in live-fire events. Training in a simulated environment allows us to test many situations and train the agent for exposure to a vast number of scenarios that would otherwise be impossible in real life. As a result, we get a vastly experienced pilot capable of presenting quick recommendations to a wide variety of situations. The presentation of this technology is meant to serve as the basis for building a navigation assistant in future work. We propose using AI as the first step to providing services for response teams and survivors, focusing on supporting immediate search and rescue efforts, such as efficient and accurate navigation and localization in indoor and outdoor crisis zones through the production of evacuation routes and plans. 
\begin{figure}
    \centering
    \includegraphics[width=1\linewidth]{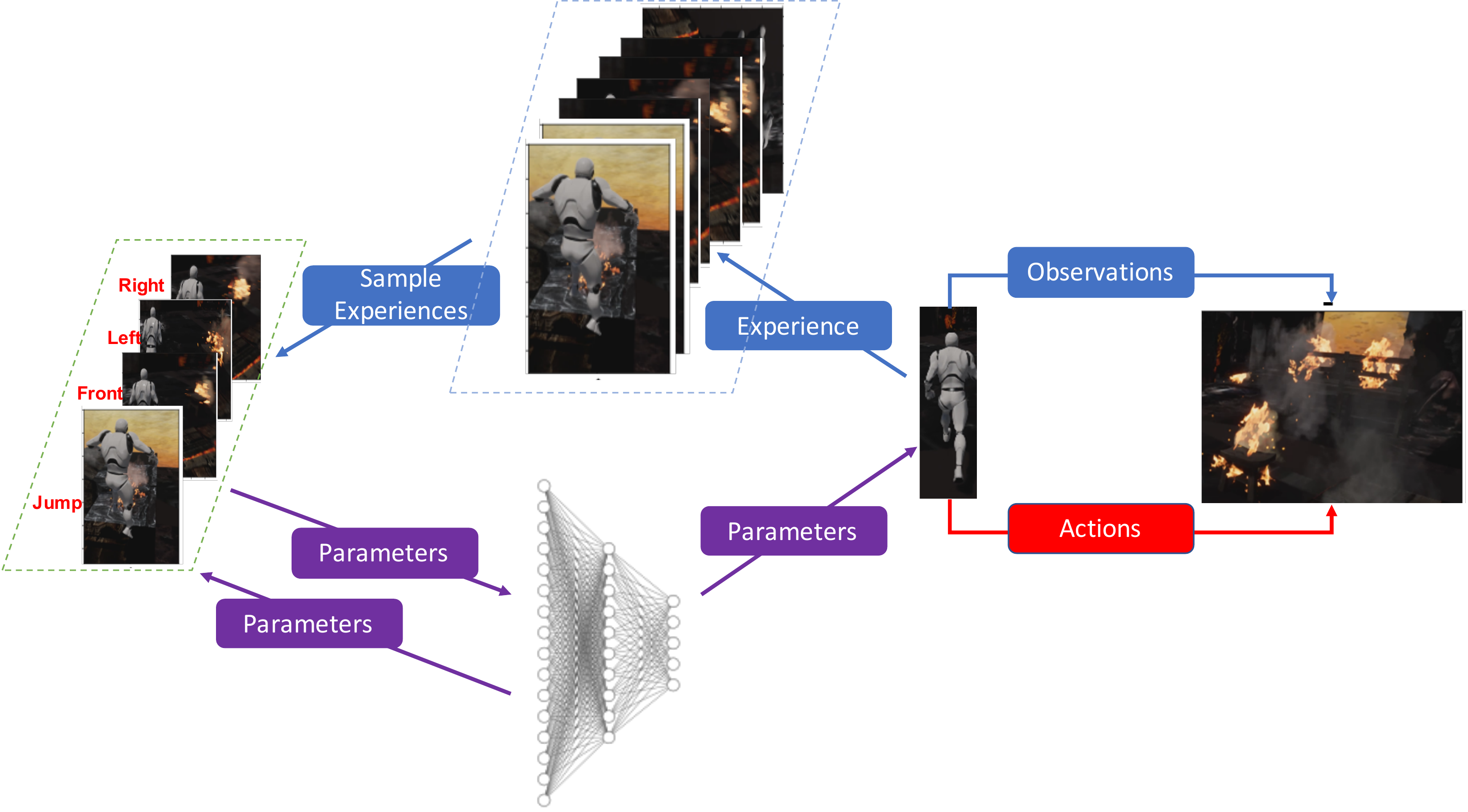}
    \caption{deep Q-Network implementation}
    \label{fig:dqn}
\end{figure}


\section{Preliminaries}

In this work, we introduce a DRL  approach to train an agent in a simulated fire environment. Taking advantage of simulation, we can expose our agent to a vast number of scenarios and dynamics that would be cost and safety prohibitive in real life. Still, the results of the training can be applied to actual live fire events. The resulting algorithm can be used in conjunction with other deep learning/machine learning approaches to produce a robust navigation assistant that can operate in real-time, effectively guiding firefighters through a fire scene and helping their decision-making by supplementing information gaps. Situational awareness lags through the correct interpretation of the scenes they have passed or are currently in.


\section{ Problem Description and RL Architecture}
The virtual environment is achieved in a gaming platform Unreal Engine\cite{qiu2016unrealcv} depicting a fire scenario of burning objects and smoke. The gaming engine uses computational fluid dynamics(CFD)\cite{anderson1995computational} based physics models to simulate a real-life dynamic situation where the parameters are a function of time. This gaming environment allows an external interaction where the agent can navigate the scene via external controls such as a keyboard or head movement in Virtual reality(VR) device. We take advantage of an interface software AirSim \cite{shah2018airsim} that allows communication to and from the gaming environment to a deep learning framework (Tensorflow \cite{abadi2016tensorflow}). AirSim can grab various gaming parameters such as RGB feed, Infrared feed, depth, semantic map information corresponding to the scene, and provide the feed to the python block. The python block then processes this information and dictates an agent's movement such as forward, backward, right, left, jump based on the deep reinforcement learning(RL) algorithm, and passes to the AirSim. AirSim further provides these control commands to the Unreal Engine environment, which emulates these motions. We deploy a deep Q-learning agent that is trained on a  policy-reward mechanism along with experience replay. For the experience replay and storing agent self-play, we also recorded the user interactions with the environment where they were asked to safely navigate the environment, avoiding the fire and reaching the target in the given scene. With each new start, the user is asked to take different routes with the virtual agent to reach the destination while avoiding fire, and the video frames and controls are recorded. During the RL agent training, the sequence of frames and controls from the experience replay memory is provided to accelerate the training process and make the non-differentiable optimization problem converge in a reasonable time with better accuracy.  The knowledge gained by the virtual agent on how to navigate the virtual scene successfully can then be transferred to a cyber-human system that can use this knowledge to interpret a real scene and provide step-by-step directions to firefighters to assist them in avoiding fire or other dangerous obstructions. 
The overview of the proposed DQN is shown in Figure \ref{fig:dqn}.
\par
Now, we define our objectives and various parameters associated with the proposed DRL framework. 
\subsection{Objective}
The proposed deep Q-learning-based agent's goal is to reach the destination while safely navigating the fire in a dynamic environment. Safe navigation is defined as avoiding any contact with simulated fire. During the test, the agent needs to make instantaneous decisions when fire appears unexpectedly in the chosen navigation path. To achieve the best decisions under such situations, the agents can be subjected to many worst-case situations during the training phase. The rewards need to be defined precisely to handle such task-driven learning. 
\subsection{Observations}
The Q-learning framework observations are collected through the agent's field of view(FOV) from the virtual simulation environment (i.e., Unreal Engine) using the AirSim app. The Python deep learning environment receives the observations in various feeds, including RGB, infrared, depth, and semantic map frames. We are particularly interested in the infrared frames as the CNN framework, 
is developed to perform recognition on thermal imagery. Infrared cameras are the only feed type which can withstand extreme fire and smoke situations and enable a see through smoke. The virtual environment can also provide information about the camera position and the agent position, which will help locate the agent in the given 3D environment. 
\subsection{Actions}
We have transformed the action space from continuous to discrete space for ease of implementation and proof of concept, which comprises five primary agent motions. This discretization of the agent space also helps to reduce the model complexity. The five actions are front, back, left, right, and jump. With these motions, the agent can navigate in a structural building containing obstacles like ladders and furniture. The same set of motions also enables the agent to navigate in wildfire scenarios. The agent may take one or a combination of these actions to navigate the fire scene to reach the destination. 
\subsection{Rewards}
It is imperative to define the direction of the goal while training an RL agent. To achieve a task-driven learning objective, it is vital to define the rewards to the navigating agent. The ultimate goal is to find a safe and minimal trajectory length to the navigation target. Unlike the trivial objective of finding the minimal trajectory length, the additional constraint of finding the safe path makes the optimization algorithm more complex. This results in a time-varying decision system whose instantaneous decisions are based on current information on the environment. We introduce a reward and a penalty for familiarizing the agent with the fire environment. The fire has a penalty of -10, and a reward for reaching the goal is 10. 

\subsection{Problem statement}
The RL agent tends to pivot the actions in the direction of maximizing the rewards. The DRL system optimizes the hyperparameters of the Q-neural network to encode the agent's experience for navigation. The backbone deep network aids the navigation by detecting the objects of interest for navigation. This information is then fed to the Q-network, choosing the optimal actions to guide the agent. The idea of a DRL system is to provide an end to end learning framework for transforming the pixel information into actions \cite{mnih2015human}. Most of the DRL systems aim to learn the neural network parameters to find a transformation from state representations $s$ to policy $\pi(s)$. It is also desirable to have an agent that can learn the navigation from a single environment and generalize the experience to various environments. To achieve that, the aim is to learn a stochastic policy function $\pi$, which can process a representation of current state $s_i$ and target state $s_t$ to produce a probability distribution over action space $\pi(s_i,s_t)$. During the test, the agent samples an action from this distribution until it can achieve the destination. To summarize, the objective function that is used to assess the model performance is given in the form
\begin{equation}
    z = g(x;\theta) = g(x;\beta(\theta);\theta) 
\end{equation}
where $g$ is the navigation problem i.e., finding the optimal actions with a DQN of parameters $\theta$ and $\beta$ are the navigation agent parameters. $z$ describes the navigation, and $J(z)$ measures the optimality of the estimated navigation decisions. For a given fixed set of neural net parameters $\beta$, the optimizer tends to seek optimal $\theta$ that determines the agent's actions. To measure the proposed actions' effectiveness, we use a scoring function $J$ that operates on $\theta$ parameters. 
\begin{figure}[t!] 
\centering
    \includegraphics[width=\linewidth]{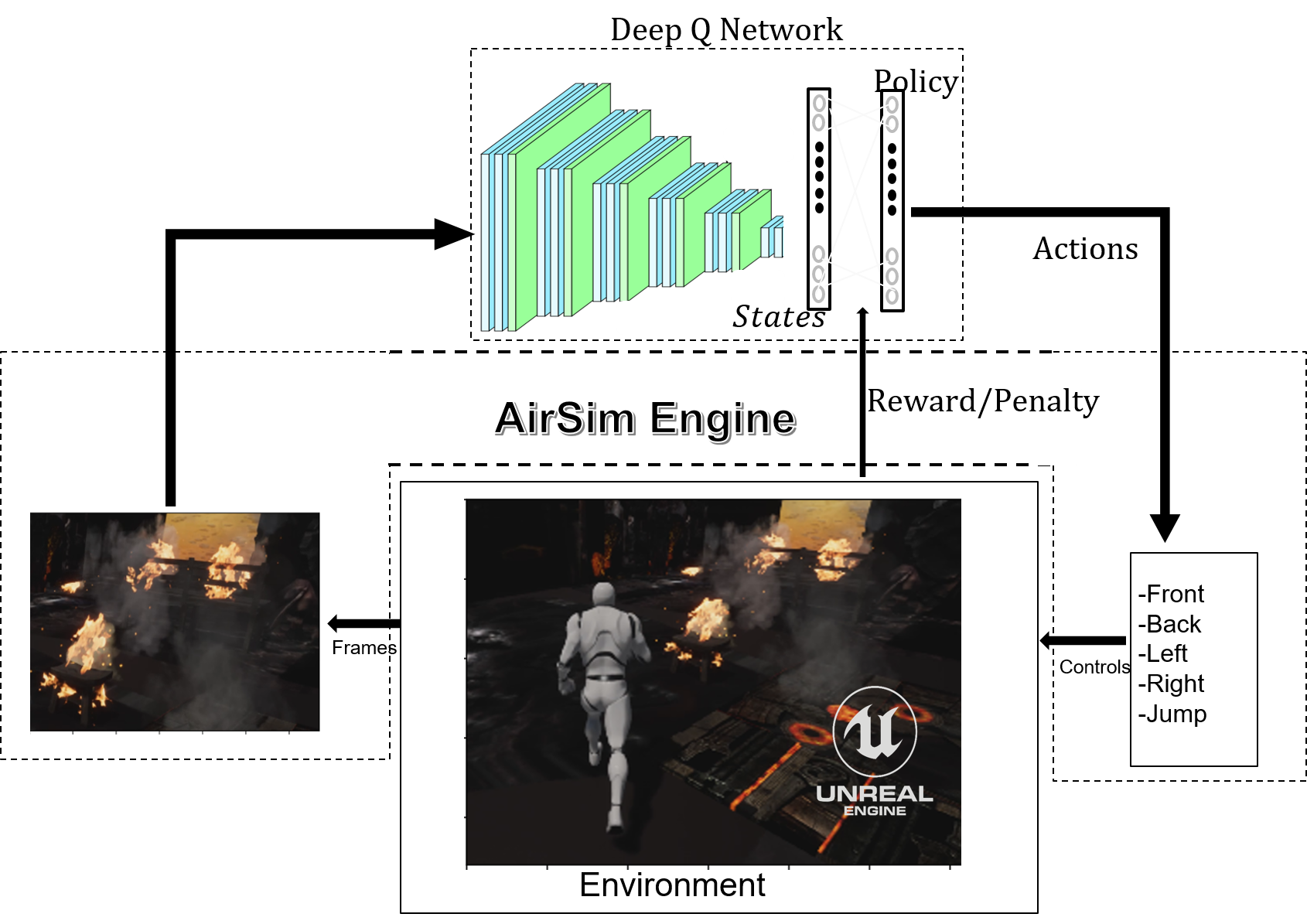}
    \caption{Architecture of Path Planning and Navigation system.  
 }
    \label{fig:sceneunderstanding1}
\end{figure}

\subsection{Model}
This work aims to find an optimal policy that can aid a firefighter to navigate in a fire setting via deep reinforcement learning. A deep neural network is trained for a non-linear approximation of the policy function $\pi$, where action $a$ at time $t$ is sampled as :
\begin{equation}
    a \sim \pi(s_i,s_t| \beta) 
\end{equation}
where $\beta$ corresponds to NN model parameters, $s_i$ is the current observation frame, $s_t$ is the target observation to which navigation is to be performed with action sequence $a$. Here, $s_t$ belongs to a discrete set, $\pi$ is a distribution function. The target scene can comprise fire victims who need help for rescue. So, once the DL model estimates the target to be rescued, the RL agent proposes navigation paths that successfully rescue the victim. 

\subsection{Q-learning and Deep Q-learning}
We employ a variant of Q-learning called Deep Q-learning (DQL)\cite{mnih2013playing} to train an agent for navigating the fire to reach the destination safely~\cite{icaart21}. In this section, we briefly give an overview of the Q-learning and Deep Q-learning algorithms. 
\par
Q-learning learns the action-value function $Q(s, a)$ to quantify the effectiveness of taking action at a particular state. Q is called the action-value function (or Q-value function ).  In Q-learning, a  lookup table/memory table $Q[s, a]$ is constructed during training to store Q-values for all possible combinations of states $s$ and actions $a$. An action is sampled from the current state, followed by computation of reward R (if any) and then the new state $s' $. The next action $a' $ is determined based on the maximum of $Q(s', a')$ from the memory table. 
After this, an action $a$ is performed to seek a reward of $R$. Based on this one-step look ahead, the target $Q(s,a) $ is set to 
\begin{equation}
    target = R(s,a,s') + \gamma max_{a'} Q_k (s',a')
\end{equation}
The update equations are called Bellman equations \cite{bellman1966dynamic} and are performed iteratively with dynamic programming. 
As this update is performed iteratively until convergence, a running average for Q is maintained.


\begin{algorithm}[!ht]
    \caption{Deep Q-learning algorithm for path planning agent} \label{alg:q-learning}
    \begin{algorithmic}[1] 
    \State  Initialize replay memory $\mathcal{R}$ to capacity N. 
    \State Initialize the Q-function $Q(s,a)$ for all $s$,$a$ with random weights.
    \For{episode in 1,2,.. M} \Comment{each execution sequence}
    \State Initialize sequence $s_1 = {x_1}$
    \For{t in 1,2..T} \Comment{decision epoch}
    \State With probability $\epsilon$, select a random action, otherwise select 
    $a_t=max_a{Q^*(s,a;\beta)}$ \Comment{\textbf{Exploration vs Exploitation Step}}
    \State Action $a_t$ is performed by agent in the environment and corresponding rewards $r_t$ and scene $x_{t+1}$ is observed. 
    \State Set $s_{t+1} = s_t,a_t,x_{t+1}$ 
    \State Store $s_{t+1}$ in $\mathcal{R}$ . 
    \State Sample a batch of transitions  $e_k = (x_k,a_k,r_k,x_{k+1})$ from $\mathcal{R}$. 
    \If {$x_{t+1}$ is terminal}
    \State $y_k$= $r_k$
    \Else{}
    \State $y_k = r_k + \gamma max_{a'} Q(s_{k+1},a';\beta)$
    \EndIf
    \State Compute loss $(y_k-Q(s_k,a_k;\beta_k))^2$ and then update neural net parameters $\beta$ with gradient descent and back-propagation as per equations 4,5 and 6 .
\EndFor
\State  
\EndFor
     \end{algorithmic}   
\end{algorithm}

However, for solving a real-world problem such as path planning and navigation, where the combinations of states and actions are too large, the memory and the computational requirements for Q is costly and intractable in some cases. To address that issue, a deep Q-network (DQN) framework was introduced to approximate Q(s, a) with the aid of neural network parameters. The associated learning algorithm is called Deep Q-learning. Based on this approach. we can approximate the Q-value function with the neural network rather than constructing a memory table for Q-function for state and actions. \par

An RL system needs to know the current state and actions to compute the Q-function. However, for our proposed simulation environment, the internal state information is not available. In one way, the state information can be constructed based on a recognition system that can identify the object of interest in the scene resulting in a discretization of the observation space by assigning pixels discrete values based on their identity. This objective is out of the scope of this work and will be pursued in the future. We only focus on observing a frame $x_t$ from the emulator for this implementation, which is a grayscale infrared image.   Based on the action performed in the environment, the agent receives a reward $r_t$, along with a change in the internal state of the environment. Since we have defined a finite reward/penalty corresponding to specific states, the agent might need to go through a series of actions before observing any reward/penalty.

To estimate the Q-function, we consider the sequence of actions and observations for a game play episode. It is given as $s_t=x_1,a_1,x_2,a_2,…..a_{t-1},x_t$. Considering $t$ is a finite time where the game terminates either by reaching the target or getting burnt in a fire, this sequence can be formulated as a markov decision process (MDP).  The goal of the agent is to choose the action that maximizes the sum of future rewards where the reward at time $t$ is given as $R_t=
\sum_{t_1=t}^T \gamma^{t_1-t} r_{t_1}$ for T being episode time. We then use a $Q^*(s,a)$ as optimal action-value function for a given sequence $s$ and action $a$ where
$Q^*(s,a) = max_{\pi} \mathbb{E}[R_t|s_t=s,a_t=a,\pi]$, $\pi$ being the distribution over actions . To estimate this $Q^*(s,a)$, we use the deep neural network (Q-network) of parameters $\beta$ as a non-linear function approximator  in the form $Q(s,a;\beta)$ where it is expected that $Q(s,a;\beta) \approx Q^*(s,a)$.  This network is trained with an objective of minimizing a sequence of loss functions $L_k(\beta_k)$ where,

\begin{equation}
L_k(\beta_k) =\mathbb{E}_{s,a\sim\psi(.)}[(y_k-Q(s,a;\beta_k))^2],
\end{equation}

Where $y_k = E_s'[r+\gamma  max_a' Q(s',a';\beta_{k-1})|s,a]$ is the target for iteration k and $\psi(.)$ is the probability distribution of sequences s and actions a.

The neural net then back-propagates the gradient given as

\begin{equation}
\begin{aligned}
\nabla_{\beta_k} L_i(\beta_k) = \mathbb{E}_{s,a\sim\psi(.)} \Big[\Big(r+\gamma max_{a'} Q(s',a';\beta_{k-1}) \\
-Q(s,a;\beta_{k})\Big) \nabla_{\beta_k} Q(s,a;\beta_k)\Big]
\end{aligned}
\end{equation}

The parameters of the neural network are updated as

\begin{equation}
\beta_{k+1} = \beta_{k} - \alpha \nabla_{\beta_k} L_i(\beta_k)
\end{equation}
where $\alpha$ is the learning rate of the neural network.


Furthermore, a technique called experience replay \cite{ mnih2013playing} is used to improve convergence. This occurs by exposing the model to human-controlled navigation and decision making. To implement experience replay, the agent’s experience $e_t$ = ($s_t,a_t,r_t,s_{t+1}) $ at each time step $t$  is stored, where $s_t$ is the current state, $a_t$ is the action, $r_t$ is the reward, $s_{t+1}$ is the next state on taking action $a_t$. The experience calculations presented result from the human-controlled navigation training and interaction with the environment. Based on the human interaction, various episodes $\{e_i\}_{i=1}^N$ are stored in a memory buffer $M$. During the inner loop Q-learning updates, a sample of experiences is drawn randomly from the memory buffer $M$. The agent then selects and executes an action based on an $\epsilon-$greedy policy. The sampling approach randomly from experience replay enables the agent to learn more rapidly via improved exposure to reactions to different environmental conditions during an episode of training and allows the model parameters to be updated based on diverse and less correlated state-action data. The algorithm corresponding to the deep Q-learning is presented in Algorithm \ref{alg:q-learning}, and implementation methodology is presented in Figure \ref{fig:dqn}.  

\subsection{Network Architecture}
The DQN framework is built on top of a VGG-net-like framework \cite{bhattarai2020deep} as a backbone and is shown in Figure  \ref{fig:sceneunderstanding1}. The backbone framework is a feature extractor that produces 4096-d features on a 224x224 infrared/thermal image. The VGG framework is frozen during the training. A stack of 4 history frames is used as state inputs to account for the agent's past sequence of actions. Then the concatenated feature set comprising $4 \times 4096$ is projected into a 512-d embedding space. This vector is then passed through a fully-connected layer producing 5 policy outputs, which give the probability over actions and value output.

\begin{figure*} 
\centering
    \includegraphics[scale=0.6]{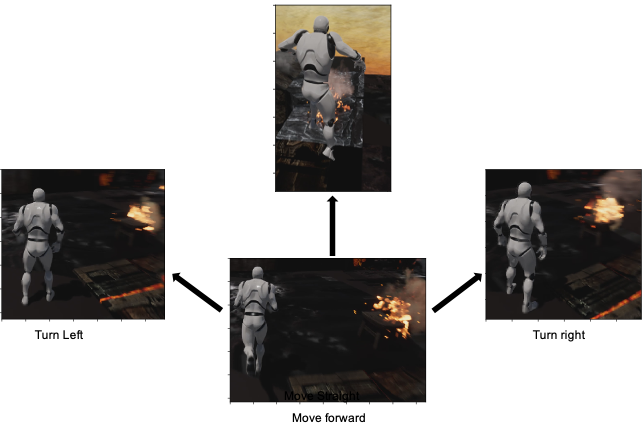}
    \caption{\textbf{ Demonstration Of  Agent Actions In Fire Environment As Dictated By Reinforcement Algorithm.}. Four primary actions demonstrated respectively are turning left, right, jump, and moving forward. This enables the agent to avoid fire and obstacles and safely reach a given destination. 
 }
    \label{fig:Pathplanning}
\end{figure*}
\section{Experimental Results}
The DQN model's implementation was done in Tensorflow \cite{abadi2016tensorflow} on a dual NVIDIA GeForce 1080Ti GPU. The DQN framework is trained with an RMSProp optimizer \cite{tieleman2012lecture} with a learning rate of $10^{-4}$ and a batch size of 32. The training was performed with $\epsilon-$ greedy with $\epsilon$ started at one and decayed to 0.1 over 5,000 frames. The training afterward was continued with $\epsilon$ of 0.1. The whole training was performed with as many as 100000 frames with a replay memory of 20,000 frames.  
While training, the gradients are back-propagated from the Q-layer outputs back to lower-level layers while the backbone model is frozen.  The navigation performance was measured by the agent's ability to reach 100 different targets set in a given fire environment. The target was placed at various locations in the virtual building on different floors, where the agent needed to navigate using a combination of all actions. The actions of the agent in successful navigation are shown in Figure \ref{fig:sceneunderstanding}. It is complicated to report the average trajectory length due to the constant changes occurring along the virtual path. In this simulation, the agent needs to avoid fire, and the trade-off for that is time. Imposing a time constraint and weighing the reinforcement model for rewarding or penalizing according to both strictures is a goal for future work.  To prove the efficacy of learning deployed by the proposed method, we intensify the situation by increasing fire prevalence. The fire volume per scene was increased from 10\% to 80\% ground coverage of the scene, with fires occurring at random across each game. The agent trained using the shortest path planning strategy (A*) failed to reach the destination when fire coverage reached 30\% while the DQL trained agent was able to consistently navigate across a scene to the destination with a fire coverage up to 76\%. For extreme fire conditions, we carefully increased the rewards and defined additional penalties (distance to fire) to better the agent's learning condition.

\par 
The proposed algorithm's main goal is to find the least number of combinations of actions that help the agent navigate from the current position to the destination while avoiding fire. Due to the dynamic nature of the environment, when we attempted to solve this problem with other path planning techniques, including shortest path technique, breadth-first search(BFS) \cite{beamer2012direction}, depth-first search(DFS) \cite{tarjan1972depth}, A*\cite{lavalle2006planning} and random walk\cite{spitzer2013principles}, the probability of the agent reaching the destination was very low (less than 5\%) under the simulation environment.  Since these methods use a single shot decision map to navigate the agent to the destination, the agent could not quickly adapt to the continuously varying surroundings. When the agent encountered fire that was not present before the decision, the agent failed to reach the target most of the cases. In contrast, the agent governed by our proposed method was able to reach the destination with a probability greater than 80\%.  
\par
During the navigation, when the DQL trained agent cannot find the path to proceed, we design penalties, so the agent is constrained to either staying in the same position until the course is cleared or retracing its steps back to the previous possible path. The authors note this part of the algorithm requires further attention, as remaining in one place in a real fire scenario is not realistic. 

\section{Movement Planning Through Deep Q Learning for Firefighting Application}
We have demonstrated the potential deep Q-learning based algorithms hold as a base framework off of which a successful navigation assistant can be built. This methodology can provide an efficient decision-making system for aiding firefighters whose decision-making abilities may be impaired due to disorientation, anxiety, and heightened stress levels. This work presents a novel approach to eliminating faulty decisions made under duress by applying AI planning paradigms.  The paths followed by the firefighters are useful to determine their positions, which is particularly important in search and rescue. 
\par
Existing path planning algorithms can process all paths followed by the firefighters but fail under the fire ground's continuously changing nature, making a previously defined rescue plan unavailable. Also, smoke and other visual impairments could make difficult the rapid identification of these incidents by a firefighter.  Incidents in the fire ground are hardly predictable by a machine learning system. However, machine learning does perform well in rapid assessment and production of a decision given the current set of circumstances. In other research outside of this work's scope, \cite{bhattarai2020deep,bhattarai2020embedded} have developed a machine learning-based methodology that detects and tracks objects of interest such as doors, ladders, people, and fire in the thermal imagery generated by firefighter's thermal cameras. Such information may be valuable to improve further the reinforcement learning algorithm's ability to understand aspects of the environment that may be used in navigation or escape. Future work looks to incorporate a similar object detection work with the path planning work described here to make a robust navigation assistant capable of understanding the surrounding environment outside of fire presence and then recommend the best paths to firefighters. To deploy the agent in a real fire situation, we also aim to construct the 3D map based on multimodal data(RGB, infrared, and depth map) collected from various sensors attached to the firefighter's body sensors. Such a map can be imported to the emulator to train the agent in a more natural look-a-like environment.   
\par
Assuming that the system has previously determined a path using the information coming from the rescuer's camera, the rescuer has access to an initial rescue path. The system tells the rescuer to take a direction, which is the present action. The states will be represented by the objects present in the scene. The objects of interest can be expressed in a matrix that contains the extracted feature's image. Each detected feature has a different reward. Fire and obstacles have associated penalties, while a clear path has a positive reward. If an obstacle is detected, then the firefighter is told to take a different direction. The new state will be computed for the action taken, and an original path will be traced. The recursion repeats until the rescuer has reached the desired position. The plan for this part of the research will include an initial model constructed by simulation. This will help determine the right design for the neural network in terms of stability and convergence speed in different simulated situations. When incorporating information from other paths followed by other firefighters in a real scene, it is not evident that their past experiences can determine all obstacles due to the dynamic nature of a real fire scene.
Nevertheless, the parallax estimation obtained from sequences of cameras in motion can be helpful. Parallax data can determine the depth of a given path because it gives the distances between the camera and the critical points detected by the SIFT algorithm. This information only needs to be stored and compared with future sequences of the same path. We can consider that an obstacle has been found in a previously clear way if the estimated depth has dramatically changed in this case. The direction pointed by the camera will be given a low reward instead of a high one.

 Movement planning is the most crucial outcome of this research. Fires create actively changing environmental conditions. Within such an environment, the system needs to inform firefighters of the best possible movement options available to them while accounting for these real-time changes. 
 The proposed reinforcement-based learning approach demonstrates the feasibility of such methods deployed in the field to address these constraints and provide optimal path planning. We apply a Q-learning algorithm to the acquired environmental/changing state information and use it to generate the best path alternatives. This approach provides an efficient decision-making system that aids firefighters whose decision-making abilities may be impaired due to disorientation, anxiety, and heightened stress levels and presents a novel approach to eliminating faulty decisions made under duress.
\par
 The paths followed by the firefighters are useful to determine their positions, which is particularly important in search and rescue. Existing path planning algorithms can process all paths followed by the firefighters but fail under the fire ground's continually changing nature, making a previously defined rescue plan unavailable. Also, smoke and other visual impairments could make difficult the rapid identification of these incidents by a firefighter.  Incidents in the fire ground are hardly predictable by a machine learning system. However, machine learning does perform well in rapid assessment and production of a decision given the current set of circumstances. In other research outside of this work's scope, we have developed a machine learning-based methodology that detects objects of interest in the thermal imagery generated by firefighter cameras such as doors, ladders, people, and fire. Other items of interest can be readily incorporated into the detectors upon the deep learning structures' training. Future work looks to include the object detection work with the path planning work described here to make a robust navigation assistant capable of understanding the surrounding environment outside of fire presence and then recommend the best paths to firefighters.
\par
With the successful creation of a deep Q-learning algorithm capable of split-second decision making, our next step is to make use of the detection and classification of objects present in the scene to improve further the reinforcement learning algorithm's ability to understand aspects of the environment that may be used in navigation or escape. Assuming that the system has previously determined a path using the information coming from the rescuer's camera, the rescuer has access to an initial rescue path. The system tells the rescuer to take a direction, which is the present action. The states will be represented by the objects present in the scene.

The objects of interest can be represented in a matrix that contains the extracted features image. Each detected feature has a different reward. Fire and obstacles have zero rewards, while a clear path has a positive reward. If an obstacle is detected, then the firefighter is told to take a different direction. The new state will be computed for the action taken, and a new path will be traced. The recursion repeats until the rescuer has reached the desired position. The plan for this part of the research will include an initial model constructed by simulation. This will help determine the right design for the neural network in terms of stability and convergence speed in different simulated situations. The radio connectivity can help in the path redesign. This information will be further incorporated into the simulation. Unless fire, other obstacles that were not previously on the scene are not easy to detect. When integrating information from different paths followed by fighters may be useful, it is not evident that all obstacles can be determined.
Nevertheless, the parallax estimation obtained from sequences of cameras in motion can be helpful. Parallax data can determine the depth of a given path because it gives the distances between the camera and the key points detected by the SIFT algorithm. This information only needs to be stored and compared with future sequences of the same path. We can consider that an obstacle has been found in a previously clear path if the estimated depth has dramatically changed in this case. The direction pointed by the camera will be given a low reward instead of a high one. 
\par
To incorporate this information and other refinements, real environment simulations need to be tested with an infrared camera. Thus, the next step of the design will be to program a real-time acquisition system tested with deep Q learning. The tests can be performed in a typical environment with a predetermined path, where obstacles will be intentionally placed to test the obstacle detection capabilities. This will need the recording of sequences in clear directions and test with the real-time acquisition system. We will simulate the presence of fire by placing cardboard panels with a given sign. The machine can be easily trained to detect these signs instead. Final evaluations will be tested in a fire ground training environment.
\par

To deploy the agent in a real fire situation, we aim to construct the 3D map based on multimodal data collected from various sensors attached to firefighter's body sensors. We propose to fuse RGB, Infrared, and depth information collected using different imaging cameras live from fire scenes. We will combine these different data types into a blended 3d scene, which can be viewed on augmented reality (AR) / virtual reality device. This map can also be parallelly be accessed by the fire commander to access the risk of danger of fire, situational awareness of the building in 360 video/ 3D that can be used for decision making and setting effective control plans for the firefighters. The multispectral imaging helps to access the temperature conditions at fire regions and structural damages of those zones, and the level of firefighting required to address the needs. Such an environment can be imported into a gaming environment such as unreal engine/ Unity to train the AI agent using deep reinforcement learning (RL). Given the feed of a real environment with burning objects and simulated fire, the AI agent can be trained to avoid fire and navigate safely through the fire with the aid of a model that works based on rewards and penalties. The model can be developed with the deep RL so that the agent learns the best to achieve the objectives by maximizing the sum of rewards on the game, safely navigating along fire to reach the destination. Once the agent learns the scene, it can be deployed in a real-world situation to assist the firefighter in path planning and navigation by using the data collected from the same source used to construct the gaming environment. 

Such a simulated AR environment can also provide an immersive experience of a real fire to firefighters in a different scenario and make them more prepared to be effective under different firefighting environments. 
\chapter{Tensor Decomposition for Firefighting}

\section{Motivation}

The current sensor technology needs human-based decision making to analyze the sensor gathered data. As the size of data produced by these sensors grows along with the number of sensors, the massive data impose a challenging problem for a human to make an efficient and reliable decision. There is a need for a mechanism to allow some level of automated decision-making where data size is not a bottleneck to address this challenge. AI-powered sensors are smart technology that enables the data from sensors to be analyzed by AI to make human-level decisions on a piece of hardware with machine learning algorithms. So, we propose one such AI-based mobile application platform to capture the data acquired by the sensors and make inferences on them in-situ with machine learning techniques to inform about any imminent incidences on mission-critical such as firefighting applications. 
\par
In modern technology, the sensors have been deployed to collect data from human bodies via electrodes to distant galaxies via telescope arrays. Sensors are regarded as the sixth sense for the current technological world. However, these wide fidelity sensors deployed across various domains have produced a massive collection of data that creates a bottleneck for most scientific processes.  Also, this huge data seeks for the need for massive storage and parallel cluster even for simpler computations. So there is a need for a compression/dimensionality reduction technique at the sensor's acquisition phase so that the post-processing bottleneck can be avoided, enabling efficient data storage, transfer, and analysis. Scientific data is naturally multidimensional, where each dimension may reflect the spatial/ temporal/intensity components, e.g., CT, MRI, and Videos. Most of these data can be decomposed to a lower-dimensional multi-linear representation so that the decomposition only encodes the most significant latent representation. It is well-known that the high-dimensional tensor data can be decomposed to a lower-rank representation via various tensor factorization methods, almost without a loss of information [3].
\par
We have developed efficient unsupervised dimensionality reduction techniques for finding the most efficient and optimal lower-dimensional representation for any dimension tensors. The limitations of the available in-situ computational resources mandate new algorithms for machine learning (ML), feature extraction, recognition, and decision making, which we propose to develop here.

\par
Deep learning frameworks such as convolutional neural networks(CNNs), Generative Adversarial Neural networks(GANs)\cite{goodfellow2014generative}, Dynamic graph neural networks (DGNN)\cite{Ma2018DynamicGN}, graph convolutional policy networks(GCPNs)\cite{you2018graph} and attention networks\cite{jaderberg2015spatial} have demonstrated their state of art performances in supervised machine learning applications such as classification, augmentation, segmentation, clustering, path planning, and predictions. Despite their proven accuracy, most of the above frameworks suffer from performance bottleneck regarding the size and volume of data for training and convergence time. The models can only take data of specific size and distribution as input and be well labeled to ensure superior performance.  As it would be challenging to feed the raw data from the sensor to these frameworks, instead we can provide the data processed by the unsupervised dimensionality reduction techniques such as tucker decomposition\cite{kim2007nonnegative}, PARAFAC\cite{bro1997parafac}, and other tensor decomposition techniques and still ensure we can achieve superior performances. The other bottleneck on the implementation of neural nets is imposed by the fully connected layers due to their colossal memory requirement to store the connection weights and the need for computational resources of large matrix multiplications within these layers. This would lead to a significant issue in low-end embedded devices. \cite{novikov2015tensorizing} proposes a tensor decomposition based technique for compressing the parameters of fully connected layers of a deep neural network. 
\par
Tensor factorization (TF) based methods won the Netflix Prize for their predictive power, and their easy interpretable results \cite{koren2009matrix}. TF has since been used in multiple industrial and real-world domains with increasing success. In recent years, various tensor factorization approaches have been applied alongside neural networks, including deep learning-based ones. While the tensor factorization methods allow for easily interpretable factors, they tend to be multi-linear. Hence, they have a relatively weak predictive power compared to the non-linear neural network-based deep learning's complex predictive power. However, the latter suffers from being mostly highly black boxes, often uninterpretable, and, being strongly non-linear, they are vulnerable under adversarial noise. Starting most notably with Recurrent Recommender Networks (RRNs), deep learning and tensor factorization methods have been recently combined, thereby partly modeling the temporal evolution of user and item preferences. Recent research of Microsoft has highlighted a possibility for merging these two techniques to integrate the nonlinearities inherent in the neural networks while preserving the human-understandable results as in tensors \cite{yu2012deep,yang2019wide,hutchinson2012tensor}. 
\par

For an unsupervised task, the major challenge in any model analysis is the determination of model hyperparameters.  Considering the tensor decomposition, the model hyperparameters estimate the correct latent dimension in the data \cite{anandkumar2014tensor}, which is an NP-hard problem.
Some of the earlier work for model selection is the  Automatic Relevance Determination(ARD)\cite{morup2009automatic} or generalized class of information criteria for tensors with specific properties \cite{chen2008extended}.  Particularly for blind source separation and anomaly identification, it is imperative to compute the exact number of latent features to identify the hidden mechanisms,  causes,  propagating channels, and group of events. 
To determine the number of significant features/latent dimensionality in the NMF,  we utilize a  model determination technique,  called  NMFk,\cite{alexandrov2014blind,alexandrov2013deciphering}. This technique was initially proposed to decompose human cancer  genomes\cite{alexandrov2013signatures}.  NMFk is based on a pipeline where the decomposed/extracted factors by NMF undergo custom clustering and Silhouette statistics\cite{rousseeuw1987silhouettes} to measure the stability of the latent features. In this work, we extend the use of NMFk for determining the latent dimensions for non-negative CPD and non-negative Tucker. 
\par
NMF is an unsupervised, parts-based learning methodology \cite{lee2001algorithms} that has been useful in pattern recognition, text mining, sparse coding, multimedia data analysis, speech recognition, information retrieval, social network analysis, and more, but remains under-explored in computer vision and pattern recognition. Due to the non-negativity constraint in NMF, the extracted latent features are physically interpretable as they are parts of the data. Also, many variables, e.g., pixels, density, counts, etc., are naturally non-negative, and the extracted features will not have physical meaning if the non-negativity is not in place. 

In this work, NMF allows addressing concerns regarding possible feature interactions and obtain a latent feature space that reveals organizations of features into clusters. Specifically, we propose two NMF-based methods, one to obtain a best cluster and another to obtain the best latent representation for the purpose of feature selection. 
\section{Methods}
A vector is a one dimensional(1-D) array, matrix is a 2-D array and tensor is a higher order multidimensional(N-D) array. In this work, we use lowercase bold letters $\vect{x}$ to denote vector,  uppercase letters $\mat{X}$ to denote matrices and uppercase script letters, $\ten{X}$ to denote tensors. The $L_2$ norm which is also called the Frobenius norm, of a vector, matrix and Tensor is respectively given as 
$||\vect{x}||_F = \sqrt{\sum_{i} \vect{X}_{i}}$, 
$||\mat{X}||_F = \sqrt{\sum_{i,j} \mat{X}_{i,j}}$ and $||\ten{X}||_F= \sqrt{\sum_{i,j,k} \ten{X}_{i,j,k}}$. Here 
$\vect{x}_i$ correspond to a scalar, $\mat{X}_{i}$ correspond to a vector and $\ten{X}_i$ correspond to a matrix.

\begin{figure}[!ht]
    \centering
    \includegraphics[width=.8\textwidth]{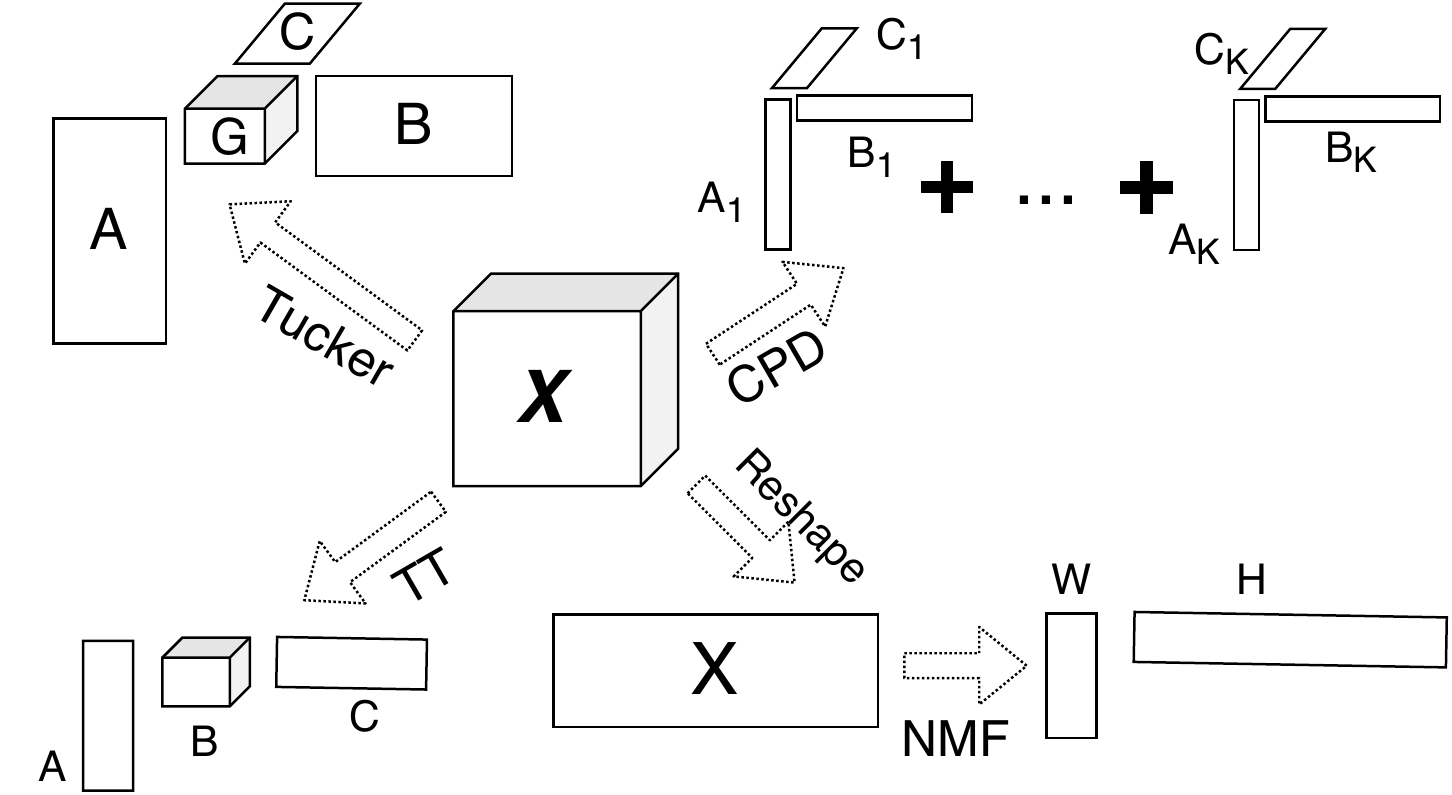}
    \caption{Tensor decomposition illustration}
    \label{fig:tensor_Decomp}
\end{figure}

\subsection{Tensor Factorization}

A big-data analysis is often difficult to directly link the data to the generating processes' parameters since the datasets are formed exclusively by directly observable quantities. In contrast, the underlying processes/features remain unobserved, hidden, or latent \cite{everett2013introduction}. Extracting these latent features reveals valuable information about hidden causality and mechanisms, but it also reduces the dimensionality by revealing the low-dimensional latent structure representing the whole dataset. \par
Most datasets are high-dimensional and are represented by \textit{tensors}, or multidimensional arrays. Such tensors typically describe multiple concurrent latent processes imprinting their signatures in the observable state variables in different dimensions. \textit{Tensor factorization}, which is the higher-dimensional analog of matrix factorization, is an unsupervised learning method that represents a cutting-edge approach for factor analysis. Its main objective is to decompose a high-dimensional tensor into factor matrices 
where the factor matrices carry the latent features in each tensor dimension \cite{kolda2009tensor}. There are various tensor decomposition techniques reported in the literature. Out of them, the prominent techniques are Canonical Polyadic Decomposition (CPD)\cite{bro1997parafac}, Tucker decomposition \cite{tucker1966some} and Tensor-train decomposition\cite{oseledets2011tensor}. CPD and tucker are primarily used in the latent feature extraction, whereas the tensor train decomposition is used for compression of big data and complex structure of Neural nets. It is a complex problem to identify the correct number of features for CPD as it's unique. However, tucker decomposition is not unique, but by Tucker decomposition, one can extract the minimal subspaces \cite{falco2012minimal}, representing the latent features in the corresponding dimensions \cite{alexandrov2019nonnegativeA}. A brief overview of these decomposition techniques is presented in the following sections. 

\subsubsection{Matrix Factorization}

One of the most powerful tools for extracting latent features is factor analysis. In two dimensions factor analysis can be performed by various versions of Principle Component Analysis (PCA) \cite{jolliffe1986principal}, Independent Component Analysis (ICA) \cite{amari1996new}, or Non-negative Matrix Factorization (NMF) \cite{paatero1994positive}. The presence of the non-negativity constraint in NMF makes the extracted latent features physically interpretable since they are parts of the data \cite{lee1999learning}. Importantly, many variables, e.g., pixels, density, counts, etc., are naturally non-negative, and the extracted features will not have physical meaning if the non-negativity is not in place. 

We exploit an NMF-based framework that decomposes a matrix $\mat{X}$ into non-negative factors $\mat{W}$ and $\mat{H}$ where $\mat{W}$ is the matrix of basis patterns and $\mat{H}$ comprises the coefficients. The linear combination of columns of $\mat{W}$ describes and represents the matrix $\mat{X}$.  
Considering $\mat{X} \in \mathbb{R_+}^{M\times N}$,  $\mat{W} \in \mathbb{R_+}^{M\times k}$ and $\mat{H} \in \mathbb{R_+}^{k\times N}$, then $\mat{X}_{i,j} \approx \sum_{l=1}^K \mat{W}_{i,l}\mat{H}_{l,j}$. NMF decomposes $\mat{X}$ into $\mat{W}$ and $\mat{H}$ with a virtue of minimizing a distance between $\mat{X}$ and $\mat{W}\mat{H}$ i.e $min ||\mat{X}-\mat{W}\mat{H}||_{dist}$ using non-negative constraints $\mat{W}\geq 0$ and $\mat{H}\geq 0$. The distance can be either the eucledian distance considering the gaussian distribution of the data or KL-divergence for which the data distribution is poisson. The norm for minimization based on the distribution of the data is based on the expectation-minimization(EM) algorithm. In matrix $\mat{X}$, the component $\mat{X}_{i,j}$ correspond to the $i-th$ feature of the $j-th$ observation. 
Here, it is vital to know the latent dimenisonality $k$ priori, which is a challenging problem in most of the unsupervised dimensionality reduction technique. Some of the earlier works in estimation of K are based on automatic relevance determination{ARD}\cite{morup2009automatic}, neural net based\cite{nebgen2020neural}.  In the next section, we present a technique that is used for the purpose of selection of K based on Silhouette analysis \cite{alexandrov2019nonnegative}. 
\begin{figure*}
    \centering
    \includegraphics[width=.8\linewidth]{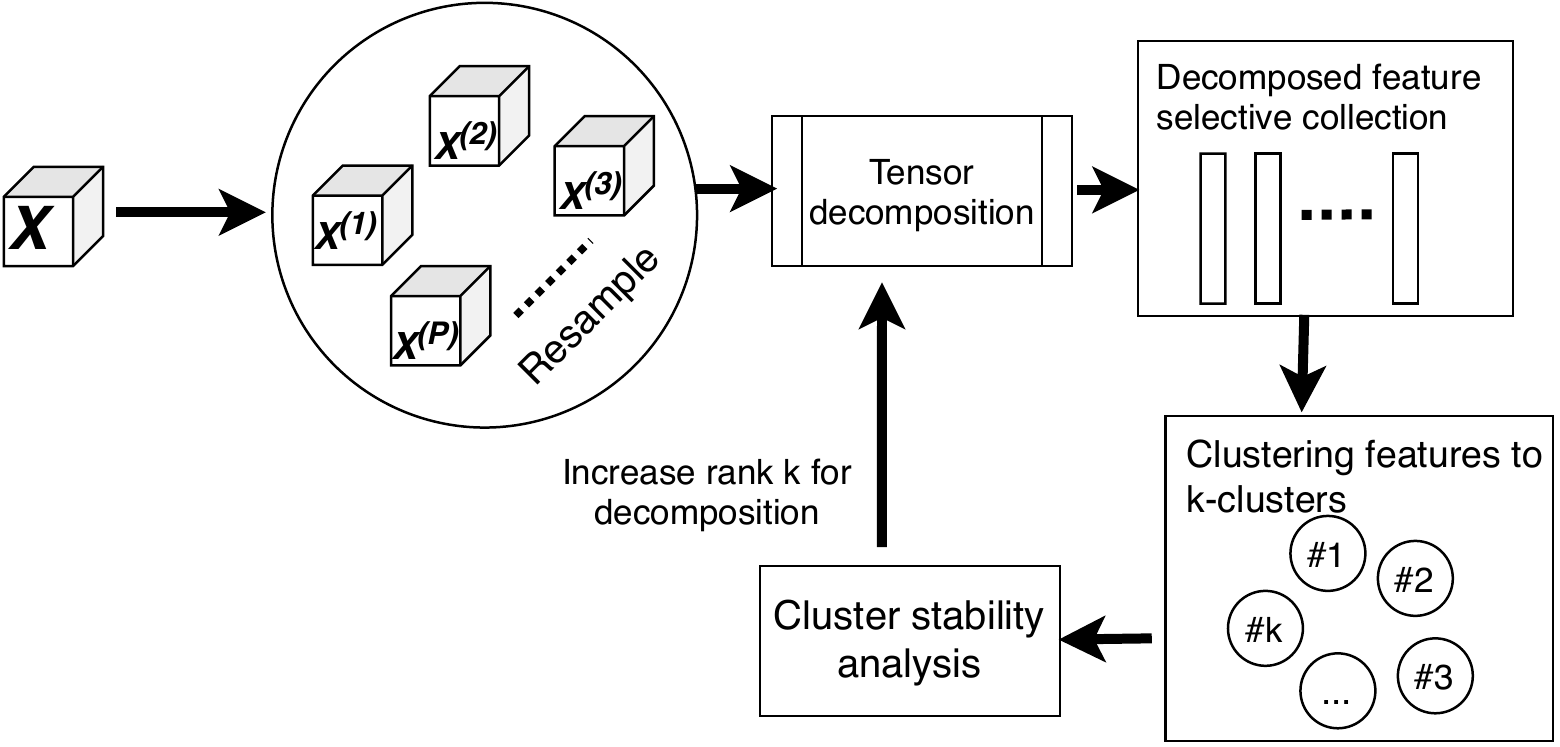}
    \caption{Latent dimension k estimation for tensor decomposition}
    \label{fig:nmfk}
\end{figure*}

To perform the tensor decomposition, we adopt a nTD-1 equivalent method, which unfolds/flattens the original volumetric data-tensor, $\ten{X}(t, x, y, z)$, along its time dimension. Thus, nTD-1 converts  the $4$-dimensional tensor, $ \ten{X}(t, x, y, z)$, to a $2d$ representation, $ \mat{A}^{(1)}(t, x*y*z)$, see Figure (\ref{fig:tensor_Decomp}). Further, nTD-1 uses NMF to extract the latent structure of $ \mat{A}^{(1)}(t, x*y*z)$. Specifically, NMF approximates, $\mat{A}^{(1)}\in \mathbb{R}_{+}^{t \times x*y*z}$, with a product of two non-negative factor matrices, $\mat{W} \in \mathbb{R}_{+}^{t\times K}$ and  $\mat{H} \in \mathbb{R}_{+}^{K \times x*y*z}$ (Figure (\ref{fig:tensor_Decomp})), such that the difference,
\begin{equation}\label{eq:nmf_cost}
\begin{aligned}
O = \Arrowvert \mat{A}^{(1)}(t, x*y*z)-\sum\displaylimits_{s=1}^{K}W_{s}(t)  H_{s}(x*y*z)\Arrowvert_{dist}\, \\
W_{s}(t)\geqslant0;\; H_{s}(x*y*z)\geqslant0.
\end{aligned}
\end{equation}
is minimal under some distance, $\Arrowvert ... \Arrowvert_{dist}$ and for a given small latent dimension $K$. Here, for $\Arrowvert ... \Arrowvert_{dist}$ we use Kullback–Leibler divergence: $D_{KL}(X || Y) = \sum_{i,j} X_{i,j} \log(\frac{X_{i,j}}{Y_{i,j}}) - X_{i,j} + Y_{i,j}$.
\subsubsection{CP-decomposition}
Canonical Polyadic decomposition(CPD/PARAFAC/CANDECOMP) also known is a tensor decomposition technique that decomposes a tensor into sum of component rank-one tensor. It is the generatlization of matrix singular value decomposition(SVD) to higher order tensors. For example, a 3-D tensor $\ten{X}\in \mathbb{R}^{I \times J\times K}$ can be expressed as 
\begin{equation}
    \ten{X} \approx \sum_{p=1}^{K} a_p \circ b_p \circ c_p ,
\end{equation}
 where K is the rank of decomposition, and factors $a_k \in \mathbb{R^{I}}$, $b_k \in \mathbb{R^{J}}$ and $c_k \in \mathbb{R^{K}}$ for $k=1,.....,K$. Alternately, equation can also be expressed as 
 \begin{equation}
    \ten{X}_{i,j,k} \approx \sum_{p=1}^K a_{ip}b_{jp}c_{kp},
 \end{equation}
for $i=1,\hdots ,I$,  $j=1,\hdots ,J$ and $k=1,\hdots ,K$.

Non-negative CPD decomposition is a special case of CPD where each of the factors are constrained to be non-negative. For example, in the previous example, $a_k \in \mathbb{R_+^{I}}$, $b_k \in \mathbb{R_+^{J}}$ and $c_k \in \mathbb{R_+^{K}}$

\subsubsection{Tucker decomposition}
Tucker decomposition, also a from of Higher order SVD(HODSVD)/PCA  decomposes a tensor into a core multiplied by a matrix along each mode. For a 3-D tensor $\ten{X} \in \mathbb{R}^{I \times J \times K}$, the decompositon is expressed as 
\begin{equation}
    \ten{X} = G \times_1 A \times_2 B \times_3 C \\
    = \sum_{p=1}^P \sum_{q=1}^Q \sum_{r=1}^R g_{pqr} a_p \circ b_q \circ c_r ,
\end{equation}
where $A \in \mathbb{R}^{I\times P}$,$B \in \mathbb{R}^{J\times Q}$ and $C \in \mathbb{R}^{K\times R}$ are called factors and $G\in \mathbb{R}^{P\times Q\times R}$ is the core tensor which shows interaction between different factors. Here, $\times_n$ represents n-th mode product. 
Elementwise, we have
\begin{equation}
    \ten{X}_{ijk} = \sum_{p=1}^P \sum_{q=1}^Q \sum_{r=1}^R g_{pqr} a_{ip} \circ b_{jq} \circ c_{rk} ,
\end{equation}
for $i=1,\hdots ,I$,  $j=1,\hdots ,J$ and $k=1,\hdots ,K$.
For a non-negative Tucker decomposition, the factors and the core are constrained to be non-negative, i.e., $A \in \mathbb{R}_+^{I\times P}$,$B \in \mathbb{R}_+^{J\times Q}$, $C \in \mathbb{R}_+^{K\times R}$ and $G_+\in \mathbb{R}^{P\times Q\times R}$

\subsubsection{Tensor train decomposition}
Tensor train(TT) decomposition decomposes a higher order tensor into a train of tensors of order 3. It is an efficient representation of a higher order tensor in term of storage requirements and computational robustness.  Tensor train decomposes a $3$-dimensional tensor $\ten{X}\in \mathbb{R}^{I \times J \times K}$ into \textit{d} 3-dimensional tensors $\ten[i]{G} \in \mathbb{R}^{r_{i-1}\times n_{i}\times r_{i}}$, where $r_{0} = r_{3} = 1$ (so $\ten[1]{G}$ and $\ten[3]{G}$ are actually matrices) and $r_{k} \geq 1$ for $k = 1,2$,  such that


\begin{equation}
\begin{aligned}
    \ten{X} = \ten[1]{G} \circ \ten[2]{G} \circ \ten[3]{G},
    \end{aligned}
\end{equation}
 Here, the 3-dimensional tensors $\ten[i]{G}$ are called  \textit{TT cores} and the numbers $r_{1},r_{2}$ are called  \textit{TT ranks}. 
Considering $\ten{A}=\ten[1]{G}$, $\ten{B}=\ten[2]{G}$ and $\ten{C}=\ten[3]{G}$, then
\begin{equation}
\begin{aligned}
    \ten{X} = \ten{A} \circ \ten{B} \circ \ten{C},
    \end{aligned}
\end{equation}

Based on the above, any element of the tensor $\ten{A}$ can be further be represented as

\begin{equation}
\begin{aligned}
\ten[][i_{1},i_{2},i_{3}]{X} = \sum_{k_1, k_2}^{r_1,r_2} \ten[][i_1,k_1]{A} \ten[][k_1,i_2,k_2]{B} \ten[][k_{2},i_3]{C}.
\end{aligned}
\end{equation}
Similar to non-negative CPD and non-negative Tucker,  a non-negative Tensor train requires  the TT-cores to be non-negative i.e , $A \in \mathbb{R}_+^{I\times P}$,$B \in \mathbb{R}_+^{P\times J \times Q}$, $C \in \mathbb{R}_+^{Q\times R}$. 

\subsection{Model Selection}

The model selection process corresponds to estimating the optimal number of latent features for the dataset. This is achieved via a clustering procedure first introduced in \cite{alexandrov2013deciphering}. For each decomposed factor, the procedure involves three stages to find the latent dimension k. They are i) bootstrapping by resampling data, ii) decompose the bootstrapped data with NMF, and iii) Analyze the cluster stability of the solutions. The details is presented in \ref{fig:nmfk}. 
 An ensemble of factors $\{\mat{X}^{(1)},\mat{X}^{(1)}.....\mat{X}^{(P)}\}$ sampled from $X^{(p)}_{i,j} \sim U(1-\epsilon,1+\epsilon)* \mat{X_{i,j}}$ where $U(a,b)$ is a uniform distribution over interval $[a,b]$. These bootstrapped variants of data $\mat{X}$ are then decomposed with NMF based on multiplicative updates/KL divergence into factors W and H. Each $\mat{X}^{(k)}$ is decomposed into corresponding $\mat{W}^{(k)}$ and $\mat{H}^{(k)}$. Next, a custom clustering based on \cite{} is used to exploit the stability of NMF's solutions corresponding to the bootstrapped data. Each cluster should contain precisely one feature vector from each $W^{(p)}$ based on the custom clustering. The algorithm based on K-means then iterates over each $W^{(p)}$ to assign each vector to an appropriate centroid where the assignment is done greedily applied to cosine similarity between columns of $W^{(p)}$ and current centroids. Next, silhouette statistics\cite{rousseeuw1987silhouettes} is then applied to analyze the quality of clustering, which measures the closeness of a point to other points in the same cluster and farness of the same point to the closest point in other clusters. 
 Silhouette is computed as ,
 \begin{equation}
     Silhouette(j) = \frac{b(j)-a(j)}{max(a(i),b(j))},
 \end{equation}
 where $a(i)$ is the measure of average distance of point j from other points in same cluster, and $b(j)$ is the minimum avergage distance from the point j to points in different cluster. 
 Based on this analysis, one can observe a well-separated cluster's compactness with a high silhouette score and an overlapping cluster with a low silhouette score. A combination of both minimum and mean silhouette score is used to quantify the cluster quality in this work. 
Further, to select the optimal number of latent features, the silhouette score and the relative error are used. The correct latent dimension ideally results in perfect clustering of solutions and the smallest relative error.

\subsection{The Salient Time Step Selection}

NMF is underpinned by a statistical generative model of superimposed components that can be treated as latent variables of Gaussian, Poisson, or other mixed model \cite{fevotte2009nonnegative}. In our case, the $K$-columns, $W_{s}(t)$ of $\mat{W}$, represent the latent time-features, while the $K$-columns, $H_{s}(x*y*z)$ of the transposed matrix $\mat{H}$, are the corresponding space-factors. After the extraction of the factor-matrices, $\mat{W}(t, K)$ and $\mat{H}(K, x*y*z)$, we reshape the matrix $\mat{H}(K, x*y*z)$ to construct the $3$-dimensional tensors, $\ten{H}_s(x, y, z)$. Each $\ten{H}_s(x, y, z)$ corresponds to a $3$-dimensional space-feature, and we have,
\begin{equation}
\ten{X}(t,x,y,z) = \sum_{s=1}^{K}W_{s}(t)\ten{H}_{s}(x,y,z)+\ten{\mathcal{E}}(t,x,y,z),
\label{NMF}
\end{equation}
Figure (\ref{fig:tensor_Decomp}), panel C. In (Eq. \ref{NMF}) $\ten{\mathcal{E}}$ is the tensor error of minimization.

We leverage this observation (Eq. \ref{NMF}) to select the timesteps most strongly associated with each space-feature\cite{pulido}. This is done simply by finding the largest value index in each of the $K$ columns of $\mat{W}$. This mapping provides us with $K$  easy interpretable features, $\ten{H}_{s}(x,y,z)$, each associated with a specific "influential" time point, the set of which we call the optimal latent salient timesteps.

\section{Experiments and Results}

\begin{figure}[!ht]
    \centering
    \includegraphics{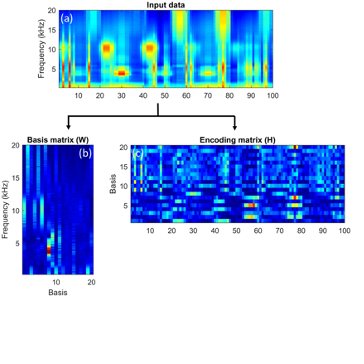}
    \caption{Spectral decomposition with NMF}
    \label{fig:nmf_res}
\end{figure}

\begin{figure}[t!]
    \centering
    \begin{subfigure}[t]{0.45\textwidth}
    \includegraphics[width=\textwidth]{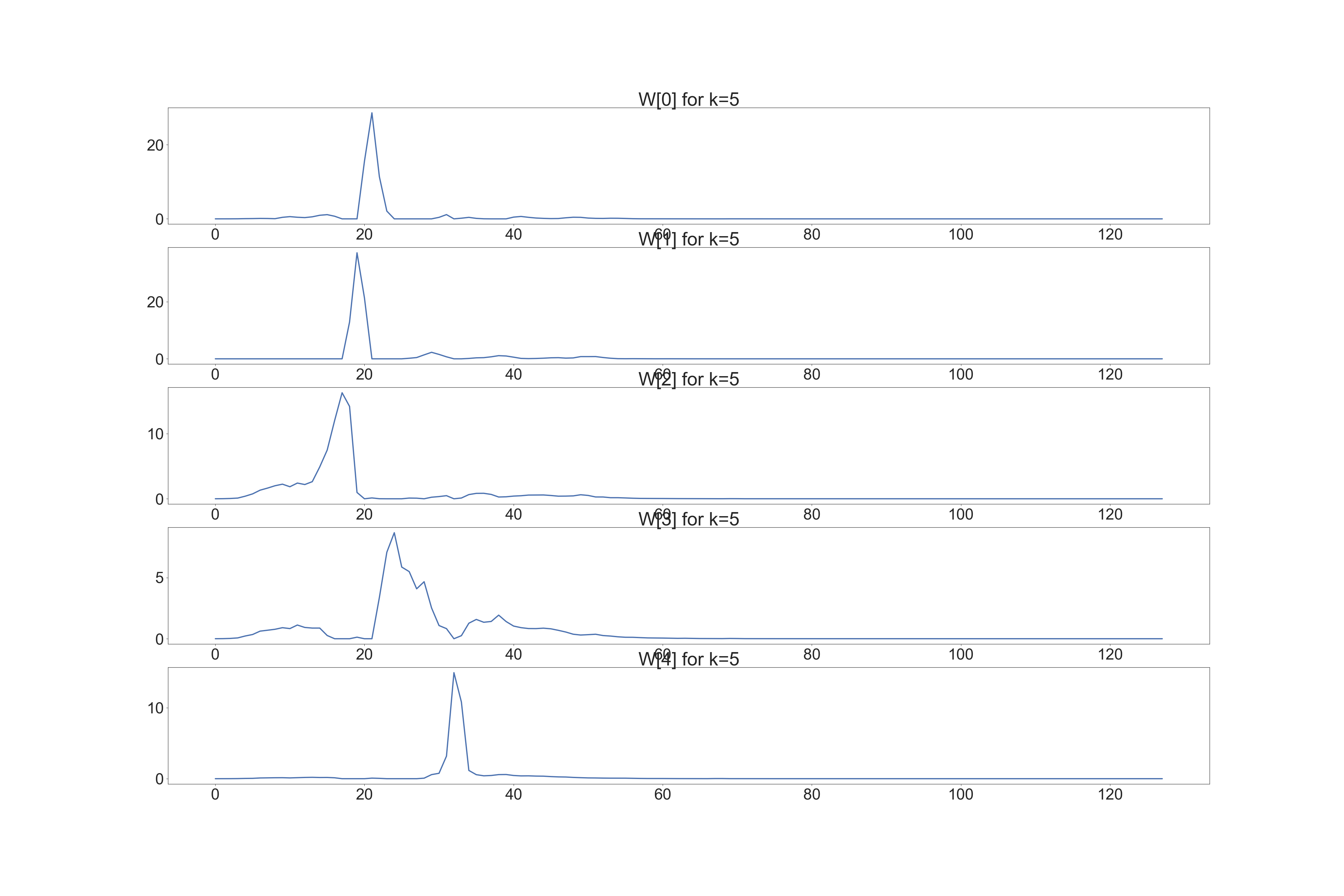}
    \caption{W features}
    \label{fig:W_feat}
    \end{subfigure}
    \hfill
    \begin{subfigure}[t]{0.45\textwidth}
    \includegraphics[width=\textwidth]{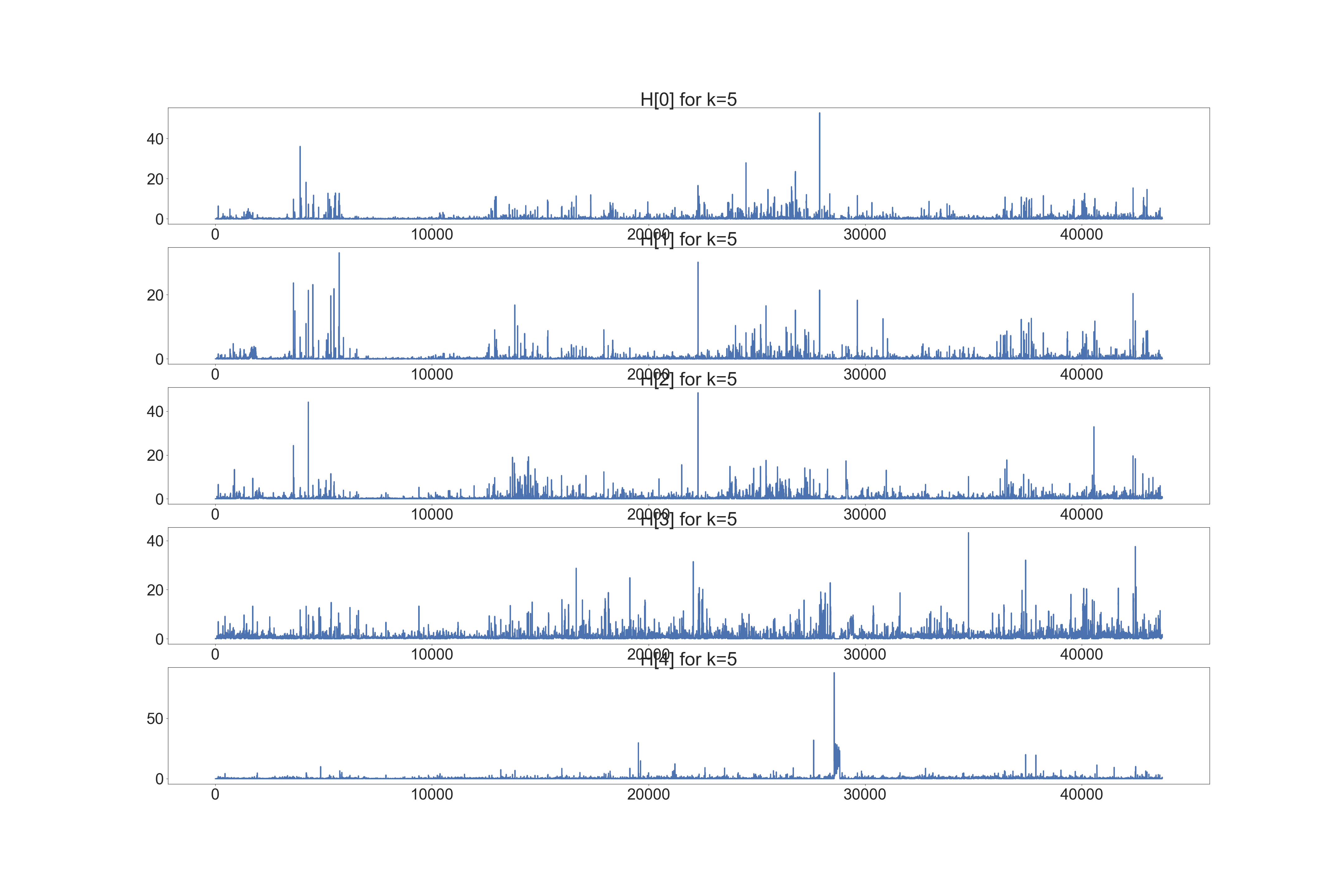}
    \caption{H features}
    \label{fig:H_feat}
    \end{subfigure}
    \caption{Decomposed feature components}
    \label{fig:feature}
\end{figure}

This section describes the proposed method's performance for speech and video datasets for anomaly detection and source separation.

Firstly, the audio recorded from the firefighter's microphone is used to analyze the firefighter's stress levels. This information can also be combined with other sensors such as heat, blood pressure, and body sensors to provide more accurate monitoring of firefighter's health on duty. The audio recordings of the firefighter were used for matrix factorization with appropriate preprocessing steps. Also, we utilized the videos recorded by the thermal camera to analyze the heat levels and temperature using tensor decomposition. 
\par
We used a multiplicative update-based NMF for decomposing the audio file, whereas Tucker and CPD were used to decompose the video file. The preprocessing step for the audio file involved computing the time-frequency spectra with the Mel spectrogram, which results in a  better decomposition compared to Short Time Fourier Transform(STFT). Further, a KL-divergence based NMF was used for decomposing the Mel spectrogram to decompose the time-frequency matrix into time and frequency components. The number of the latent features K was computed with the NMFk algorithm presented in 7.2.2. Here W returns the frequency components, and H returns the timestamps where the frequency components are dominant. A speech activity can be located precisely in time, and the strength is determined. For any explosions, high breath levels, specific fighter speaking, NMF can identify the activity, and an alarm system can be based on this. \par
An example is presented in figure \ref{fig:nmf_res}, which displays the Mel spectrogram and the factor matrices. Further, \ref{fig:feature} shows the time and frequency components in detail. For a specific audio recording, in \ref{fig:feature}, we can observe the fundamental five latent components corresponding to different audio sources. This could be multiple firefighters, background noise, firefighters breathing, and explosion sound. The W features correspond to the frequency of the sources where the narrow frequencies correspond to a source, whereas the broad ones correspond to the background noise. The H matrix components correspond to the activation coefficients, i.e., the time instances for the activity. This source separation technique aids in finding anomalies and unusual incidents when passed through a system trained to identify such events. 
\par

\begin{figure}[!ht]
    \centering
    \includegraphics[width=.9\linewidth]{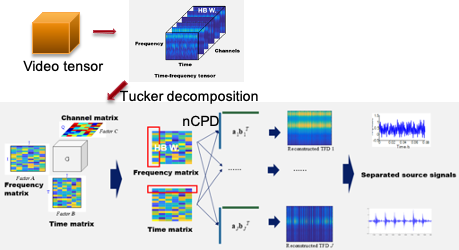}
    \caption{Tensor decompositions 
for detecting anomalous activity
from video frames}
    \label{fig:tensor_decomp}
\end{figure}
We applied Tucker and CPD based tensor decomposition to identify the rise in heat levels and anomaly events for the videos. The video tensor is first converted into a time-frequency matrix for each video frame channel with a discrete-time Fourier transform(DTFT). A tucker decomposition is then applied to this tensor to extract the time, frequency, and channel components.  Again, we used the NMFk algorithm presented in section 7.2.2 to compute the number of latent features along each tensor dimension. Next, we applied nCPD decomposition to the factor matrices to find the underlying sources and time and frequency. With this complete procedure, the anomalous event's time is identified, and nature is determined based on the frequency. Further, a lightweight neural net system can be trained to identify the event based on the decomposed features. With this approach, we can observe the change in directions of time, channel, and space. To be precise, we can precisely decompose the Spatio-temporal correlations along with each component for better understanding and inferences. The details of the proposed algorithm are shown in \ref{fig:tensor_decomp}. 
\par

The implementation was performed with LANL developed software libraries \cite{bhattaraipyDNMFk} and \cite{bhattaraipyDNTNK}. 

\chapter{Conclusions and Future Works}

\section{Conclusion}
The scene of a structural fire is chaotic, dangerous, and disorienting. Heavy smoke, near-zero visibility, extreme heat, and flames create a perfect storm for stress-induced misjudgments that can affect even seasoned firefighters. These are prime conditions for computer-aided assistance and artificial intelligence solutions to supplement human knowledge and experience. Poor judgment in high-stress situations is cited among the top reasons for fire-related fatality. The evidence suggests that aiding firefighters battling their disorientation and maintaining a constant understanding of their surroundings will prevent fatalities. Our research provides an artificial intelligence solution that can supplement firefighters with real-time information and offer guidance by automatically interpreting the fireground from the information provided by the handheld thermal cameras already in use by firefighters.

 This work provides a foundational deep learning framework capable of identifying objects within the environment and incorporating those objects into its decision-making process to deliver safe, navigable routes to firefighters successfully.  We present a deep learning-based technology that is the first of its kind in the firefighting environment. It can accurately detect objects of interest utilizing thermal imagery actively recorded on the scene. While previously published, Convolutional Neural Networks has demonstrated outstanding performance in object detection on RGB imagery; very little work has been applied to thermal imagery. The CNN developed for this application is the first of its kind to achieve such high accuracy on thermal imagery and is unique in its ability to detect, track, and segment multiple objects simultaneously within a frame. Our CNN can also conduct this processing, analysis, and result generation in real-time if a camera is connected to a simple single-board commercial computer endowed with GPU capabilities (for example, an NVIDIA JETSON). This is crucial in near-zero visibility conditions where it can be very easy to become disoriented and then miss key details that would allow for escape. Our system can identify key objects of interest, such as doors, windows, ladders, and people, even when those objects are partially obscured. We also incorporated a Natural Language Processing (NLP) system, which is capable of scene description and allows the framework to autonomously make human-understandable descriptions of the environment to help firefighters improve their understanding of the immediate surroundings and assist them when anxiety levels are heightened. The framework’s ability to not only identify human figures but also differentiate between postures further adds to its utility in search and rescue as posture can provide a rough estimate of the health status of a victim (for instance, a prone posture may indicate a person who has succumbed to the effects of smoke inhalation and requires immediate evacuation and paramedic assistance). Thus posture recognition and differentiation could assist in the prioritization of rescue. Our framework performs with greater than 95\% accuracy on the detection of these objects of interest and more than 90\% accuracy on posture identification. We also present an evaluation of computational time to accuracy achievement trade-off and show that the model performs above 70\% even at the lowest convolutional depth, making it highly adept for usage in the field where computational power is likely to be limited. The improvements gained through implementing this technology could significantly enhance the firefighter’s ability to maintain a constant and consistent situational awareness, considerably improving their ability to correctly interpret surroundings and maximize rescue efficiency and effectiveness while minimizing risk.\par

 Creating a virtual guide capable of tracking objects of interest and presenting an escape plan to a disoriented firefighter would add a finishing touch to creating an artificially intelligent solution to aid firefighters. To lay the foundation for this future work, we also present a deep Q-learning based agent trained in a virtual environment that can quickly adapt to a rapidly changing environment and adaptively make navigation decisions. Previous works have not achieved the ability to change course once a path is chosen, but our agent can backtrack and discern new routes as soon as an obstacle appears in its path. The Unreal engine was used to simulate the fire environment, and AirSim was used to communicate data and controls between the virtual environment to the deep learning model. The agent was successfully able to navigate extreme fires based on its acquired knowledge and experience.

\par
We developed a prototype embedded system to implement this detection to showcase its ability to analyze, process, and relay relevant scene information back to firefighters in real-time. This prototype is capable of initializing and capturing real-time thermal images from a FLIR One G2 camera, processing those images for object detection, object tracking, instance segmentation, and scene description on an NVIDIA Jetson TX2 embedded GPU development platform, and returning the output as a real-time live stream over a secure wireless network to first responders and commanding officers alike. Besides, the embedded system is also able to acquire and process RGB color and depth images captured by an Intel RealSense D435i camera on an NVIDIA Jetson AGX Xavier embedded GPU development platform and to return the output as a second real-time live stream over the same secure wireless network.

There are several challenges to the successful implementation of a smart and connected firefighter system, and primary among them is the challenge of providing robust, real-time data acquisition, processing, communication, and augmentation. This research provides a foundation on which solutions to these challenges can be built. Among these challenges is the integration of such technology into the gear of first responders. In numerous interviews with first responder crews, the main concern raised centers on this integration of the proposed system in their gear that is not cumbersome or distracting given their small field of view through their masks. It is challenging for them to add equipment to their standard gear without disturbing their regular operation. Therefore, the accommodation of this equipment and the design of suitable interfaces for firefighters needs further investigation.
Nevertheless, the present proposal is a proof of concept of the usage of these technologies by first responders. The design of a usable prototype is out of the research scope but is the objective of future work. The final prototype must consider the usability of the interface and the elimination of all interference with the emergency response.

 \section{Future Works}


The proposed system is intended to be integrated into a geographic and visual environment with data of the floor plan, which will be updated as firefighters progress through the scene to include information about the fire locations, doors, windows, detected firefighters, health condition of the firefighters and other features that are collected from the sensors carried in the firefighter gear, which will be transmitted over a robust communication system to an incident commander to produce a fully flexed situational awareness system. This can be achievied via the use of the deep learning-based results such as object detection, tracking, and segmentation to create a more informative situational awareness map of the reconstructed 3d scene. 
\par


Our next steps will focus on fusing the data collected from the different cameras to form a 3D map of the building. The data will form the foundation of a new AI system that assists the firefighters with path planning through the cataloging of doors, windows, and paths clear of fire. We will also address the hardware issues that currently limit the embedded system’s application at the fireground. We also propose using advanced AI techniques such as generative adversarial neural nets (GAN’s) to identify objects from obscured/ partially perceivable features from imagery. GAN system is also capable of filling in the missing information for completing the sparse 3d reconstructed map.
Most importantly, they will be used to predict the flashovers/ fire explosions in advance to aid in firefighters’ safety. Next, a foundation for developing a real-time situational map of the structural configuration of a building that is actively built and updated via the live thermal imagery being recorded by firefighters moving through the scene is proposed. This map, which is updated in real-time, could be used by firefighters to safely navigate the burning structure and improve the situational awareness necessary in decision making by tracking exits that may become blocked and finding alternatives.

\par
Next, we also aim to fuse RGB, Infrared, and depth information collected using different imaging cameras live from fire scenes to reconstruct the 3D multispectral map of the building. This can be accessed by the fire commander to access the risk of fire, situational awareness of the building in 360 video/ 3D that can be used for decision making and setting effective control plans for the firefighters. The multispectral imaging helps to access the temperature conditions at fire regions and structural damages of those zones, and the level of firefighting required to address the needs. We will combine these different information types into a blended 3d scene, which can be viewed on augmented reality (AR) / virtual reality device. Such an environment can be imported into a gaming environment such as unreal engine/ Unity to train the AI agent using deep reinforcement learning (RL). Given the feed of a real environment with burning objects and simulated fire, the AI agent can be trained to avoid fire and navigate safely through the fire with the aid of a model that works based on rewards and penalties. The model can be developed with the deep RL so that the agent learns the best to achieve the objectives by maximizing the sum of rewards on the game, safely navigating along fire to reach the destination. Once the agent learns the scene, it can be deployed in a real-world situation to assist the firefighter in path planning and navigation by using the data collected from the same source used to construct the gaming environment. Such a simulated AR environment can also provide an immersive experience of a real fire to firefighters in different scenarios and make them more prepared to be effective under different firefighting environments. 
\par
Ultimately, this research aims to provide the foundation for building a system capable of automatically creating a geographic and visual environment using a combination of floor plan data and streamed data collected from firefighters traveling through the building. This environment can depict active and up-to-date fires locations, firefighter locations, and conditions and be communicated reliably over wireless networks. The results can be visualized on an Augmented Reality system by the incident commanders and improve their awareness of the fire environment’s physical conditions. This improved understanding of the ground conditions can improve their ability to instruct firefighters in the navigation of the 3d floor plan of the structure and better predict fire change through a live temperature map of the building for advanced structure assessment and necessary planning. This information can lead to better knowledge for strategic, tactical, communication, and safety decisions to achieve SOP for firefighters.

\bibliographystyle{ieee}
\bibliography{main}

\end{document}